# BoundMPC: Cartesian path following with error bounds based on model predictive control in the joint space

**Thies Oelerich[1], Florian Beck[1], Christian Hartl-Nesic[1] and Andreas Kugi[1,2]**

**Abstract**

*This work introduces the* BoundMPC *strategy, an innovative online model-predictive path-following approach for robot manipulators. This joint-space trajectory planner allows the following of Cartesian reference paths in the end-effector's position and orientation, including via-points, within the desired asymmetric bounds of the orthogonal path error. These bounds encode the obstacle-free space and additional task-specific constraints in Cartesian space. Contrary to traditional path-following concepts,* BoundMPC *purposefully deviates from the Cartesian reference path in position and orientation to account for the robot's kinematics, leading to more successful task executions for Cartesian reference paths. Furthermore the simple reference path formulation is computationally efficient and allows for replanning during the robot's motion. This feature makes it possible to use this planner for dynamically changing environments and varying goals. The flexibility and performance of* BoundMPC *are experimentally demonstrated by five scenarios on a 7-DoF KUKA LBR iiwa 14 R820 robot. The first scenario shows the transfer of a larger object from a start to a goal pose through a confined space where the object must be tilted. The second scenario deals with grasping an object from a table where the grasping point changes during the robot's motion, and collisions with other obstacles in the scene must be avoided. The adaptability of* BoundMPC *is showcased in scenarios such as the opening of a drawer, the transfer of an open container, and the wiping of a table, where it effectively handles task-specific constraints. The last scenario highlights the possibility of accounting for collisions with the entire robot's kinematic chain. The code is readily available at https://github.com/TU-Wien-ACIN-CDS/BoundMPC, inspiring you to explore its potential and adapt it to your specific robotic tasks.*

## 1. Introduction

Robot manipulators are multi-purpose machines with many different applications in industry and research, which include bin picking (Bencak et al., 2022), physical human–robot interaction (Ortenzi et al., 2021), and industrial processes like welding (Lei et al., 2020). During the execution of these applications, various constraints have to be considered, such as safety bounds, collision avoidance, task-specific constraints, and kinematic and dynamic limitations. In general, this leads to challenging trajectory-planning problems. A well-suited control framework for such applications is model predictive control (MPC) (Faulwasser et al., 2009). Using MPC, the control action is computed by optimizing an objective function over a finite planning horizon at every time step while respecting all constraints. Online trajectory planning with MPC is particularly useful for robots acting in dynamically changing environments. Especially when there is uncertainty about actors in the scene, it is crucial to appropriately react to changing situations, for example, when interacting with humans (Li et al., 2021). Trajectory planners that plan the full trajectory from a start to a goal pose (e.g., Jankowski et al., 2023; Marcucci et al., 2023) lack the computational efficiency to react online in these situations.

Robotic tasks are often specified as Cartesian reference paths, parametrized by a path parameter. To execute these tasks accurately, path-following controllers are essential.

[1]Automation and Control Institute (ACIN), TU Wien, Wien, Austria
[2]AIT Austrian Institute of Technology GmbH, Wien, Austria

**Corresponding author:**
Thies Oelerich, Automation and Control Institute (ACIN), TU Wien, Gußhausstraße 27-29 / E376, Wien 1040, Austria.
Email: oelerich@acin.tuwien.ac.at

These controllers ensure that the robot's end-effector follows the specified path with precision and efficiency (Van Duijkeren et al., 2016). Examples include opening a door, where the end-effector has to follow an arc-shaped path in Cartesian space, opening a drawer with a linear motion, and the transfer of an open container, which must be held upright during the motion. These tasks are challenging for a planner that only plans trajectories based on joint-space objectives. In contrast, a Cartesian reference path consists of a series of Cartesian via-points that the robot's end-effector needs to pass exactly. It is advantageous to decompose the error between the current position and the desired reference path point (Romero et al., 2022) into a tangential error, representing the error along the path and an orthogonal error. The tangential error is minimized to progress along the path quickly while the orthogonal error is bounded (Arrizabalaga and Ryll, 2022).

Providing a reference path or series of via-points in Cartesian space is intuitive. However, due to the complex robot kinematics, trajectory planning is complicated compared to reference paths or via-points in the robot's joint space. Furthermore, Cartesian reference paths with path error bounds can simplify collision avoidance and can be used to encode task-specific constraints. Specifying these bounds in the joint space is complicated and requires large computational effort, as shown in Marcucci et al. (2023). Existing work often minimizes the path error to follow the path exactly (Arbo et al., 2017; Astudillo et al., 2022b). This works well if the given path is known to be executable but may lead to the robot getting stuck for challenging paths.

This work introduces a novel online trajectory planner in the joint space, *BoundMPC*, to follow a Cartesian reference path with the robot's end-effector while bounding the orthogonal path error. Furthermore, position and orientation reference paths with via-points are considered. A schematic drawing of the geometric relationships is shown in Figure 1. By bounding the orthogonal path error and minimizing the tangential path error, the MPC can find the optimal trajectory according to the robot dynamics and other constraints. The robot purposefully deviates from the path to take the robot's kinematics into account, which allows the robot to follow even more challenging paths. Simple piecewise linear reference paths with via-points are used to simplify online replanning, and (asymmetric) polynomial error bounds allow us to adapt to different scenarios and obstacles quickly. The end-effector's position and orientation reference paths are parameterized by the same path parameter, ensuring synchronicity at the via-points. The Lie theory for rotations is employed to describe the end-effector's orientation, which yields a novel way to decompose and constrain the orientation path error. Moreover, the decomposition of the orthogonal error using the desired basis vectors of the orthogonal error plane allows for meaningful asymmetric bounding of both position and orientation path errors.

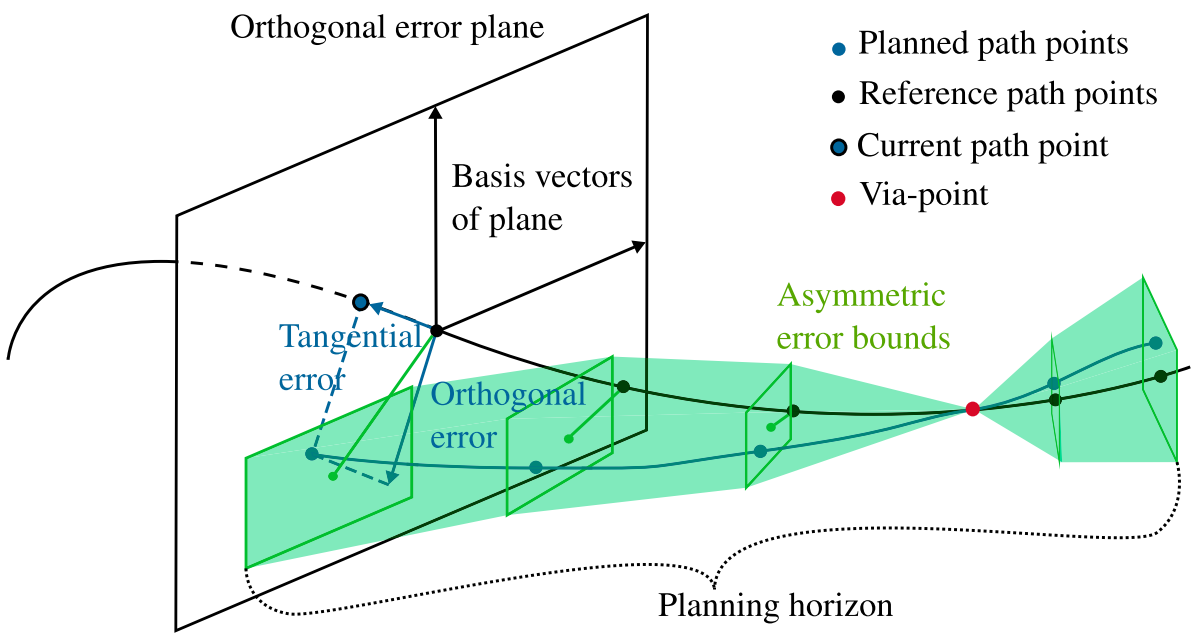

**Figure 1.** Schematic of the position trajectory planning using *BoundMPC* in the 3D Cartesian space. The tangential and orthogonal errors are shown for the initial position with the orthogonal error plane spanned by two basis vectors. The planned path over the planning horizon is within the asymmetric error bounds, depicted as shaded green area.

The paper is organized as follows: Related work is summarized in Section 2, and the contributions extending the state of the art are given in Section 3. The general MPC framework *BoundMPC* is developed in Section 4. The used orthogonal error bounds are described in Section 5. The use of a piecewise linear reference path and its implications are presented in Section 6. Afterward, Section 7 shows the online replanning. The implementation details for the simulations and experiments are given in Section 8. Section 9 deals with the parameter tuning of the MPC. Six experimental scenarios on a 7-DoF KUKA LBR iiwa 14 R820 robot are demonstrated in Section 10, which emphasize the ability of *BoundMPC* to replan trajectories online during the robot's motion and to specifically bound the orthogonal errors asymmetrically. Comparisons with existing works are performed, which show the advantages of the proposed framework. Section 11 gives some conclusions and an outlook on future research activities.

## 2. Related work

Path and trajectory planning in robotics is an essential topic with many different solutions, such as sampling-based planners (Elbanhawi and Simic, 2014; Karaman and Frazzoli, 2011; Persson and Sharf, 2014), learning-based planners (Mac et al., 2016; Mukherjee et al., 2022; Osa, 2022), and optimization-based planners (Faulwasser et al., 2009; Romero et al., 2022; Van Duijkeren et al., 2016), which can be further classified into offline and online planning. Online planning is needed to be able to react to dynamically changing environments. Sampling-based planners have the advantage of being probabilistically complete, meaning they will eventually find a solution if the problem is feasible. However, online planning is limited by computational costs and the need to smooth the obtained trajectories (Ferguson et al., 2006; Zucker et al., 2007). Constrained motion planning with sampling-based methods is discussed in Kingston et al. (2019). Optimization-based trajectory planners, such as TrajOpt (Schulman et al., 2014), CHOMP (Zucker et al., 2013), and CIAO (Schoels et al., 2020b), take dynamics and collision avoidance into account. However, they are generally too slow for online planning due to the nonconvexity of the planning problems. Receding horizon control plans a trajectory for a finite horizon starting at the current state to reduce computational complexity. For example, CIAO-MPC (Schoels et al., 2020b) extends CIAO to receding horizon control for point-to-point motions.

More complex applications require the robot to follow a reference path. Receding horizon control in this setting was demonstrated for quadrotor racing in Arrizabalaga and Ryll (2022) and Romero et al. (2022) and robot manipulators in Arbo et al. (2017); Astudillo et al. (2022b); and Mehrez et al. (2017). Additionally, Romero et al. (2022) include via-points in their framework to safely pass the gates along the racing path. Adapting the current reference path during the motion to react to a changing environment and a varying goal is crucial. While, in theory, the developed MPC frameworks in Arrizabalaga and Ryll (2022), Astudillo et al. (2022b), Lam et al. (2013), and Romero et al. (2022) could handle a change in the reference path, this still needs to be demonstrated for industrial manipulators. Furthermore, these frameworks need pre-optimized reference paths, which often requires the solution of a nonlinear optimization problem (e.g., Verschueren et al., 2016). This step must be performed before the path-following task, making the online replanning computationally infeasible. Alternatively, the orthogonal error bounds can be adapted to reflect a dynamic change in the environment, as done in Arrizabalaga and Ryll (2022).

To be able to use the path following formulations, an arc-length parametrized path is needed. Since the system dynamics are parametrized in time, a reformulation is needed to couple the reference path to the system dynamics as done in Arrizabalaga and Ryll (2022), Böck and Kugi (2016), Debrouwere et al. (2014), Spedicato and Notarstefano (2018), and Van Duijkeren et al. (2016). Such a coupling allows for a very compact formulation but complicates constraint formulations. Therefore, Lam et al. (2013) decouple the system dynamics from the path and uses the objective function to minimize the tangential path error to ensure that the path system is synchronized to the system dynamics.

Tracking the reference path by the planner results in a path error. The goal of the classical path following control is to minimize this error (Astudillo et al., 2022b; Faulwasser et al., 2009), which is desirable if the reference path is optimal. Since optimal reference paths are generally not trivial, exploiting the path error to improve the performance has been considered in Arrizabalaga and Ryll (2022) and Romero et al. (2022). Further control on utilizing the path error is given by decomposing it into tangential and orthogonal components. To safely traverse via-points, Romero et al. (2022) use a dynamical weighting of the orthogonal path error, achieving collision-free trajectories. Furthermore, the orthogonal path error lies in the plane orthogonal to the path and can thus be decomposed further using basis vectors of the plane. This way, surface-based path following was proposed in Hartl-Nesic et al. (2021), where the direction orthogonal to the surface was used to assign a variable stiffness. The choice of basis vectors is application specific. Concretely, Arrizabalaga and Ryll (2022) use the Frenet–Serret frame to decompose the orthogonal path error. It provides a continuously changing frame along the path but is not defined for path sections with zero curvature. An alternative to the Frenet–Serret frame is the parallel transport frame used in Bischof et al. (2017). The decomposition of the orthogonal error allows for asymmetric bounding in different directions of the orthogonal error plane, but this has yet to be demonstrated in the literature.

Many path-following controllers only consider a position reference trajectory and thus, a position path error. For quadrotors, this is sufficient since the orientation is given by the dynamics (Romero et al., 2022). The orientation was included in Astudillo et al. (2022a, 2022b) for robot manipulators, where the respective orientation error formulation allows arbitrarily large errors. However, their implementation minimizes the orientation path error, which is only feasible for a path known to be kinematically executable by the robot. For arbitrary paths, this may lead to the robot getting stuck in, for example, the joint limits. The orientation in Van Duijkeren et al. (2016) is even considered constant. In Hartl-Nesic et al. (2021), the orientation is set relative to the surface and is not freely optimized. Thus, the literature has yet to treat orientation reference paths without minimizing the orientation path error. Furthermore, a decomposition of the orientation path error analogous to the position path error is missing in the literature, which can further improve the performance of path-following controllers. In this work, orientation reference paths are included using the Lie theory for rotations, which is beneficial since the rotation propagation can be linearized. In Torres Alberto et al. (2022), this property was exploited for pose estimation based on the theory in Solà et al. (2021). Forster et al. (2017) use the same idea for visual-inertial odometry. In this work, the Lie theory for rotations is exploited to decompose the orientation path error and the propagation of this error over time.

## 3. Contributions

The proposed MPC framework *BoundMPC* combines concepts discussed in Section 2 in a novel way. It uses a receding horizon implementation to follow a reference path in the position and orientation with asymmetric error bounds in distinct orthogonal error directions. A path parameter system similar to Lam et al. (2013) ensures path progress. Using linear reference paths with via-points yields a simple representation of reference paths and allows for fast online replanning without providing optimal reference paths. Additionally, the orientation has a distinct reference path, and the orientation path errors are freely optimized to account for the robot's kinematics instead of minimizing them, as in previous works. The novel asymmetric error bounds allow more control over trajectory generation and obstacle avoidance. Through its simplicity of only providing the desired via-points, *BoundMPC* finds the optimal position and orientation path within the given error bounds. The main contributions are:

- A path-following MPC framework to follow position and orientation paths with synchronized via-points is developed. The robot kinematics are considered in the optimization problem such that a joint trajectory is optimized while adhering to (asymmetric) Cartesian constraints.
- The position and orientation path errors, computed based on the Lie theory for rotations, are decomposed and bounded to be freely optimized within the given bounds. This allows for more freedom during the trajectory optimization compared to existing approaches that minimize tracking errors.
- Piecewise linear position and orientation reference paths with via-points can be replanned online to quickly react to dynamically changing environments and new goals during the robot's motion.
- The framework is demonstrated on a 7-DoF KUKA LBR iiwa 14 R820 manipulator with planning times less than 100 ms. A video of the experiments can be found at https://www.acin.tuwien.ac.at/42d0/.

## 4. BoundMPC formulation

This section describes the BoundMPC formulation, which is a path-following concept for Cartesian position and orientation reference paths. This concept is schematically illustrated in Figure 1. The proposed formulation computes a trajectory in the joint space, requiring an underlying torque controller to follow the planned trajectory.

First, the formulation of the dynamical systems, that is, the robot manipulator and the jerk input system, is introduced. Second, the orientation representation based on Lie algebra and its linearization is used to calculate the tangential and orthogonal orientation path errors based on arc-length parameterized reference paths. A similar decomposition is also introduced for the position path error, finally leading to the optimal control problem formulation.

### 4.1. Dynamical system

The robot manipulator dynamics for $n$ joints are described in the joint positions $\mathbf{q} \in \mathbb{R}^n$, velocities $\dot{\mathbf{q}} \in \mathbb{R}^n$, and accelerations $\ddot{\mathbf{q}} \in \mathbb{R}^n$ as the rigid-body model

$$\mathbf{M}(\mathbf{q})\ddot{\mathbf{q}} + \mathbf{C}(\mathbf{q}, \dot{\mathbf{q}})\dot{\mathbf{q}} + \mathbf{g}(\mathbf{q}) = \boldsymbol{\tau}, \tag{1}$$

with the mass matrix $\mathbf{M}(\mathbf{q})$, the Coriolis matrix $\mathbf{C}(\mathbf{q}, \dot{\mathbf{q}})$, the gravitational forces $\mathbf{g}(\mathbf{q})$, and the joint input torque $\boldsymbol{\tau} \in \mathbb{R}^n$. For trajectory following control, the computed-torque controller (Siciliano et al., 2009)

$$\begin{aligned}\boldsymbol{\tau} = \mathbf{M}(\mathbf{q})\left(\ddot{\mathbf{q}}_\mathrm{d}(t) - \mathbf{K}_2\dot{\mathbf{e}}_\mathrm{q} - \mathbf{K}_1\mathbf{e}_\mathrm{q} - \mathbf{K}_0\int \mathbf{e}_\mathrm{q}\mathrm{d}t\right) \\ + \mathbf{C}(\mathbf{q}, \dot{\mathbf{q}})\dot{\mathbf{q}} + \mathbf{g}(\mathbf{q}),\end{aligned} \tag{2}$$

with the desired joint-space trajectory $\mathbf{q}_\mathrm{d}(t)$ and its time derivatives, is applied to (1). The tracking error $\mathbf{e}_\mathrm{q} = \mathbf{q} - \mathbf{q}_\mathrm{d}$ is exponentially stabilized in the closed-loop system (1), (2) by choosing suitable matrices $\mathbf{K}_0$, $\mathbf{K}_1$, and $\mathbf{K}_2$, typically by pole placement.

*4.1.1. Joint jerk parametrization.* To obtain the desired trajectory $\mathbf{q}_\mathrm{d}(t)$, a joint jerk input signal $\dddot{\mathbf{q}}_\mathrm{d}(t)$ is introduced, which is parameterized using hat functions similar to Hausberger et al. (2019). The hat functions $H_k(t)$, $k = 0, \ldots, K - 1$, given by

$$H_k(t) = \begin{cases} \dfrac{t - t_k}{T_\mathrm{s}} & t_k \le t < t_k + T_\mathrm{s} \\ \dfrac{t_k + 2T_\mathrm{s} - t}{T_\mathrm{s}} & t_k + T_\mathrm{s} \le t \le t_k + 2T_\mathrm{s} \\ 0 & \text{otherwise,} \end{cases} \tag{3}$$

are parametrized by the starting time $t_k$ and the width $2T_\mathrm{s}$. The index $k$ refers to an equidistant time grid $t_k = kT_\mathrm{s}$ on the continuous time $t$. For each joint at the discretization points, the desired jerk values are denoted by the vectors $\mathbf{j}_k \in \mathbb{R}^n$, $k = 0, \ldots, K - 1$. Thus, the continuous joint jerk input

$$\dddot{\mathbf{q}}_\mathrm{d}(t) = \sum_{k=0}^{K-1} H_k(t + T_s)\mathbf{j}_k \tag{4}$$

results from the summation of all hat functions over the considered time interval. The joint jerk (4) is a linear interpolation of the joint jerk values $\mathbf{j}_k$. The desired trajectory for the joint accelerations $\ddot{\mathbf{q}}_\mathrm{d}(t)$, joint velocities $\dot{\mathbf{q}}_\mathrm{d}(t)$, and joint positions $\mathbf{q}_\mathrm{d}(t)$ are found analytically by integrating (4) with respect to the time $t$ for $t_0 = 0$

$$\ddot{\mathbf{q}}_\mathrm{d}(t) = \ddot{\mathbf{q}}_{\mathrm{d},0} + \sum_{k=0}^{K-1} \int_0^t H_k(\tau + T_\mathrm{s})\mathrm{d}\tau \mathbf{j}_k \tag{5a}$$

$$\dot{\mathbf{q}}_\mathrm{d}(t) = \dot{\mathbf{q}}_{\mathrm{d},0} + \ddot{\mathbf{q}}_{\mathrm{d},0}t + \sum_{k=0}^{K-1} \iint_0^t H_k(\tau + T_\mathrm{s})\mathrm{d}\tau \mathbf{j}_k \tag{5b}$$

$$\begin{aligned}\mathbf{q}_\mathrm{d}(t) = \mathbf{q}_{\mathrm{d},0} + \dot{\mathbf{q}}_{\mathrm{d},0}t + \ddot{\mathbf{q}}_{\mathrm{d},0}\frac{t^2}{2} \\ + \sum_{k=0}^{K-1} \iiint_0^t H_k(\tau + T_\mathrm{s})\mathrm{d}\tau \mathbf{j}_k.\end{aligned} \tag{5c}$$

The integration constants $\mathbf{q}_\mathrm{d}(0) = \mathbf{q}_{\mathrm{d},0}$, $\dot{\mathbf{q}}_\mathrm{d}(0) = \dot{\mathbf{q}}_{\mathrm{d},0}$, and $\ddot{\mathbf{q}}_\mathrm{d}(0) = \ddot{\mathbf{q}}_{\mathrm{d},0}$ are given by the initial values at the time $t_0 = 0$.

*4.1.2. Linear time-invariant system.* In the following, (5) is described by a linear time-invariant (LTI) system with the state $\mathbf{x}^\mathrm{T} = [\mathbf{q}_\mathrm{d}^\mathrm{T}, \dot{\mathbf{q}}_\mathrm{d}^\mathrm{T}, \ddot{\mathbf{q}}_\mathrm{d}^\mathrm{T}]$. The corresponding discrete-time LTI system with the sampling time $T_\mathrm{s}$ then reads as

$$\mathbf{x}_{k+1} = \mathbf{\Phi}\mathbf{x}_k + \mathbf{\Gamma}_0\mathbf{u}_k + \mathbf{\Gamma}_1\mathbf{u}_{k+1}, \tag{6}$$

with the state $\mathbf{x}_k$ and the input $\mathbf{u}_k = \mathbf{j}_k$.

*4.1.3. Robot kinematics.* The forward kinematics of the robot are mappings from the joint space to the Cartesian space and describe the Cartesian pose of the robot's end-effector. Given the joint positions $\mathbf{q}$ of the robot, the Cartesian position map for the forward kinematics reads as $\mathbf{p}_\mathrm{c}(\mathbf{q}) : \mathbb{R}^n \to \mathbb{R}^3$ and for the orientation $\mathbf{R}_\mathrm{c}(\mathbf{q}) : \mathbb{R}^n \to \mathbb{R}^{3\times3}$, with the rotation matrix $\mathbf{R}_\mathrm{c}$.

By using the geometric Jacobian $\mathbf{J}(\mathbf{q}) \in \mathbb{R}^{6\times n}$, the Cartesian velocity $\mathbf{v}_\mathrm{c}$ and the angular velocity $\boldsymbol{\omega}_\mathrm{c}$ of the end-effector are related to the joint velocity $\dot{\mathbf{q}}$ by

$$\begin{bmatrix} \mathbf{v}_\mathrm{c} \\ \boldsymbol{\omega}_\mathrm{c} \end{bmatrix} = \mathbf{J}(\mathbf{q})\dot{\mathbf{q}}, \tag{7}$$

which yields the Cartesian and angular accelerations as

$$\begin{bmatrix} \mathbf{a}_\mathrm{c} \\ \dot{\boldsymbol{\omega}}_\mathrm{c} \end{bmatrix} = \frac{\mathrm{d}}{\mathrm{d}t}\begin{bmatrix} \mathbf{v}_\mathrm{c} \\ \boldsymbol{\omega}_\mathrm{c} \end{bmatrix} = \mathbf{J}(\mathbf{q})\ddot{\mathbf{q}} + \dot{\mathbf{J}}(\mathbf{q})\dot{\mathbf{q}}. \tag{8}$$

If $n > 6$, the robot is kinematically redundant and the Jacobian $\mathbf{J}(\mathbf{q})$ becomes a non-square matrix. Redundant robots, such as the KUKA LBR iiwa 14 R820 used in this work, have a joint nullspace, which has to be considered for planning and control.

## 4.2. Orientation representation

The orientation of the end-effector in the Cartesian space can be represented in different ways, such as rotation matrices, quaternions, and Euler angles. In this work, the Lie algebra

$$\boldsymbol{\tau} = \theta\mathbf{u} \tag{9}$$

of the rotation matrix $\mathbf{R}$ is used. An orientation is thus specified by an angle $\theta$ around the unit length axis $\mathbf{u}$. The most relevant results of the Lie theory for rotations, utilized in this work, are summarized from Solà et al. (2021) in the following.

*4.2.1. Mappings.* To transform a rotation matrix $\mathbf{R}$ into its equivalent (non-unique) element $\boldsymbol{\tau}$ in the Lie space, the exponential and logarithmic mappings

$$\mathrm{Exp}: \quad \mathbb{R}^3 \to \mathbb{R}^{3\times3} : \boldsymbol{\tau} \to \mathbf{R} = \mathrm{Exp}(\boldsymbol{\tau}) \tag{10}$$

$$\mathrm{Log}: \quad \mathbb{R}^{3\times3} \to \mathbb{R}^3 : \mathbf{R} \to \boldsymbol{\tau} = \mathrm{Log}(\mathbf{R}) \tag{11}$$

are used, with the definitions

$$\mathrm{Exp}(\theta\mathbf{u}) = \mathbf{I} + \sin(\theta)[\mathbf{u}]_\times + (1 - \cos(\theta))[\mathbf{u}]_\times^2 \tag{12}$$

$$[\mathrm{Log}(\mathbf{R})]_\times = \frac{\phi(\mathbf{R} - \mathbf{R}^\mathrm{T})}{2\sin(\phi)}, \tag{13}$$

the identity matrix $\mathbf{I}$, and $\phi = \cos^{-1}\left(\frac{\mathrm{trace}(\mathbf{R})-1}{2}\right)$. The operator

$$[\mathbf{u}]_\times = \left[\begin{bmatrix} u_x \\ u_y \\ u_z \end{bmatrix}\right]_\times = \begin{bmatrix} 0 & -u_z & u_y \\ u_z & 0 & -u_x \\ -u_y & u_x & 0 \end{bmatrix} \tag{14}$$

converts a vector into a skew-symmetric matrix.

*4.2.2. Concatenation of rotations.* A rotation matrix may result from the concatenation of multiple rotation matrices. An example is the Euler angles which are the result of three consecutive rotations around distinct rotation axes. Concretely, let us assume that

$$\mathbf{R} = \mathbf{R}_3\mathbf{R}_2\mathbf{R}_1 = \mathrm{Exp}(\boldsymbol{\tau}_3)\mathrm{Exp}(\boldsymbol{\tau}_2)\mathrm{Exp}(\boldsymbol{\tau}_1) = \mathrm{Exp}(\boldsymbol{\tau}) \tag{15}$$

is the combination of the rotation matrices $\mathbf{R}_1$, $\mathbf{R}_2$, and $\mathbf{R}_3$. By using the Lie theory for rotations, (15) can be approximated around $\mathbf{R}_2$ as (Solà et al., 2021)

$$\begin{aligned} \boldsymbol{\tau} &= \mathrm{Log}(\mathbf{R}_3\mathbf{R}_2\mathbf{R}_1) \\ &\approx \mathrm{Log}(\mathbf{R}_3\mathbf{R}_2) + \mathbf{J}_\mathrm{r}^{-1}(\mathrm{Log}(\mathbf{R}_3\mathbf{R}_2))\boldsymbol{\tau}_1 \\ &\approx \boldsymbol{\tau}_2 + \mathbf{J}_\mathrm{l}^{-1}(\boldsymbol{\tau}_2)\boldsymbol{\tau}_3 + \mathbf{J}_\mathrm{r}^{-1}(\mathrm{Log}(\mathbf{R}_3\mathbf{R}_2))\boldsymbol{\tau}_1, \end{aligned} \tag{16}$$

where the left and right inverse Jacobians

$$\mathbf{J}_\mathrm{l}^{-1}(\boldsymbol{\tau}) = \mathbf{I} - \frac{1}{2}[\boldsymbol{\tau}]_\times + \left(\frac{1}{\theta^2} - \frac{1+\cos\theta}{2\theta\sin\theta}\right)[\boldsymbol{\tau}]_\times^2 \tag{17}$$

$$\mathbf{J}_\mathrm{r}^{-1}(\boldsymbol{\tau}) = \mathbf{I} + \frac{1}{2}[\boldsymbol{\tau}]_\times + \left(\frac{1}{\theta^2} - \frac{1+\cos\theta}{2\theta\sin\theta}\right)[\boldsymbol{\tau}]_\times^2 \tag{18}$$

are defined using $\theta = \|\boldsymbol{\tau}\|_2$. Approximations around $\mathbf{R}_1$ and $\mathbf{R}_3$ are obtained analogously to (16). Note that for transposed rotation matrices the additions in (16) become subtractions.

## 4.3. Reference path formulation

In this work, each path point is given as a Cartesian pose, which motivates a separation of the position and orientation path as $\boldsymbol{\pi}_\mathrm{p}(\phi) \in \mathbb{R}^3$ and $\boldsymbol{\pi}_\mathrm{o}(\phi) \in \mathbb{R}^3$, respectively, with the same path parameter $\phi$. The position reference path $\boldsymbol{\pi}_\mathrm{p}(\phi)$ is assumed to be arc-length parametrized. The arc-length parametrization implies $\|\frac{\partial}{\partial\phi}\boldsymbol{\pi}_\mathrm{p}\|_2 = \|\boldsymbol{\pi}_\mathrm{p}'\|_2 = 1, \forall\phi \in [0, \phi_\mathrm{f}]$, where $\phi_\mathrm{f}$ is the total arc length of the path. The orientation path uses the same path parameter $\phi$ and is therefore, in general, not arc-length parametrized. To couple the two paths, the orientation path $\boldsymbol{\pi}_\mathrm{o}(\phi)$ is parametrized such that the via-points coincide with the position path $\boldsymbol{\pi}_\mathrm{p}(\phi)$ and the orientation path also ends at $\phi_\mathrm{f}$. Hence, the orientation path derivative is in general not normalized, which will become important later.

## 4.4. Path error

In this section, the position and orientation path errors are computed and decomposed based on the tangential path directions $\boldsymbol{\pi}_\mathrm{p}' = \frac{\partial}{\partial\phi}\boldsymbol{\pi}_\mathrm{p}$ and $\boldsymbol{\pi}_\mathrm{o}' = \frac{\partial}{\partial\phi}\boldsymbol{\pi}_\mathrm{o}$. First, the position path error and its decomposition are described. Second, the orientation path error is decomposed similarly using the formulation from Section 4.2.

*4.4.1. Position path error.* The position path error between the current end-effector position $\mathbf{p}_c(\mathbf{q})$ and the reference path in the Cartesian space

$$\mathbf{e}_p(\mathbf{q}, \phi(t)) = \mathbf{p}_c(\mathbf{q}) - \boldsymbol{\pi}_p(\phi(t)) \tag{19}$$

is a function of the joint positions $\mathbf{q}$ and the path parameter $\phi(t)$ using the forward kinematics $\mathbf{p}_c(\mathbf{q})$ of the robot.

*4.4.2. Position path error decomposition.* The position path error (19) is decomposed into a tangential and an orthogonal error. Figure 2 shows a visualization of the geometric relations. The tangential part

$$\mathbf{e}_p^{\parallel}(\mathbf{q}, \phi(t)) = \left(\boldsymbol{\pi}_p'(\phi(t))^{\mathrm{T}} \mathbf{e}_p(\mathbf{q}, \phi(t))\right) \boldsymbol{\pi}_p'(\phi(t)) \tag{20}$$

is the projection onto the tangent $\boldsymbol{\pi}_p'(\phi(t))$ of the reference path. Thus, the remaining error is the orthogonal path error, reading as

$$\mathbf{e}_p^{\perp}(\mathbf{q}, \phi(t)) = \mathbf{e}_p(\mathbf{q}, \phi(t)) - \mathbf{e}_p^{\parallel}(\mathbf{q}, \phi(t)). \tag{21}$$

The time derivatives of (20) and (21) yield

$$\dot{\mathbf{e}}_p = \mathbf{v}_c(\mathbf{q}) - \boldsymbol{\pi}_p'(\phi(t))\dot{\phi}(t) \tag{22a}$$

$$\ddot{\mathbf{e}}_p = \mathbf{a}_c(\mathbf{q}) - \boldsymbol{\pi}_p''(\phi(t))\dot{\phi}^2(t) - \boldsymbol{\pi}_p'(\phi(t))\ddot{\phi}(t) \tag{22b}$$

$$\begin{aligned}\dot{\mathbf{e}}_p^{\parallel} = &\left(\left(\boldsymbol{\pi}_p''(\phi(t))\dot{\phi}(t)\right)^{\mathrm{T}} \mathbf{e}_p(\mathbf{q}, \phi(t))\right)\boldsymbol{\pi}_p'(\phi(t)) \\ &+ \left(\boldsymbol{\pi}_p'(\phi(t))^{\mathrm{T}} \dot{\mathbf{e}}_p(\mathbf{q}, \phi(t))\right)\boldsymbol{\pi}_p'(\phi(t)) \\ &+ \left(\boldsymbol{\pi}_p'(\phi(t))^{\mathrm{T}} \mathbf{e}_p(\mathbf{q}, \phi(t))\right)\boldsymbol{\pi}_p''(\phi(t))\dot{\phi}(t)\end{aligned} \tag{22c}$$

$$\dot{\mathbf{e}}_p^{\perp} = \dot{\mathbf{e}}_p(\mathbf{q}, \phi(t)) - \dot{\mathbf{e}}_p^{\parallel}(\mathbf{q}, \phi(t)). \tag{22d}$$

In the following, the function arguments will be omitted for clarity of presentation. The orthogonal path error $\mathbf{e}_p^{\perp}$ from (21) is further decomposed using the orthonormal basis vectors $\mathbf{b}_{p,1}$ and $\mathbf{b}_{p,2}$ of the orthogonal error plane. The orthogonal position path errors $\mathbf{e}_p^{\perp,1}$, $\mathbf{e}_p^{\perp,2}$ are then obtained by projection onto these basis vectors as

$$\mathbf{e}_{p,\mathrm{proj}}^{\perp} = \begin{bmatrix} e_{p,\mathrm{proj},1}^{\perp} \\ e_{p,\mathrm{proj},2}^{\perp} \end{bmatrix} = \begin{bmatrix} \mathbf{b}_{p,1}^{\mathrm{T}} \\ \mathbf{b}_{p,2}^{\mathrm{T}} \end{bmatrix} \mathbf{e}_p^{\perp} \tag{23a}$$

$$\mathbf{e}_p^{\perp,i} = e_{p,\mathrm{proj},i}^{\perp} \mathbf{b}_{p,i}, i = 1, 2. \tag{23b}$$

The basis vectors $\mathbf{b}_{p,1}$ and $\mathbf{b}_{p,2}$ are obtained using the Gram–Schmidt procedure by providing a desired basis vector $\mathbf{b}_{p,d}$ and using the reference velocity vector $\mathbf{v}_r = \boldsymbol{\pi}_p'(\phi(t))$, leading to

$$\mathbf{w}(\mathbf{v}_r, \mathbf{b}_{p,d}) = \left(\frac{\mathbf{v}_r^{\mathrm{T}}}{\|\mathbf{v}_r\|_2}\mathbf{b}_{p,d}\right)\frac{\mathbf{v}_r}{\|\mathbf{v}_r\|_2} \tag{24a}$$

$$\mathbf{u}(\mathbf{v}_r, \mathbf{b}_{p,d}) = \mathbf{b}_{p,d} - \mathbf{w}(\mathbf{v}_r, \mathbf{b}_{p,d}) \tag{24b}$$

$$\mathbf{b}_{p,1} = \frac{\mathbf{u}(\mathbf{v}_r, \mathbf{b}_{p,d})}{\|\mathbf{u}(\mathbf{v}_r, \mathbf{b}_{p,d})\|_2} \tag{24c}$$

as the first normalized basis vector. The cross product $\mathbf{b}_{p,2} = \mathbf{v}_r \times \mathbf{b}_{p,1}$ gives the second basis vector. Note that $\mathbf{b}_{p,d}$ and $\mathbf{v}_r$ must be linearly independent.

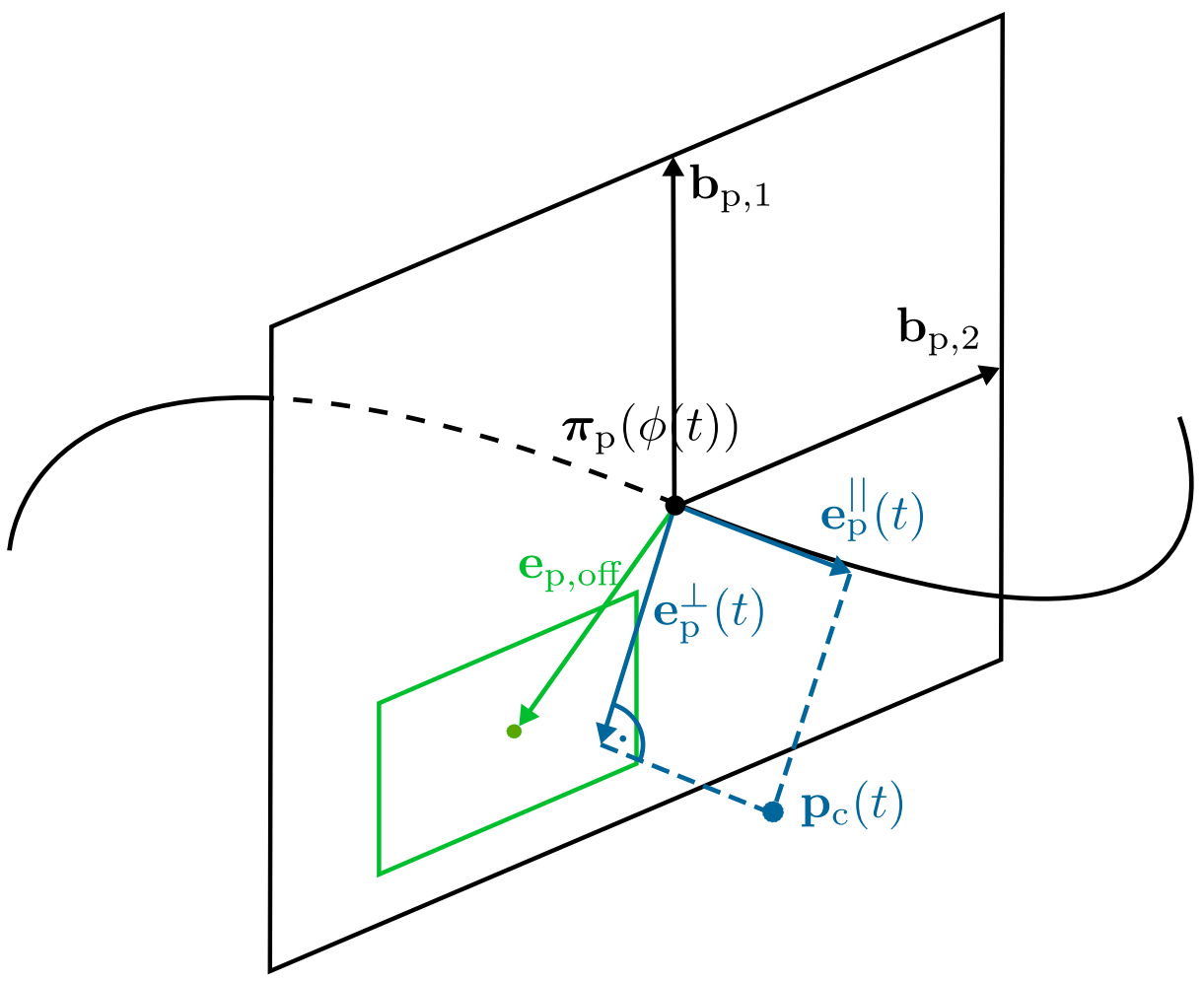

**Figure 2.** Position path error decomposition into the orthogonal and tangential path direction. The black rectangle visualizes the orthogonal error plane spanned by the basis vectors $\mathbf{b}_{p,1}$ and $\mathbf{b}_{p,2}$. The current reference path position is $\boldsymbol{\pi}_p(\phi(t))$. The blue lines indicate the errors of the current end-effector position $\mathbf{p}_c(t)$. The error bounds are indicated by the green rectangle, which is offset from the path by $\mathbf{e}_{p,\mathrm{off}}$.

*4.4.3. Orientation path error.* In the following, the orientation path errors are derived using rotation matrices. Then the errors are approximated by applying the Lie theory for rotations from Section 4.2. The orientation path error between the planned and the reference path

$$\mathbf{e}_o(t) = \mathrm{Log}\left(\mathbf{R}_c(t)\mathbf{R}_r^{\mathrm{T}}(\phi(t))\right) \tag{25}$$

is a function of the time $t$, with the rotation matrix representation of the orientation reference path $\mathbf{R}_r(\phi(t)) = \mathrm{Exp}(\boldsymbol{\pi}_o(\phi(t)))$ and the current state rotation matrix $\mathbf{R}_c(t) = \mathbf{R}_c(\mathbf{q}(t))$. These rotation matrices are related to the respective angular velocities $\boldsymbol{\omega}_c$ and $\boldsymbol{\omega}_r$ in the form

$$\dot{\mathbf{R}}_c(t) = [\boldsymbol{\omega}_c(t)]_\times \mathbf{R}_c(t) \tag{26a}$$

$$\frac{\partial \mathbf{R}_r(\phi)}{\partial \phi} = [\boldsymbol{\omega}_r(\phi(t))]_\times \mathbf{R}_r(\phi(t)), \tag{26b}$$

where the reference path depends on the path parameter $\phi(t)$. For constant angular velocities $\boldsymbol{\omega}_c$ and $\boldsymbol{\omega}_r$, the closed-form solutions for (26) over a time span $\Delta t$ exist in the form

$$\mathbf{R}_c(t+\Delta t) = \mathrm{Exp}(\boldsymbol{\omega}_c \Delta t)\mathbf{R}_c(t) \tag{27a}$$

$$\mathbf{R}_r(\phi(t)+\Delta\phi) = \mathrm{Exp}(\boldsymbol{\omega}_r \Delta\phi)\mathbf{R}_r(\phi(t)), \tag{27b}$$

with $\Delta\phi = \phi(t+\Delta t) - \phi(t)$. Using (16) and $\Delta\phi \approx \dot{\phi}(t)\Delta t$, the orientation path error $\mathbf{e}_o(t)$ propagates as

$$\begin{aligned}
\mathbf{e}_o&(t+\Delta t)\\
&= \mathrm{Log}(\mathrm{Exp}(\boldsymbol{\omega}_c\Delta t)\mathbf{R}_c(t)\mathbf{R}_r^{\mathrm{T}}(\phi(t))\mathrm{Exp}(\boldsymbol{\omega}_r\Delta\phi)^{\mathrm{T}})\\
&\approx \mathbf{e}_o(t) + \Delta t\big[\mathbf{J}_l^{-1}(\mathbf{e}_o(t))\boldsymbol{\omega}_c\\
&\quad -\mathbf{J}_r^{-1}(\mathrm{Log}(\mathrm{Exp}(\boldsymbol{\omega}_c\Delta t)\mathbf{R}_c(t)\mathbf{R}_r^{\mathrm{T}}(\phi(t))))\boldsymbol{\omega}_r\dot{\phi}(t)\big]\\
&\approx \mathbf{e}_o(t) + \Delta t\big[\mathbf{J}_l^{-1}(\mathbf{e}_o(t))\boldsymbol{\omega}_c - \mathbf{J}_r^{-1}(\mathbf{e}_o(t))\boldsymbol{\omega}_r\dot{\phi}(t)\big],
\end{aligned} \tag{28}$$

where the last step approximates the right Jacobian around the time $t$. For $\Delta t \to 0$, (28) can be written as

$$\dot{\mathbf{e}}_o(t) = \mathbf{J}_l^{-1}(\mathbf{e}_o(t))\boldsymbol{\omega}_c(t) - \mathbf{J}_r^{-1}(\mathbf{e}_o(t))\boldsymbol{\omega}_r(\phi(t))\dot{\phi}(t), \tag{29}$$

which is integrated to obtain the orientation path error. Figure 3 shows a comparison of this approximation with the true error (25).

*4.4.4. Orientation path error decomposition.* The orientation path error is decomposed similarly to the position path error. The rotation path error matrix $\mathbf{R}^e = \mathrm{Exp}(\mathbf{e}_o)$ consists of the orthogonal path errors $\mathbf{R}^{\perp,1}$ and $\mathbf{R}^{\perp,2}$ as well as the tangential path error $\mathbf{R}^{\parallel}$, given by

$$\mathbf{R}^e = \mathbf{R}^{\perp,2}\mathbf{R}^{\parallel}\mathbf{R}^{\perp,1} = \mathbf{R}_c\mathbf{R}_r^{\mathrm{T}}, \tag{30}$$

with

$$\mathbf{R}^{\perp,1} = \mathrm{Exp}(\mathbf{e}_o^{\perp,1}) = \mathrm{Exp}(\alpha\mathbf{b}_{o,1}) \tag{31a}$$

$$\mathbf{R}^{\parallel} = \mathrm{Exp}(\mathbf{e}_o^{\parallel}) = \mathrm{Exp}(\beta\boldsymbol{\omega}_r) \tag{31b}$$

$$\mathbf{R}^{\perp,2} = \mathrm{Exp}(\mathbf{e}_o^{\perp,2}) = \mathrm{Exp}(\gamma\mathbf{b}_{o,2}) \tag{31c}$$

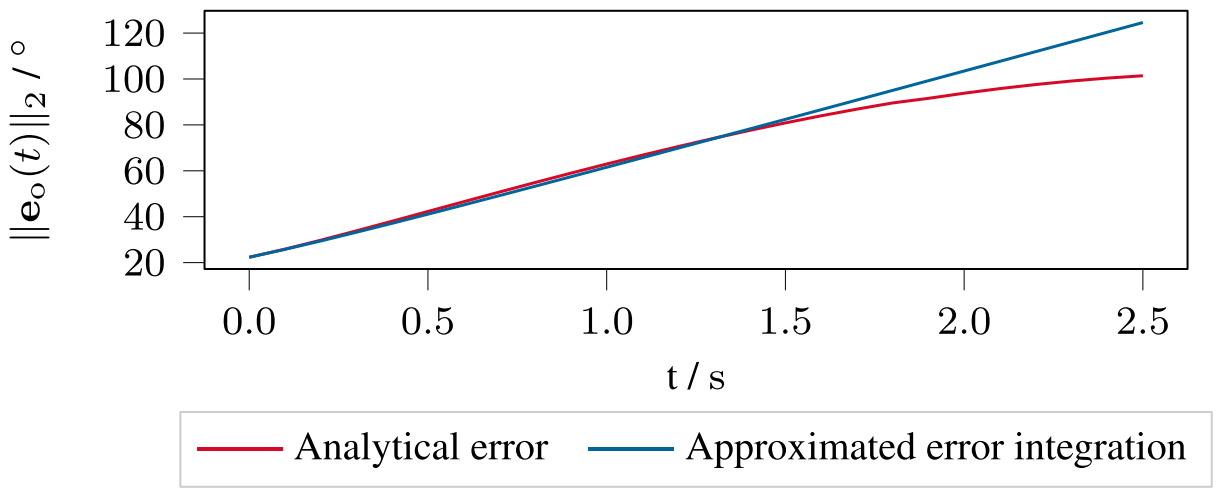

**Figure 3.** Comparison of the orientation path error computations. The reference and the current orientation path are computed over a time interval of 2.5 s using the constant angular velocities $\boldsymbol{\omega}_r$ and $\boldsymbol{\omega}_c$. The true evolution of the norm of (25) in red is compared to the approximate computation of the norm according to (29) in blue.

and the orthonormal orientation axes $\mathbf{b}_{o,1}$, $\mathbf{b}_{o,2}$, and $\boldsymbol{\omega}_r$. The basis vectors $\mathbf{b}_{o,1}$ and $\mathbf{b}_{o,2}$ are obtained by following the procedure of the position path error in (24) with the angular velocity $\boldsymbol{\omega}_r$ instead of the Cartesian velocity $\mathbf{v}_r$ and with the desired basis vector $\mathbf{b}_{o,d}$. In this work, the projection of the error rotation matrix $\mathbf{R}^e$ onto the orientation axes $\mathbf{b}_{o,1}$, $\mathbf{b}_{o,2}$, and $\boldsymbol{\omega}_r$ is defined as

$$\underset{\alpha,\beta,\gamma}{\arg\min}\|\mathrm{Log}(\mathbf{R}^e(\mathbf{R}^{\perp,1})^{\mathrm{T}}(\mathbf{R}^{\parallel})^{\mathrm{T}}(\mathbf{R}^{\perp,2})^{\mathrm{T}})\|_2, \tag{32}$$

which gives the optimal values $\alpha^*$, $\beta^*$, and $\gamma^*$ that minimize the deviation from $\mathbf{R}^e$. Note that this formulation is equivalent to roll, pitch, and yaw (RPY) angles, when $\mathbf{b}_{o,1}$, $\mathbf{b}_{o,2}$, and $\boldsymbol{\omega}_r$ are the $z$, $x$, and $y$ axes, respectively (Ang and Tourassis, 1987). Thus, solving (32) efficiently is done by computing the RPY angles for

$$\mathbf{R}_{\mathrm{RPY}} = (\mathbf{R}_{\mathrm{RPY}}^e)^{\mathrm{T}}\mathbf{R}^e\mathbf{R}_{\mathrm{RPY}}^e, \tag{33}$$

with

$$\mathbf{R}_{\mathrm{RPY}}^e = [\,\mathbf{b}_{o,2}|\boldsymbol{\omega}_r|\mathbf{b}_{o,1}\,] \tag{34}$$

as the rotation matrix between the axes for the RPY angles and the desired axes for the error computation. The RPY representation has singularities when $\beta$ becomes 90° or 270° (Ang and Tourassis, 1987). In this work, the tangential orientation path error $\mathbf{R}^{\parallel}$ is minimized, as will be shown in Section 4.5. Thus, the singularities are avoided by letting $\mathbf{R}^{\parallel}$ represent the pitch angle $\beta$ and minimizing it, which also explains the order of the matrix multiplication in (30).

Using the Lie representation and the approximation (16), the optimization problem (32) can be written as

$$\begin{aligned}
\underset{\alpha,\beta,\gamma}{\arg\min}\|\mathbf{e}_o &- \gamma\mathbf{J}_r^{-1}(\mathrm{Log}(\mathbf{R}^e(\mathbf{R}^{\perp,1})^{\mathrm{T}}(\mathbf{R}^{\parallel})^{\mathrm{T}}))\mathbf{b}_{o,2}\\
&-\beta\mathbf{J}_r^{-1}(\mathrm{Log}(\mathbf{R}^e(\mathbf{R}^{\perp,1})^{\mathrm{T}}))\boldsymbol{\omega}_r\\
&-\alpha\mathbf{J}_r^{-1}(\mathrm{Log}(\mathbf{R}^e))\mathbf{b}_{o,1}\|_2.
\end{aligned} \tag{35}$$

The Jacobians are not independent of the optimization variables $\alpha$, $\beta$, and $\gamma$, which makes solving (35) a challenging task. To further simplify the problem it is assumed for the Jacobians that the optimal values $\alpha^*$, $\beta^*$, and $\gamma^*$ are close to an initial guess $\alpha^0$, $\beta^0$, and $\gamma^0$. This motivates to compute the Jacobians only for the initial guess to obtain the vectors

$$\mathbf{r}_1 = \mathbf{J}_r^{-1}(\mathrm{Log}(\mathbf{R}^e))\mathbf{b}_{o,1} \tag{36a}$$

$$\mathbf{r}_2 = \mathbf{J}_r^{-1}(\mathrm{Log}(\mathbf{R}^e(\mathbf{R}^{\perp,1}(\alpha^0))^{\mathrm{T}}))\boldsymbol{\omega}_r \tag{36b}$$

$$\mathbf{r}_3 = \mathbf{J}_r^{-1}(\mathrm{Log}(\mathbf{R}^e(\mathbf{R}^{\perp,1}(\alpha^0))^{\mathrm{T}}(\mathbf{R}^{\parallel}(\beta^0))^{\mathrm{T}}))\mathbf{b}_{o,2}, \tag{36c}$$

which are constant in the optimization problem (35). Using (36) in (35) results in a quadratic program, and the optimal solution $\alpha^*$, $\beta^*$, and $\gamma^*$ can be obtained from

$$\begin{bmatrix} \mathbf{r}_1^T\mathbf{r}_1 & \mathbf{r}_1^T\mathbf{r}_2 & \mathbf{r}_1^T\mathbf{r}_3 \\ \mathbf{r}_2^T\mathbf{r}_1 & \mathbf{r}_2^T\mathbf{r}_2 & \mathbf{r}_2^T\mathbf{r}_3 \\ \mathbf{r}_3^T\mathbf{r}_1 & \mathbf{r}_3^T\mathbf{r}_2 & \mathbf{r}_3^T\mathbf{r}_3 \end{bmatrix} \begin{bmatrix} \alpha \\ \beta \\ \gamma \end{bmatrix} = \begin{bmatrix} \mathbf{e}_o^T\mathbf{r}_1 \\ \mathbf{e}_o^T\mathbf{r}_2 \\ \mathbf{e}_o^T\mathbf{r}_3 \end{bmatrix} \tag{37}$$

yielding

$$\alpha^* \approx \boldsymbol{\rho}_\alpha^T(\mathbf{b}_{o,1}, \mathbf{b}_{o,2}, \boldsymbol{\omega}_r)\mathbf{e}_o \tag{38a}$$

$$\beta^* \approx \boldsymbol{\rho}_\beta^T(\mathbf{b}_{o,1}, \mathbf{b}_{o,2}, \boldsymbol{\omega}_r)\mathbf{e}_o \tag{38b}$$

$$\gamma^* \approx \boldsymbol{\rho}_\gamma^T(\mathbf{b}_{o,1}, \mathbf{b}_{o,2}, \boldsymbol{\omega}_r)\mathbf{e}_o \tag{38c}$$

as the approximate optimal solution of (32) with the solution vectors $\boldsymbol{\rho}_\alpha$, $\boldsymbol{\rho}_\beta$, and $\boldsymbol{\rho}_\gamma$, which are not given explicitly here.

Equation (38) provides the approximate solution to (32) at one point in time but the vectors $\mathbf{b}_{o,1}$, $\mathbf{b}_{o,2}$, and $\boldsymbol{\omega}_r$ change along the path. It is possible to use the approximate solution (38) to compute the optimal values for $\alpha^*$, $\beta^*$, and $\gamma^*$ at each time step when the robot's end-effector traverses the path. However, the quality of the approximation in (38) decreases when $\alpha$, $\beta$, and $\gamma$ grow. For the model predictive control formulation in Section 4.5, the orientation path errors $\mathbf{e}_o^{\perp,1}(t)$, $\mathbf{e}_o^{\perp,2}(t)$, and $\mathbf{e}_o^{\parallel}(t)$ are computed over a time horizon spanning from $t_0$ to $t_1$. This motivates to obtain an initial guess $\alpha^0$, $\beta^0$, and $\gamma^0$ using (32) at the time $t_0$ and then use the change of the approximate solution (38) to propagate it over the time horizon. Thus, the time evolution of the errors are computed as

$$\mathbf{e}_o^{\perp,1}(t) = \mathbf{e}_o^{\perp,1}(t_0) + \int_{t_0}^{t} \dot{\mathbf{e}}_o^{\perp,1}(\tau)\mathrm{d}\tau \tag{39a}$$

$$\mathbf{e}_o^{\perp,2}(t) = \mathbf{e}_o^{\perp,2}(t_0) + \int_{t_0}^{t} \dot{\mathbf{e}}_o^{\perp,2}(\tau)\mathrm{d}\tau \tag{39b}$$

$$\mathbf{e}_o^{\parallel}(t) = \mathbf{e}_o^{\parallel}(t_0) + \int_{t_0}^{t} \dot{\mathbf{e}}_o^{\parallel}(\tau)\mathrm{d}\tau, \tag{39c}$$

where the initial guesses for $\alpha^0$, $\gamma^0$, and $\beta^0$ are used to compute the initial orientation path errors $\mathbf{e}_o^{\perp,1}(t_0)$, $\mathbf{e}_o^{\perp,2}(t_0)$, and $\mathbf{e}_o^{\parallel}(t_0)$, respectively, meaning that (39) is an approximation around the initial solution at time $t_0$. The time derivatives $\dot{\mathbf{e}}_o^{\perp,1}$, $\dot{\mathbf{e}}_o^{\perp,2}$, and $\dot{\mathbf{e}}_o^{\parallel}$ required in (39) are computed using (38)

$$\begin{aligned} \dot{\mathbf{e}}_o^{\parallel} &= (\dot{\boldsymbol{\rho}}_\beta^T(\mathbf{b}_{o,1}, \mathbf{b}_{o,2}, \boldsymbol{\omega}_r)\mathbf{e}_o(t))\boldsymbol{\omega}_r(\phi(t)) \\ &\quad + (\boldsymbol{\rho}_\beta^T(\mathbf{b}_{o,1}, \mathbf{b}_{o,2}, \boldsymbol{\omega}_r)\dot{\mathbf{e}}_o(t))\boldsymbol{\omega}_r(\phi(t)) \\ &\quad + (\boldsymbol{\rho}_\beta^T(\mathbf{b}_{o,1}, \mathbf{b}_{o,2}, \boldsymbol{\omega}_r)\mathbf{e}_o(t))\boldsymbol{\omega}_r'(\phi(t))\dot{\phi}(t) \\ &= \dot{\mathbf{e}}_{o,\mathrm{proj}}(\boldsymbol{\rho}_\beta(\mathbf{b}_{o,1}, \mathbf{b}_{o,2}, \boldsymbol{\omega}_r), \boldsymbol{\omega}_r) \end{aligned} \tag{40a}$$

$$\dot{\mathbf{e}}_o^{\perp,1} = \dot{\mathbf{e}}_{o,\mathrm{proj}}(\boldsymbol{\rho}_\alpha(\mathbf{b}_{o,1}, \mathbf{b}_{o,2}, \boldsymbol{\omega}_r), \mathbf{b}_{o,1}) \tag{40b}$$

$$\dot{\mathbf{e}}_o^{\perp,2} = \dot{\mathbf{e}}_{o,\mathrm{proj}}(\boldsymbol{\rho}_\gamma(\mathbf{b}_{o,1}, \mathbf{b}_{o,2}, \boldsymbol{\omega}_r), \mathbf{b}_{o,2}). \tag{40c}$$

Thus, the approximation error of (39) is limited to the change of the orientation path errors within the time horizon of the MPC.

The orthogonal orientation path errors $\mathbf{e}_o^{\perp,1}$ and $\mathbf{e}_o^{\perp,2}$ are represented similar to Section 4.4.2 by

$$\mathbf{e}_{o,\mathrm{proj}}^{\perp} = \begin{bmatrix} e_{o,\mathrm{proj},1}^{\perp} \\ e_{o,\mathrm{proj},2}^{\perp} \end{bmatrix} = \begin{bmatrix} \alpha \\ \gamma \end{bmatrix}, \tag{41}$$

in the orthogonal error plane.

**Remark.** Due to the iterative computation of the orientation path errors in (39), the reference path $\boldsymbol{\pi}_o(\phi(t))$ is only evaluated at the initial time $t_0$. For all other time points in the time horizon from $t_0$ to $t_1$, only the reference angular velocity $\boldsymbol{\omega}_r(\phi(t))$ and its derivative $\boldsymbol{\omega}_r'(\phi(t))$ with respect to the path parameter $\phi$ are needed.

## 4.5. Optimization problem

In this section, the results of the previous sections are utilized to formulate the optimal control problem for the proposed MPC framework. The goal is to follow a desired reference path of the Cartesian end-effector's position and orientation within given asymmetric error bounds in the error plane orthogonal to the path, which also includes compliance with desired via-points. At the same time, the system dynamics of the controlled robot (6) and the state and input constraints must be satisfied. To govern the progress along the path, the discrete-time path-parameter dynamics

$$\boldsymbol{\xi}_{k+1} = \boldsymbol{\Phi}_\xi \boldsymbol{\xi}_k + \boldsymbol{\Gamma}_{0,\xi} v_k + \boldsymbol{\Gamma}_{1,\xi} v_{k+1} \tag{42}$$

are introduced, with the state $\boldsymbol{\xi}_k^T = [\phi_k, \dot{\phi}_k, \ddot{\phi}_k]$ and the input $v_k = \dddot{\phi}_k$. The linear system (42) is chosen to have the same order as (6) and also uses hat functions for the interpolation of the jerk input $\dddot{\phi}_k$. Thus, the optimization problem for the MPC is formulated as

$$\min_{\substack{\mathbf{u}_1,\ldots,\mathbf{u}_N, \\ \mathbf{x}_1,\ldots,\mathbf{x}_N, \\ \boldsymbol{\xi}_1,\ldots,\boldsymbol{\xi}_N, \\ v_1,\ldots,v_N}} \sum_{i=1}^{N} l(\mathbf{x}_i, \boldsymbol{\xi}_i, \mathbf{u}_i, v_i) \tag{43a}$$

$$\text{s.t. } \mathbf{x}_{i+1} = \boldsymbol{\Phi}\mathbf{x}_i + \boldsymbol{\Gamma}_0\mathbf{u}_i + \boldsymbol{\Gamma}_1\mathbf{u}_{i+1}, \quad i = 0, \ldots, N-1 \tag{43b}$$

$$\boldsymbol{\xi}_{i+1} = \boldsymbol{\Phi}_\xi\boldsymbol{\xi}_i + \boldsymbol{\Gamma}_{0,\xi}v_i + \boldsymbol{\Gamma}_{1,\xi}v_{i+1}, \quad i = 0, \ldots, N-1 \tag{43c}$$

$$\mathbf{x}_0 = \mathbf{x}_{\mathrm{init}} \tag{43d}$$

$$\mathbf{u}_0 = \mathbf{u}_{\mathrm{init}} \tag{43e}$$

$$\boldsymbol{\xi}_0 = \boldsymbol{\xi}_{\mathrm{init}} \tag{43f}$$

$$v_0 = v_{\mathrm{init}} \tag{43g}$$

$$\underline{\mathbf{x}} \le \mathbf{x}_{i+1} \le \overline{\mathbf{x}} \tag{43h}$$

$$\underline{\mathbf{u}} \le \mathbf{u}_{i+1} \le \overline{\mathbf{u}} \tag{43i}$$

$$\phi_{i+1} \le \phi_{\mathrm{f}} \tag{43j}$$

$$\dot{\phi}_{i+1} \le \overline{\dot{\phi}} \tag{43k}$$

$$\psi_{\mathrm{p},m}(\mathbf{e}^{\perp}_{\mathrm{p,proj},i+1}) \le 0, \quad m = 1,2 \tag{43l}$$

$$\psi_{\mathrm{o},m}(\mathbf{e}^{\perp}_{\mathrm{o,proj},i+1}) \le 0, \quad m = 1,2. \tag{43m}$$

The quantities $\mathbf{x}_{\mathrm{init}}$, $\mathbf{u}_{\mathrm{init}}$, $\xi_{\mathrm{init}}$, and $v_{\mathrm{init}}$ in (43d)–(43g) describe the initial conditions at the initial time $t_0$. In (43h) and (43i), the system states $\mathbf{x}$ and the inputs $\mathbf{u}$ are bounded from below and above with the corresponding limits $\underline{\mathbf{x}}$, $\overline{\mathbf{x}}$, $\underline{\mathbf{u}}$, and $\overline{\mathbf{u}}$. The path parameter never exceeds the maximum value $\phi_{\mathrm{f}}$ (see (43j)), and its speed is bounded from above by $\overline{\dot{\phi}}$ (see (43k)). In (43l) and (43m), the Cartesian orthogonal position and orientation path errors are bounded by the constraint functions $\psi_{\mathrm{p},m}$ and $\psi_{\mathrm{o},m}$ in both desired basis vector directions, which is further detailed in Section 5. The position reference path is given by

$$\boldsymbol{\pi}_{\mathrm{p},i} = \boldsymbol{\pi}_{\mathrm{p}}(\phi_i) \tag{44}$$

for each discrete-time instance $t_0 + iT_{\mathrm{s}}$. An explicit representation of $\boldsymbol{\pi}_{\mathrm{o}}(\phi(t))$ is not required within the optimization problem (43), as discussed in Section 4.4.4. The objective function (43a),

$$\begin{aligned} l(\mathbf{x}_i, \xi_i, \mathbf{u}_i, v_i) = {} & w_{\parallel,\mathrm{p}}\|\mathbf{e}^{\parallel}_{\mathrm{p},i}\|_2^2 + w_{\parallel,\mathrm{o}}\|\mathbf{e}^{\parallel}_{\mathrm{o},i}\|_2^2 \\ & + w_{\dot{e}}\left(\|\dot{\mathbf{e}}_{\mathrm{p},i}\|_2^2 + \|\mathbf{e}_{\omega,i}\|_2^2\right) \\ & + w_{\ddot{e}}\left(\|\ddot{\mathbf{e}}_{\mathrm{p},i}\|_2^2 + \|\dot{\mathbf{e}}_{\omega,i}\|_2^2\right) + \|\xi_i - \xi_{\mathrm{d}}\|^2_{\mathbf{W}_\xi} \\ & + \|\mathbf{x}_i\|^2_{\mathbf{W}_{\mathrm{x}}} + w_{\mathrm{u}}\|\mathbf{u}_i\|_2^2 + w_{\mathrm{v}}\|v_i\|_2^2, \end{aligned} \tag{45}$$

minimizes the tangential position and orientation path errors with the tangential error weights $w_{\parallel,\mathrm{p}} > 0$ and $w_{\parallel,\mathrm{o}} > 0$. Small tangential error norms $\|\mathbf{e}^{\parallel}_{\mathrm{p},i}\|_2^2$ and $\|\mathbf{e}^{\parallel}_{\mathrm{o},i}\|_2^2$ are ensured when choosing $w_{\parallel,\mathrm{p}}$ and $w_{\parallel,\mathrm{o}}$ sufficiently large. The path error velocity term $\|\dot{\mathbf{e}}_{\mathrm{p},i}\|_2^2 + \|\mathbf{e}_{\omega,i}\|_2^2$ and acceleration term $\|\ddot{\mathbf{e}}_{\mathrm{p},i}\|_2^2 + \|\dot{\mathbf{e}}_{\omega,i}\|_2^2$, with the weights $w_{\dot{e}} > 0$ and $w_{\ddot{e}} > 0$, bring damping into the path error systems. Here, the orientation error system is damped using the angular velocity and acceleration errors

$$\mathbf{e}_{\omega,i} = \boldsymbol{\omega}_{\mathrm{c},i} - \dot{\phi}_i \boldsymbol{\omega}_{\mathrm{r},i} \tag{46a}$$

$$\dot{\mathbf{e}}_{\omega,i} = \dot{\boldsymbol{\omega}}_{\mathrm{c},i} - \ddot{\phi}_i \boldsymbol{\omega}_{\mathrm{r},i} - \dot{\phi}_i \dot{\boldsymbol{\omega}}_{\mathrm{r},i}, \tag{46b}$$

with the current angular accelerations $\dot{\omega}_{\mathrm{c},i}$ and $\dot{\omega}_{\mathrm{r},i}$ of the end-effector and the reference path, respectively. The terms (46) are used instead of (29) and its derivative, because (29) depends on the error Jacobians $\mathbf{J}_{\mathrm{l}}^{-1}(\mathbf{e}_{\mathrm{o}}(t))$ and $\mathbf{J}_{\mathrm{r}}^{-1}(\mathbf{e}_{\mathrm{o}}(t))$, which complicate the MPC problem (43).

The term $\|\xi_i - \xi_{\mathrm{d}}\|^2_{\mathbf{W}_\xi} = (\xi_i - \xi_{\mathrm{d}})^{\mathrm{T}} \mathbf{W}_\xi (\xi_i - \xi_{\mathrm{d}})$, with the positive definite diagonal weighting matrix $\mathbf{W}_\xi$, makes the state $\xi$ of the path-parameter system (43c) approach the desired state $\xi_{\mathrm{d}} = [\phi_{\mathrm{f}}, 0, 0]^{\mathrm{T}}$. However, this formulation leads to very high path velocities $\dot{\phi}$ when the desired path parameter $\phi_{\mathrm{f}}$ is large, that is, when the reference path is long. The upper limit $\overline{\dot{\phi}}$ in (43k) limits $\dot{\phi}$ which yields a more desirable behavior along the path. Movement in the joint space is regularized using the term $\|\mathbf{x}_i\|^2_{\mathbf{W}_{\mathrm{x}}}$ with the positive definite diagonal weighting matrix $\mathbf{W}_{\mathrm{x}}$. Since the joint velocity $\dot{\mathbf{q}}_{\mathrm{d},i}$ in $\mathbf{x}_i$ includes the velocity in the joint nullspace, a minimization of the joint velocity $\dot{\mathbf{q}}_{\mathrm{d},i}$ stabilizes the nullspace. More involved methods exist to project $\dot{\mathbf{q}}_{\mathrm{d},i}$ to the nullspace (Ott, 2008) but these are computationally expensive. The last two terms in (45), with the weights $w_{\mathrm{u}} > 0$, $w_{\mathrm{v}} > 0$, serve as regularization terms.

Within the MPC, the planning problem (43) is solved at each time step with the same sampling time $T_{\mathrm{s}}$. In each step, the optimal control inputs $\mathbf{u}_i$ and $v_i$ are computed over the horizon $i = 1, \ldots, N$, but only the first control inputs $\mathbf{u}_1$ and $v_1$ are applied to the system.

**Remark.** Kinematic singularities happen when the Jacobian $\mathbf{J}(\mathbf{q})$ in (7) does not have full row rank. Close to a singularity, high joint velocities $\dot{\mathbf{q}}$ are needed to achieve a desired Cartesian velocity. Due to the constraint (43h), the joint velocities $\dot{\mathbf{q}}$ are bounded and the robot can successfully pass through kinematic singularities. This is demonstrated in the experiment video at https://www.acin.tuwien.ac.at/42d0/ at 3 min 26 s.

## 5. Orthogonal path error bounds

To stay within predefined error bounds, the MPC formulation (43) bounds the orthogonal position and orientation path errors using (43l) and (43m). The bounds are specified in the Cartesian space and allow for meaningful interpretations. The applications of the bounds are threefold. First, the bounds guide the end-effector along the path, second they ensure collision-freedom, and third they support encoding task-specific constraints. Guidance is important since the end-effector must pass the via-points with zero bounds. Appropriate bounds of the Cartesian reference path ensure collision freedom. In combination with these bounds, the reference path may be specified manually or by a higher-level planner, which is further discussed in Section 6 and Section 10. This way, the bounds capture the obstacle-free space, which can, for example, be obtained using convex collision-free sets (Deits and Tedrake, 2015). This idea contrasts an often-used approach of specifying the obstacle-occupied space and finding a solution in the remaining space (Vu et al., 2020). An advantage of the proposed approach is that the reference paths with the error bounds provide a simple description of the obstacle-free space, independent of the shape or amount of obstacles. Hence, the computational complexity becomes independent of the complexity of the obstacles and the environment. Lastly, the bounds can enforce task-specific constraints along a path, such as keeping the end-effector upright for transferring an open container filled with liquid or moving in a straight line to open a drawer.

The formulation of the bounding functions is detailed in this section. First, the case of symmetric path error bounds is examined and then generalized to asymmetric error bounds with respect to the desired basis vector directions to increase the flexibility of the method. The resulting bounds are visualized for one position basis direction in Figure 4.

## 5.1. Symmetric bounds

For a symmetric bounding of the orthogonal path errors, the representations $\mathbf{e}^{\perp}_{\mathrm{p,proj}}$, given in (23) Section 4.4.2, and $\mathbf{e}^{\perp}_{\mathrm{o,proj}}$, given in (41), are used. Thus, the error bounding constraints

$$\psi^{\mathrm{sym}}_{j,m}(\mathbf{e}^{\perp}_{j,\mathrm{proj}}) = \left(e^{\perp}_{j,\mathrm{proj},m}\right)^2 - \left(\Upsilon_{j,m}(\phi)\right)^2 \le 0, \tag{47}$$

with the bounding functions $\Upsilon_{j,m}(\phi)$, define symmetric box constraints centered on the reference path. The index $j \in \{\mathrm{p,o}\}$ indicates the position and orientation error, and the index $m = 1, 2$ refers to the basis vector direction, leading to four constraints in total.

The bounding functions $\Upsilon_{j,m}(\phi)$ are used to smoothly vary the error bounds such that no orthogonal path errors are allowed at the via-points but deviations from the path are possible in between the via-points depending on the respective task and situation. Let us assume that two via-points exist at the path points $\phi_l$ and $\phi_{l+1}$. In this work, the bounding functions for $\phi_l \le \phi \le \phi_{l+1}$ are chosen as fourth-order polynomials. They provide sufficient smoothness and are easy to specify by using the conditions

$$\Upsilon_{j,m}(\phi_l) = 0 \tag{48a}$$

$$\Upsilon_{j,m}(\phi_{l+1}) = 0 \tag{48b}$$

$$\Upsilon'_{j,m}(\phi_l) = s_0 \tag{48c}$$

$$\Upsilon'_{j,m}(\phi_{l+1}) = s_{\mathrm{f}} \tag{48d}$$

$$\Upsilon_{j,m}\left(\frac{\phi_l + \phi_{l+1}}{2}\right) = \Upsilon_{j,\max}, \tag{48e}$$

where the design parameters are $s_0$, $s_{\mathrm{f}}$, and $\Upsilon_{j,\max}$. Note that in general, any sufficiently smooth function can be used to bound the orthogonal path errors. For multiple via-points, the error bounding functions are defined for each segment between the via-points, meaning that $L$ via-points lead to $L-1$ segments with four bounding functions each. The case of two segments in 2D is visualized in Figure 4.

**Remark.** The error bounding functions $\Upsilon_{\mathrm{j,m}}(\phi)$ depend on the path parameter $\phi$ meaning that the orthogonal error is only correctly bounded as long as the tangential errors $\mathbf{e}^{\|}_{\mathrm{p}}$ and $\mathbf{e}^{\|}_{\mathrm{o}}$ are close to zero. This assumption holds true if the weights $w_{\|,\mathrm{p}}$ and $w_{\|,\mathrm{o}}$ in (45) are sufficiently large compared to the other weights.

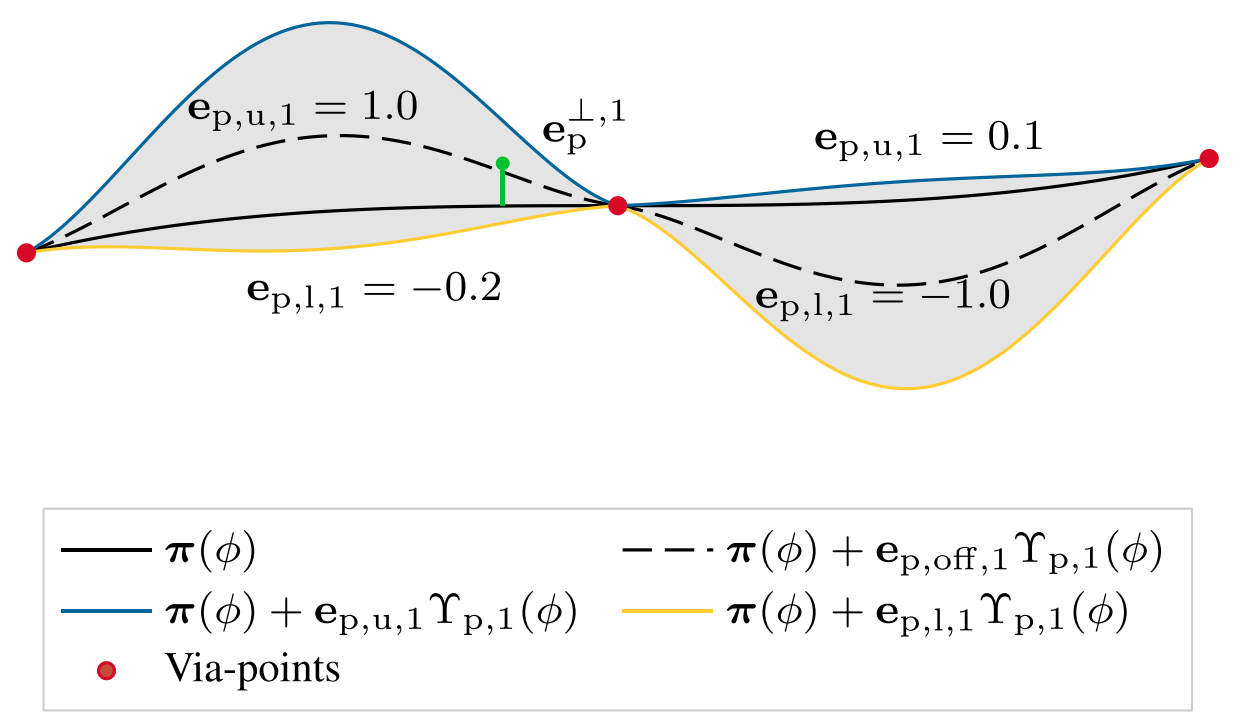

**Figure 4.** 2D Visualization of a position reference path $\boldsymbol{\pi}_{\mathrm{p}}(\phi)$ with error bounds. The symmetric error bounding functions $\Upsilon_{\mathrm{p,1}}$ are adapted by the upper and lower bounds $\mathbf{e}_{\mathrm{p,u,1}}$ and $\mathbf{e}_{\mathrm{p,l,1}}$ to asymmetrically bound the orthogonal position path error $\mathbf{e}^{\perp,1}_{\mathrm{p}}$. The shaded gray regions indicate the area where $\psi_{\mathrm{p},1}(\mathbf{e}^{\perp}_{\mathrm{p,proj}}) \le 0$. The error bounding functions $\Upsilon_{\mathrm{p,1}}$ and the upper and lower bounds $\mathbf{e}_{\mathrm{p,u,1}}$ and $\mathbf{e}_{\mathrm{p,l,1}}$ are defined for both segments between the three shown via-points.

## 5.2. Asymmetric bounds

The symmetric error bounds from the previous section are generalized in this section to allow for asymmetric bounds around the reference paths. The formulation

$$\psi_{j,m}(\mathbf{e}^{\perp}_{j,\mathrm{proj}}) \le 0, \quad j \in \{\mathrm{p,o}\}, \quad m = 1, 2, \tag{49}$$

with

$$\psi_{j,m}(\mathbf{e}^{\perp}_{j,\mathrm{proj}}) = \left(e^{\perp}_{j,\mathrm{proj},m} - e_{j,\mathrm{off},m}\right)^2 - \lambda^2_{j,m}, \tag{50}$$

generalizes (47) and is used in this work. The offsets $e_{j,\mathrm{off},m} = 1/2\left(e_{j,\mathrm{l},m} + e_{j,\mathrm{u},m}\right)\Upsilon_{j,m}$ and the bounds $\lambda_{j,m} = 1/2(e_{j,\mathrm{u},m} - e_{j,\mathrm{l},m})\Upsilon_{j,m}$ are parametrized using the upper and lower bounds $\mathbf{e}^{\mathrm{T}}_{j,\mathrm{u}} = [e_{j,\mathrm{u},1}, e_{j,\mathrm{u},2}]$ and $\mathbf{e}^{\mathrm{T}}_{j,\mathrm{l}} = [e_{j,\mathrm{l},1}, e_{j,\mathrm{l},2}]$, referred to with the index u and l, respectively, and the bounding functions $\Upsilon_{j,m}$ from Section 5.1. Note that choosing $e_{j,\mathrm{u},m} = 1$ and $e_{j,\mathrm{l},m} = -1$ reduces (50) to the symmetric case (47). Intuitively, the upper and lower bounds $e_{j,\mathrm{u},m}$ and $e_{j,\mathrm{l},m}$ define how much of the bound, defined by $\Upsilon_{j,m}$, can be used by the MPC to deviate in orthogonal direction from the reference path. For example, setting $e_{\mathrm{p,l,1}} = -0.5$ means that the orthogonal position path error can only deviate 50 % of the bounding function $\Upsilon_{\mathrm{p,1}}$ in the negative direction defined by the basis vector $\mathbf{b}_{\mathrm{p,1}}$. A 2D example is given in Figure 4.

**Remark.** Setting $\Upsilon_{j,m}(\phi_l) = 0$ and $\Upsilon_{j,m}(\phi_{l+1}) = 0$ in (48) may lead to numerical issues in the MPC (43). This is especially true for the orientation path error bounds as the orientation errors are only approximated. Therefore, relaxations $\Upsilon_{j,m}(\phi_l) = \epsilon_l > 0$ and $\Upsilon_{j,m}(\phi_{l+1}) = \epsilon_{l+1} > 0$ are used in practice. Note that in combination with asymmetric error bounds, this may entail discontinuous error bounds at the via-points. Choosing linearly varying upper and lower bounds $\mathbf{e}^{\mathrm{T}}_{j,\mathrm{u}}$ and $\mathbf{e}^{\mathrm{T}}_{j,\mathrm{l}}$ fixes these issues but is not further detailed here.

## 6. BoundMPC with linear reference paths

This section describes the application of the proposed MPC framework to piecewise linear reference paths. This is a special case of the above formulation yielding beneficial simplifications of the involved terms and expressions. The block diagram in Figure 5 provides an overview of the interaction of BoundMPC with the underlying joint-space computed-torque controller (2) and the higher-level planner that provides the Cartesian reference paths $\boldsymbol{\pi}_{\mathrm{p}}$ and $\boldsymbol{\pi}_{\mathrm{o}}$, including the error bounds. Algorithm 1 for BoundMPC with linear reference paths is described in this section. The output of BoundMPC is a joint-space trajectory sent at 10 Hz to the computed-torque controller (2). This controller interpolates the given joint-space trajectory and computes a joint torque at 8 kHz to control the robot. The reference path specification for BoundMPC is provided by a higher-level-planner, user input, or application specifications. For point-to-point movements, a higher-level planner for collision avoidance is helpful. Application specifications, for example, keeping a cup upright, are also easily encodable as a reference path with bounds. The reference path can also be manually specified if the user already has a set of desired path points. The bounds of the reference path are assumed to be obstacle-free. Admissible piecewise linear reference paths can be easily generated by just specifying via-points. This way, the reference path is defined by points to be passed with the robot's end-effector. The realized trajectory between the points is computed by the proposed MPC framework respecting the given bounds and the system dynamics. For obstacle avoidance, linear paths with Cartesian error bounds allow for a simple and accurate formulation of the collision-free space. Moreover, arc-length parametrization becomes trivial for linear paths. This section introduces section, piecewise linear reference paths, and the simplifications that follow from using such paths are derived.

### *6.1 Reference path formulation*

In this work, a piecewise linear reference path is a sequence of $L$ linear segments, that is,

$$\boldsymbol{\pi}_{\mathrm{p}}(\phi) = \begin{cases} \boldsymbol{\pi}_{\mathrm{p},0}(\phi) & \phi_0 \le \phi < \phi_1 \\ \boldsymbol{\pi}_{\mathrm{p},1}(\phi) & \phi_1 \le \phi < \phi_2 \\ \vdots & \\ \boldsymbol{\pi}_{\mathrm{p},L-1}(\phi) & \phi_{L-1} \le \phi < \phi_L \end{cases} \tag{51}$$

for the position reference path $\boldsymbol{\pi}_{\mathrm{p}}(\phi)$. The values $\phi_l$, $l = 0, \ldots, L$, indicate the locations of the via-points along the path, where $\phi_0$ refers to the starting and $\phi_L$ to the terminal point. In Figure 6, an example of two linear position reference segments is visualized. Each segment of the reference position path is parameterized by the slope $\mathbf{m}_l$ of unit length and the position $\mathbf{b}_l$ at the via-point

$$\boldsymbol{\pi}_{\mathrm{p},l}(\phi) = \mathbf{m}_l(\phi - \phi_l) + \mathbf{b}_l, \tag{52}$$

with the arc-length parameter $\phi$. To ensure continuity of $\boldsymbol{\pi}_{\mathrm{p}}(\phi)$, the following condition must hold

$$\mathbf{b}_{l+1} = \mathbf{m}_l(\phi_{l+1} - \phi_l) + \mathbf{b}_l. \tag{53}$$

Thus, the path (51) is uniquely defined by the via-points $\mathbf{b}_l$ $l = 0, \ldots, L$.

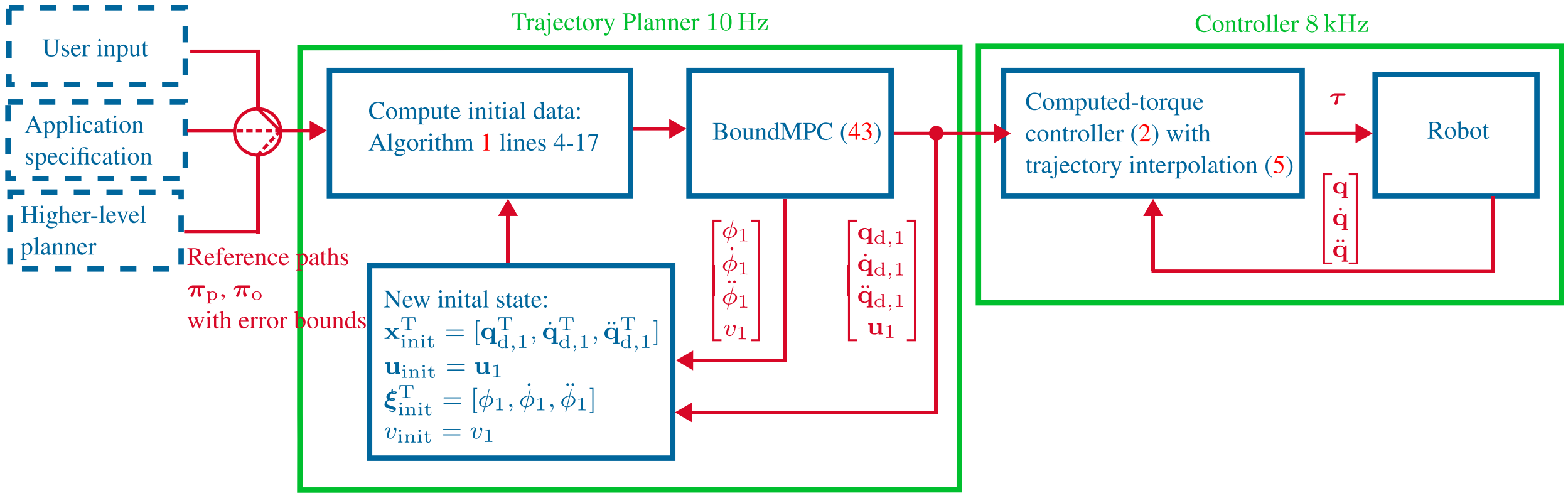

**Figure 5.** Block diagram of the application of BoundMPC (43) with the reference paths $\boldsymbol{\pi}_{\mathrm{p}}$ and $\boldsymbol{\pi}_{\mathrm{o}}$ and the error bounds as inputs. BoundMPC computes optimal joint-space trajectories $[\mathbf{q}_{\mathrm{d}}^{\mathrm{T}}, \dot{\mathbf{q}}_{\mathrm{d}}^{\mathrm{T}}, \ddot{\mathbf{q}}_{\mathrm{d}}^{\mathrm{T}}, \mathbf{u}^{\mathrm{T}}]$ that are provided to computed-torque controller (2). This controller outputs the joint torques $\boldsymbol{\tau}$ to control the robot. Since the control frequency of 8 kHz is higher than the trajectory planning frequency of 10 Hz, the desired joint trajectory is interpolated using (5). The computation of the reference paths $\boldsymbol{\pi}_{\mathrm{p}}$ and $\boldsymbol{\pi}_{\mathrm{o}}$ is beyond the scope of this work.

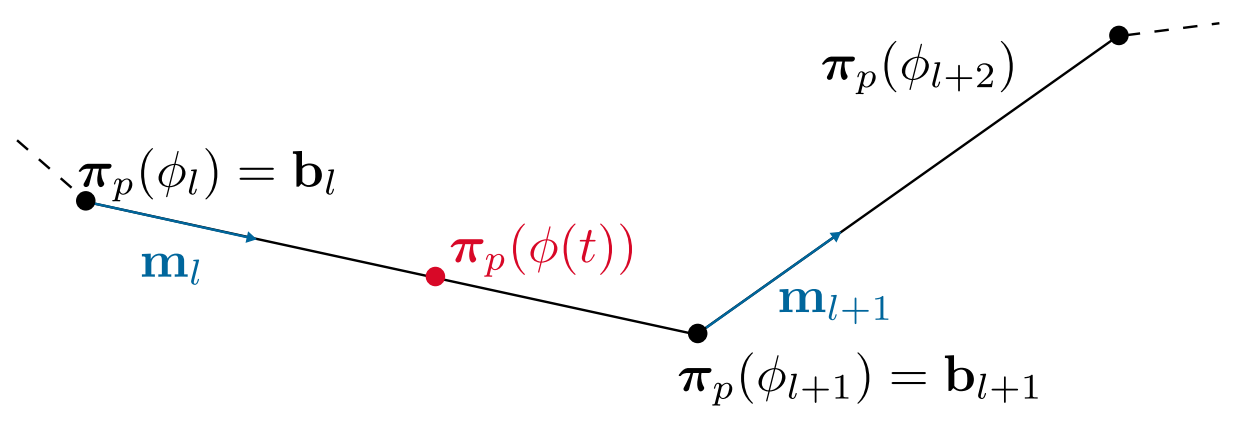

**Figure 6.** Visualization of a linear reference path. The current position along the path is $\boldsymbol{\pi}_\mathrm{p}(\phi(t))$ on the first linear segment.

The orientation reference path $\boldsymbol{\pi}_\mathrm{o}(\phi)$ is formulated analogously. Here, it is assumed that the angular reference velocity $\boldsymbol{\omega}_{\mathrm{r},l}$ between two via-points $l$ and $l+1$ is constant. Then, the path segments $\boldsymbol{\pi}_{\mathrm{o},l}(\phi)$ read as

$$\boldsymbol{\pi}_{\mathrm{o},l}(\phi) = \mathrm{Log}(\mathrm{Exp}(\boldsymbol{\omega}_{\mathrm{r},l}(\phi - \phi_l))\mathbf{R}_l), l = 0, \ldots, L-1, \tag{54}$$

where $\mathbf{R}_l$ is the end-effector's orientation at the via-point $l$. Moreover, the condition

$$\mathbf{R}_{l+1} = \mathrm{Exp}(\boldsymbol{\omega}_{\mathrm{r},l}(\phi_{l+1} - \phi_l))\mathbf{R}_l \tag{55}$$

must be satisfied to ensure continuity of the piecewise linear orientation reference path $\boldsymbol{\pi}_\mathrm{o}(\phi)$.

The piecewise linear position reference path (51) is used to compute the tangential and orthogonal position path errors for the optimal control problem (43). As discussed in Section 4.4.4, the orientation reference $\boldsymbol{\pi}_\mathrm{o}(\phi(t))$ is never explicitly used within the optimization problem (43) due to the iterative computation of the orientation path errors in (39). This is detailed for piecewise linear orientation reference paths in the next section.

**Remark.** The orthonormal basis vectors $\mathbf{b}_{\mathrm{p},1}$, $\mathbf{b}_{\mathrm{p},2}$, $\mathbf{b}_{\mathrm{o},1}$, and $\mathbf{b}_{\mathrm{o},2}$ are constant for each linear segment and will be referred to as $\mathbf{b}_{\mathrm{p,1,l}}$, $\mathbf{b}_{\mathrm{p,2,l}}$, $\mathbf{b}_{\mathrm{o,1,l}}$, and $\mathbf{b}_{\mathrm{o,2,l}}$ with $\mathrm{l} = 0, \ldots, \mathrm{L}-1$. Furthermore, the error bounding functions with the upper and lower bounds from Section 5 are denoted as $\Upsilon_{\mathrm{j,m,l}}$, $\mathbf{e}_{\mathrm{j,u,l}}$, and $\mathbf{e}_{\mathrm{j,l,l}}$, $\mathrm{j} \in \{\mathrm{p,o}\}$, $\mathrm{m} = 1, 2$, respectively.

**Remark.** Sampling-based methods, such as RRT, can be applied to efficiently plan collision-free piecewise linear reference paths in the Cartesian space, which is simpler than generating admissible paths in the joint space. The proposed MPC framework is then employed to find the joint trajectories that follow the paths within the given bounds.

## *6.2. Computation of the orientation path errors*

In this section, the orientation path error computations are derived for the piecewise linear orientation reference paths introduced in (54). Let us assume that the initial path parameter of the planner at the initial time $t_0$ is $\phi_0$. The current segment at time $t$ has the index $l_t$. The orientation path error (29) is integrated in the form

$$\begin{aligned}
\mathbf{e}_\mathrm{o}(t) &= \mathbf{e}_\mathrm{o}(t_0) + \mathbf{J}_\mathrm{l}^{-1}(\mathbf{e}_\mathrm{o}(t_0)) \int_{t_0}^{t} \boldsymbol{\omega}_\mathrm{c}(\tau)\mathrm{d}\tau \\
&\quad - \mathbf{J}_\mathrm{r}^{-1}(\mathbf{e}_\mathrm{o}(t_0)) \int_{\phi(t_0)}^{\phi(t)} \boldsymbol{\omega}_\mathrm{r}(\sigma)\mathrm{d}\sigma \\
&= \mathbf{e}_\mathrm{o}(t_0) + \mathbf{J}_\mathrm{l}^{-1}(\mathbf{e}_\mathrm{o}(t_0))(\boldsymbol{\Omega}_\mathrm{c}(t) - \boldsymbol{\Omega}_\mathrm{c}(t_0)) \\
&\quad - \mathbf{J}_\mathrm{r}^{-1}(\mathbf{e}_\mathrm{o}(t_0))(\boldsymbol{\Omega}_\mathrm{r}(\phi(t)) - \cancelto{0}{\boldsymbol{\Omega}_\mathrm{r}(\phi_0)}),
\end{aligned} \tag{56}$$

where $\boldsymbol{\Omega}_\mathrm{c}(t)$ and

$$\boldsymbol{\Omega}_\mathrm{r}(\phi(t)) = \sum_{l=0}^{l_t-1} \boldsymbol{\omega}_{\mathrm{r},l}(\phi_{l+1} - \phi_l) + \boldsymbol{\omega}_{\mathrm{r},l_t}(\phi(t) - \phi_{l_t}) \tag{57}$$

are the integrated angular velocities of the current robot motion and the piecewise linear reference path, respectively. Numerical integration is used to obtain $\boldsymbol{\Omega}_\mathrm{c}(t)$. Due to the assumption on the initial value of $\phi(t)$, the term $\boldsymbol{\Omega}_\mathrm{r}(\phi_0)$ vanishes. This formulation allows the representation of the current orientation path error $\mathbf{e}_\mathrm{o}(t)$ by the integrated angular velocities $\boldsymbol{\Omega}_\mathrm{c}(t)$ and $\boldsymbol{\Omega}_\mathrm{r}(\phi(t))$ and does not require the current orientation $\mathbf{R}_\mathrm{c}(t)$ or the current orientation reference path $\boldsymbol{\pi}_\mathrm{o}(\phi(t))$. For linear orientation reference paths, $\boldsymbol{\omega}'_\mathrm{r}(\phi(t)) = \mathbf{0}$ and $\boldsymbol{\rho}_\beta$ is constant, and the tangential orientation path error in (39) becomes

$$\begin{aligned}
\mathbf{e}_\mathrm{o}^{\|}(t) &= \mathbf{e}_\mathrm{o}^{\|}(t_0) + \int_{t_0}^{t} \Big(\boldsymbol{\rho}_\beta^\mathrm{T}(\mathbf{b}_{\mathrm{o},1}, \mathbf{b}_{\mathrm{o},2}, \boldsymbol{\omega}_\mathrm{r})\dot{\mathbf{e}}_\mathrm{o}(\tau)\Big)\boldsymbol{\omega}_\mathrm{r}(\phi(\tau))\mathrm{d}\tau \\
&= \mathbf{e}_\mathrm{o}^{\|}(t_0) + \sum_{l=0}^{l_t-1} \Big(\boldsymbol{\rho}_{\beta,l}^\mathrm{T}(\mathbf{e}_\mathrm{o}(t_{\phi_{l+1}}) - \mathbf{e}_\mathrm{o}(t_{\phi_l}))\Big)\boldsymbol{\omega}_{\mathrm{r},l} \\
&\quad + \Big(\boldsymbol{\rho}_{\beta,l_t}^\mathrm{T}(\mathbf{e}_\mathrm{o}(t) - \mathbf{e}_\mathrm{o}(t_{\phi_{l_t}}))\Big)\boldsymbol{\omega}_{\mathrm{r},l_t},
\end{aligned} \tag{58}$$

with

$$\boldsymbol{\rho}_{\beta,l} = \boldsymbol{\rho}_\beta(\mathbf{b}_{\mathrm{o},1,l}, \mathbf{b}_{\mathrm{o},2,l}, \boldsymbol{\omega}_{\mathrm{r},l}), \tag{59}$$

and the basis vectors $\mathbf{b}_{\mathrm{o},1,l}$ and $\mathbf{b}_{\mathrm{o},2,l}$ of the respective segment. Since the $t_{\phi_l}$ values (time associated with the path parameter $\phi_l$) in (58) are unknown, they are approximated as the last discrete time stamp after $\phi_l$. Furthermore, the orthogonal orientation path errors $\mathbf{e}_\mathrm{o}^{\perp,1}$ and $\mathbf{e}_\mathrm{o}^{\perp,2}$ are computed analogously to (58). Note that the values $\boldsymbol{\rho}_{\alpha,l}$, $\boldsymbol{\rho}_{\beta,l}$, and $\boldsymbol{\rho}_{\gamma,l}$, $l = 0, \ldots, L-1$ are required for the optimization problem (43). However, since they depend on values known prior to the optimization problem, they can be precomputed.

## *6.3. Position and orientation synchronization*

The position and orientation reference paths $\boldsymbol{\pi}_\mathrm{p}(\phi)$ and $\boldsymbol{\pi}_\mathrm{o}(\phi)$ are coupled by the path parameter $\phi$. Since $\phi$ is chosen to be the arc length of $\boldsymbol{\pi}_\mathrm{p}(\phi)$, the slopes $\mathbf{m}_l$ in (52) are of unit length but the angular velocities $\boldsymbol{\omega}_{\mathrm{r},l}$ in (55) are in general of different magnitudes. At the $l$-th via-point at $\phi = \phi_l$, the path velocity changes from $\dot{\phi}\mathbf{m}_{l-1}$ to $\dot{\phi}\mathbf{m}_l$ and the angular velocity along the orientation path from $\dot{\phi}\boldsymbol{\omega}_{\mathrm{r},l-1}$ to $\dot{\phi}\boldsymbol{\omega}_{\mathrm{r},l}$. Due to the slopes $\mathbf{m}_{l-1}$ and $\mathbf{m}_l$ being unit length, the position reference velocity only changes direction while the orientation reference velocity changes direction and magnitude. A very high angular acceleration is obtained at a via-point where, for example, $\mathbf{m}_{l-1} = \mathbf{m}_l$ and $\boldsymbol{\omega}_{r,l-1} = 0\boldsymbol{\omega}_{r,l}$, since the angular velocity along the path suddenly stops. In order to account for this issue, the weights of objective function (45) are chosen such that $w_{\|,p} \gg w_{\|,o}$, and the tangential orientation error $e^{\|}_{o,\text{proj}} = \beta$ is bounded similar to the orthogonal orientation errors $e^{\perp}_{o,\text{proj},1}$ and $e^{\perp}_{o,\text{proj},2}$ in (43m) as

$$\left(e^{\|}_{o,\text{proj}}(\phi)\right)^2 - \left(\Upsilon^{\|}_{o,l}(\phi)\right)^2 \leq 0 \tag{60}$$

in the $l$-th segment, $l = 0, \ldots, L$, which is a symmetric bound similar to (47) with the bounding function $\Upsilon^{\|}_{o}$. This way, the MPC can relax the synchronization of the position and orientation reference between the via-points due to $w_{\|,p} \gg w_{\|,o}$, but the synchronization at the via-points is ensured using (60). For the example case, the MPC can now arrive at the $l$-th via-point with angular velocity $\dot{\phi}\boldsymbol{\omega}_{r,l-1}$ and then exploit the bound (60) for $\phi > \phi_l$ to reduce the angular velocity. The MPC using linear reference paths is further detailed in Algorithm 1.

**Algorithm 1**. BoundMPC with linear reference paths

**Require:** sampling time $T_s$, reference paths $\boldsymbol{\pi}_p$, $\boldsymbol{\pi}_o$, initial states $\mathbf{x}_{\text{init}}, \mathbf{u}_{\text{init}}, \boldsymbol{\xi}_{\text{init}}, v_{\text{init}}$
$t \leftarrow t_0$
$\boldsymbol{\xi}^T(t_0) \leftarrow [0, 0, 0]$ ▷ Initialize state of path progress system
**while** $\phi(t) < \phi_f$ **do**
  **for** $l = 0, \ldots, L-1$ **do** ▷ For each linear segment
    obtain $\mathbf{m}_l, \boldsymbol{\omega}_{r,l}, \mathbf{b}_l$ ▷ Slopes and biases
    obtain $\Upsilon^{\|}_{o,l}$ ▷ Error bound
    **for** $j = p, o$ **do** ▷ For position and orientation
      obtain $\Upsilon_{j,1,l}$ and $\Upsilon_{j,2,l}$ ▷ Error bounds
      obtain $\mathbf{e}_{j,u,l}$ and $\mathbf{e}_{j,l,l}$
    **end for**
  **end for**
  compute $\mathbf{e}_o(t_0)$ using (25) ▷ Initial orientation errors
  compute $\mathbf{e}^{\|}_o(t_0)$, $\mathbf{e}^{\perp,1}_o(t_0)$, $\mathbf{e}^{\perp,2}_o(t_0)$ using (32)
  compute $\mathbf{J}^{-1}_l(\mathbf{e}_o(t_0)), \mathbf{J}^{-1}_r(\mathbf{e}_o(t_0))$ ▷ Jacobians
  **for** $l = 0, \ldots, L-1$ **do** ▷ For each linear segment
    compute $\boldsymbol{\rho}_{\alpha,l}$, $\boldsymbol{\rho}_{\beta,l}$ and $\boldsymbol{\rho}_{\gamma,l}$ ▷ Projection vectors
  **end for**
  solve (43) ▷ Optimization
  apply $\mathbf{u}_1$ to the robot
  $t = t + T_s$ ▷ Increment time
  $\mathbf{x}_{\text{init}} = \mathbf{x}_1$ ▷ Obtain new initial state
  $\mathbf{u}_{\text{init}} = \mathbf{u}_1$
  $\boldsymbol{\xi}_{\text{init}} = \boldsymbol{\xi}_1$
  $v_{\text{init}} = v_1$
**end while**

## 7. Online replanning

A major advantage using the proposed framework compared to state-of-the-art methods is the ability to replan a given path in real time during the motion of the robot in order to adapt to new goals or dynamic changes in the environment. Two different situations for replanning are possible based on when the reference path is adapted, that is,

1. outside of the current MPC horizon
2. within the current MPC horizon.

The first situation is trivial since it does not affect the subsequent MPC iteration. More challenging is the second situation, which requires careful consideration on how to adapt the current reference path such that the optimization problem remains feasible. Finding the replanned reference paths is left to a higher-level planner, which is beyond the scope of this work.

This situation is depicted for linear position reference paths at the current time $t$ in Figure 7 and applies similarly to linear orientation paths. In this figure, the replanned path $\tilde{\pi}_p(\phi)$ is entirely different and is valid starting from the current time $t$. Consequently, the new error bounds, depicted in green, must be chosen such that the current point $\mathbf{p}_c$ lies within the bounds to keep the optimization problem feasible. The position basis vectors of the new path $\tilde{\mathbf{b}}_{p,1}$ and $\tilde{\mathbf{b}}_{p,2}$ are used to obtain the new orthogonal errors by computing (23). These vectors are utilized to choose appropriate bounding functions. Note that these feasibility considerations do not consider the dynamics of the end-effector.

To simplify the replanning, the path state is projected onto the new reference path $\tilde{\boldsymbol{\pi}}_p(\phi)$, resulting in the new path state

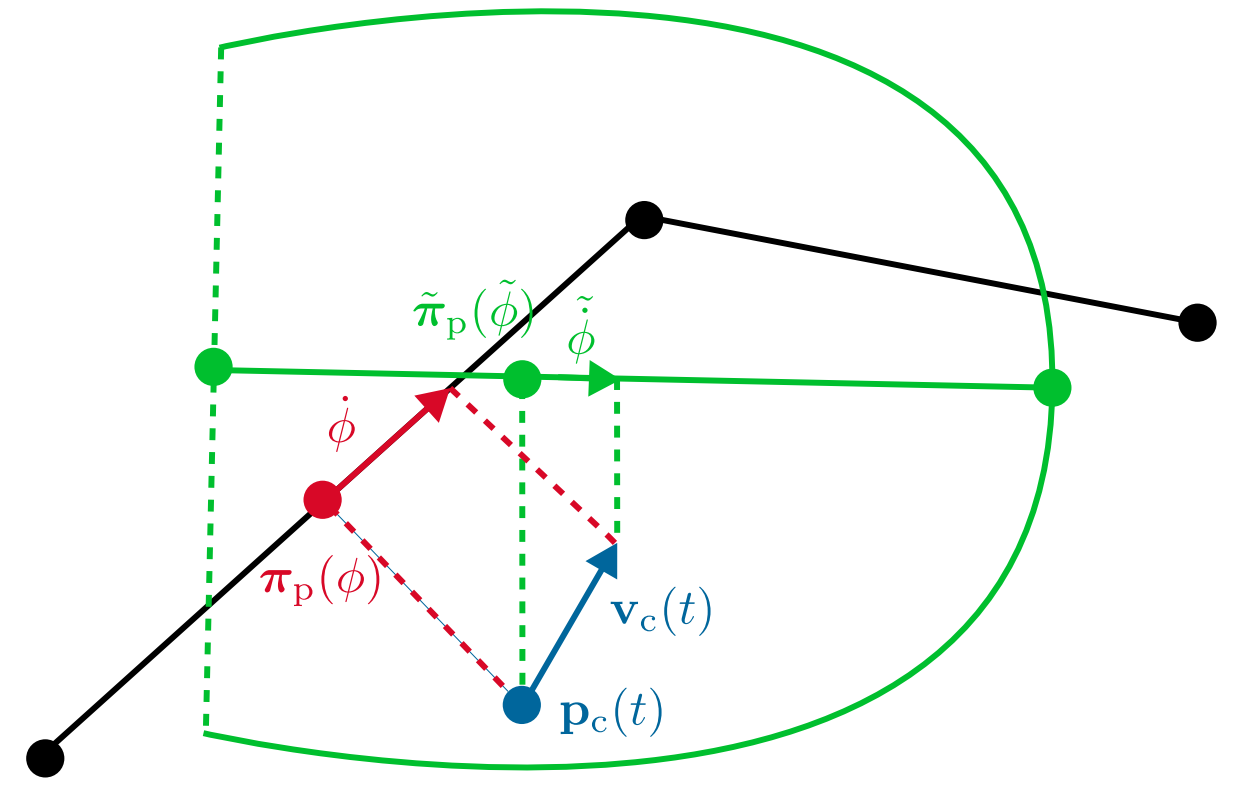

**Figure 7.** Replanning for a linear position reference path. The blue color indicates the current position $\mathbf{p}_c(t)$ and velocity $\mathbf{v}_c(t)$. The current reference point at time $t$ is $\boldsymbol{\pi}_p(\phi)$. Replanning takes place at the current path parameter $\phi$ with the path velocity $\dot{\phi}$ indicated by the red arrow along the path at the time $t$. The path state after replanning is given by the new reference path $\tilde{\pi}_p(\tilde{\phi})$ at the replanned path parameter $\tilde{\phi}$ with the path velocity $\dot{\tilde{\phi}}$.

$$\widetilde{\xi}(t) = \begin{bmatrix} \widetilde{\phi} \\ \dot{\widetilde{\phi}} \\ \ddot{\widetilde{\phi}} \end{bmatrix} = \begin{bmatrix} (\mathbf{p}_c^T(t) - \widetilde{\boldsymbol{\pi}}_p(0))\widetilde{\boldsymbol{\pi}}'_p(0) \\ \mathbf{v}_c^T(t)\widetilde{\boldsymbol{\pi}}'_p(0) \\ \mathbf{a}_c^T(t)\widetilde{\boldsymbol{\pi}}'_p(0) \end{bmatrix}, \tag{61}$$

with the current Cartesian velocity $\mathbf{v}_c(t)$, acceleration $\mathbf{a}_c(t)$ from (7) and (8), and the constant position path derivative $\widetilde{\boldsymbol{\pi}}'_p(0)$ of the first segment of the new reference path. Equation (61) ensures that the tangential position error $\mathbf{e}_p^{\|}$ in the new path state $\widetilde{\xi}(t)$ is zero and that its derivatives $\dot{\mathbf{e}}_p^{\|}$ and $\ddot{\mathbf{e}}_p^{\|}$ also become zero. The orientation path reference is adapted such that the tangential orientation error $\mathbf{e}_o^{\|}$ is zero and the orthogonal path errors $e^{\perp}_{o,proj,1}$ and $e^{\perp}_{o,proj,2}$ are within the error bounds. Since the position and orientation reference paths use the same path parameter $\phi$, the projected position path state (61) also determines the new orientation path state. Hence, $\dot{\mathbf{e}}_o^{\|}$ and $\ddot{\mathbf{e}}_o^{\|}$ are generally arbitrary, but the MPC can utilize the bound (60) to adapt to the new reference path, as shown in Section 10.

## 8. Implementation details

The proposed MPC is implemented on a KUKA LBR iiwa 14 R820 robot. For the optimization problem (43), the IPOPT optimizer (Wächter and Biegler, 2006) is used within the CasADi framework (Andersson et al., 2019). To improve convergence and robustness at the end of the paths, the objective function (45) is slightly modified to use

$$\mathbf{e}^{obj}_{p,i} = (1 - \sigma_e(\phi_i))\mathbf{e}^{\|}_{p,i} + \sigma_e(\phi_i)\mathbf{e}_{p,i} \tag{62a}$$

$$\mathbf{e}^{obj}_{o,i} = (1 - \sigma_e(\phi_i))\mathbf{e}^{\|}_{o,i} + \sigma_e(\phi_i)\mathbf{e}_{o,i} \tag{62b}$$

instead of the tangential errors $\mathbf{e}^{\|}_{p,i}$ and $\mathbf{e}^{\|}_{o,i}$. The sigmoid function

$$\sigma_e(\phi) = \frac{1}{1 + \exp(-100(\phi - (\phi_f - 0.02))} \tag{63}$$

is used to minimize the full path errors $\mathbf{e}_{p,i}$ and $\mathbf{e}_{o,i}$ at the end of the path, which ensures convergence to the final point.

**Remark.** The objective function (45) depends on the current end-effector acceleration for the terms $\|\ddot{\mathbf{e}}_{p,i}\|_2^2$ and $\|\dot{\mathbf{e}}_{\omega,i}\|_2^2$, which requires the computation of (8). As this is computationally expensive to evaluate, the derivative is numerically approximated within the MPC problem (43) as

$$\begin{bmatrix} \mathbf{a}_{c,i} \\ \dot{\boldsymbol{\omega}}_{c,i} \end{bmatrix} = \frac{1}{T_s}\left(\begin{bmatrix} \mathbf{v}_{c,i} \\ \boldsymbol{\omega}_{c,i} \end{bmatrix} - \begin{bmatrix} \mathbf{v}_{c,i-1} \\ \boldsymbol{\omega}_{c,i-1} \end{bmatrix}\right) \tag{64}$$

at horizon step $i = 1, \ldots, N$.

**Table 1.** Parameters of the MPC to be investigated.

| Parameter name | Symbol | Values |
|---|---|---|
| Horizon length | $N$ | 6, **10**, 30 |
| Path velocity limit/m s$^{-1}$ | $\overline{\dot{\phi}}$ | **0.5**, 0.75, 2 |
| Slope | $s_0 = s_f$ | **0.2**, 2 |

Bold entries indicate default values.

**Table 2.** Weights for the MPC cost function (45).

| Weight | Value |
|---|---|
| $w_{\|,p}$ | 1000 |
| $w_{\|,o}$ | 1 |
| $w_{\dot{e}}$ | 0.1 |
| $w_{\ddot{e}}$ | 0.05 |
| $\mathbf{W}_x$ | diag $(0.01\mathbf{I}_7, 0.01\mathbf{I}_7, 0.001\mathbf{I}_7)$ |
| $w_u$ | $10^{-4}$ |
| $w_v$ | 0.5 |
| $\mathbf{W}_\xi$ | diag(8, 5, 4) |

## 9. Parameter tuning

In this section, several parameters of the BoundMPC are studied in order to give further insight into the working principle and guide the user on how to choose the parameters. The considered parameters are given in Table 1, where the bold values indicate the default values, which are used when the other parameters are varied. The maximum allowable errors for the position $\Upsilon_{p,max} = 0.5$ m and orientation $\Upsilon_{o,max} = 50°$ are constant for all linear path segments and both orthogonal errors. All other parameters are given in Table 2. As discussed in Section 4.5, the weight $w_{\|,p}$ in (45) is chosen larger than the other weights, ensuring a small tangential position error. The position weight $w_{\|,p}$ is chosen larger than the orientation weight $w_{\|,o}$ to enable smooth transitions at the via-points, as discussed in Section 6.3. The constraint (60) is also added to the MPC formulation (43). The weight $w_u$, in combination with the weighting matrix $\mathbf{W}_\xi$ from Table 1, ensures smooth progress along the path, where the speed along the path is bounded from above by $\overline{\dot{\phi}}$. The weights $w_{\dot{e}}$ and $w_{\ddot{e}}$ in (45) are chosen small to regularize the orthogonal path errors but still allow exploitation of the error bounds. The weighting matrix $\mathbf{W}_x$ is used to regularize the joint position $\mathbf{q}_d$, velocity $\dot{\mathbf{q}}_d$ and acceleration $\ddot{\mathbf{q}}_d$ and stabilize the nullspace movement. Lastly, the joint jerk weight $w_u$ is chosen to be small to regularize the joint jerk input. Choosing the weight $w_v \geq w_u$ in (45) focuses on optimizing the Cartesian jerk instead of the joint jerk. In combination with the Cartesian reference paths, this leads to desirable behavior. Note that this only optimizes the Cartesian jerk along the path. Orthogonal to the path, only the Cartesian velocity and acceleration are optimized using the weights $w_{\dot{e}}$ and $w_{\ddot{e}}$.

**Table 3.** Via-points for the parameter studies. The bounds are chosen symmetrical around the reference. Hence, the basis vectors $\mathbf{b}_{j,1}$, $\mathbf{b}_{j,2}$ and the bounds $\mathbf{e}_{j,l}$ and $\mathbf{e}_{j,u}$, $j$ = p,o, are not needed for this study.

| Index $l$ | $\boldsymbol{\pi}_p(\phi_l)$/m | $\boldsymbol{\pi}_o(\phi_l)$/rad |
|---|---|---|
| 0 | $\mathbf{p}_0 = [1, 0, 0.66]^T$ | $\boldsymbol{\tau}_0 = \pi[0, 0.5, 0]^T$ |
| 1 | $\mathbf{p}_0 + [-2, 0, 0]^T$ | $\pi[0.71, 0, -0.71]^T$ |
| 2 | $\mathbf{p}_0 + [-1, 1, 0]^T$ | $\pi[-0.38, 0.38, 0.38]^T$ |
| 3 | $\mathbf{p}_0 + [-1, -1, 0]^T$ | $\pi[0.5, 0, 0]^T$ |
| 4 | $\mathbf{p}_0$ | $\boldsymbol{\tau}_0$ |

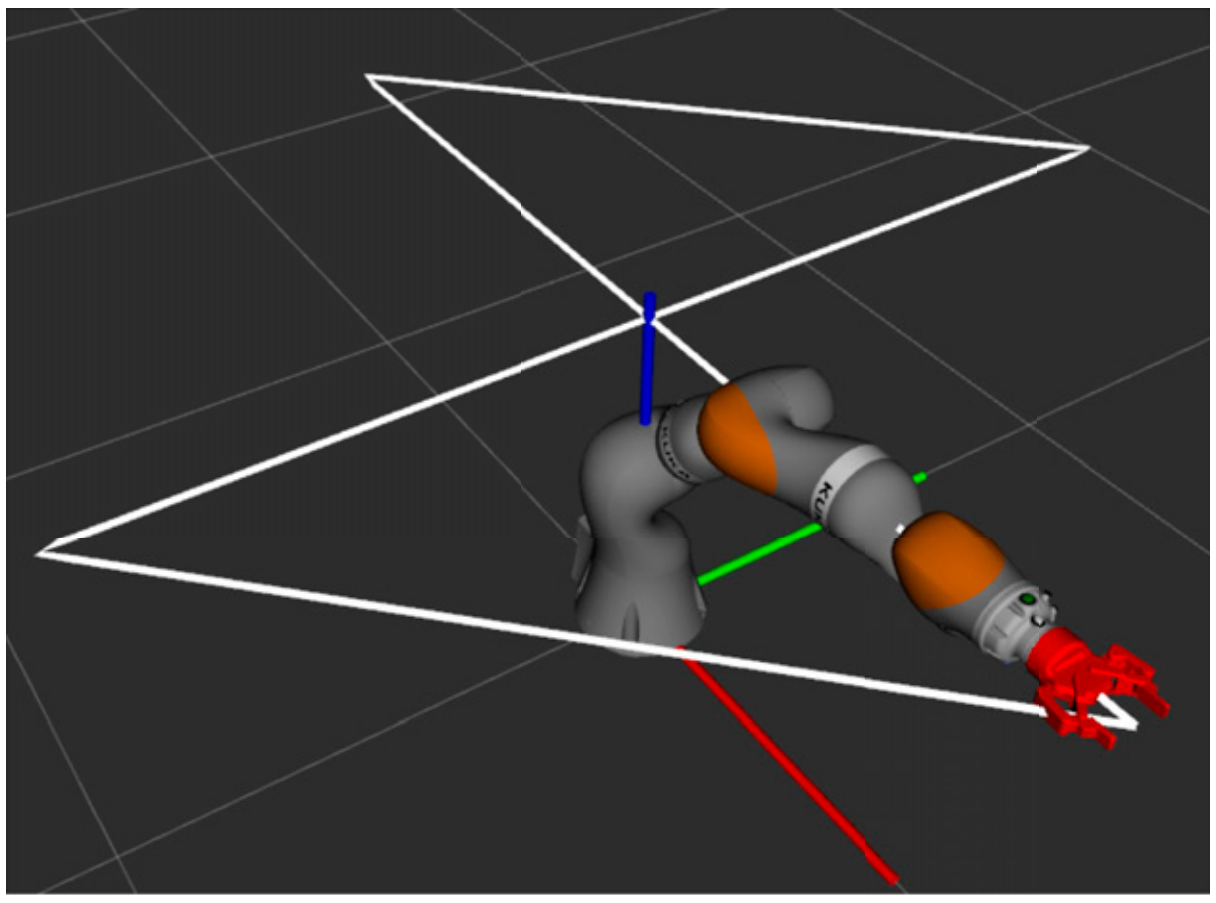

**Figure 8.** Visualization of the robot's configuration at the end of the example trajectory. The white line shows the position reference path of the end-effector. The world coordinate system is given by the RGB (xyz) axes at the robot's base. The end-effector in red is a gripper attached to the last robot link.

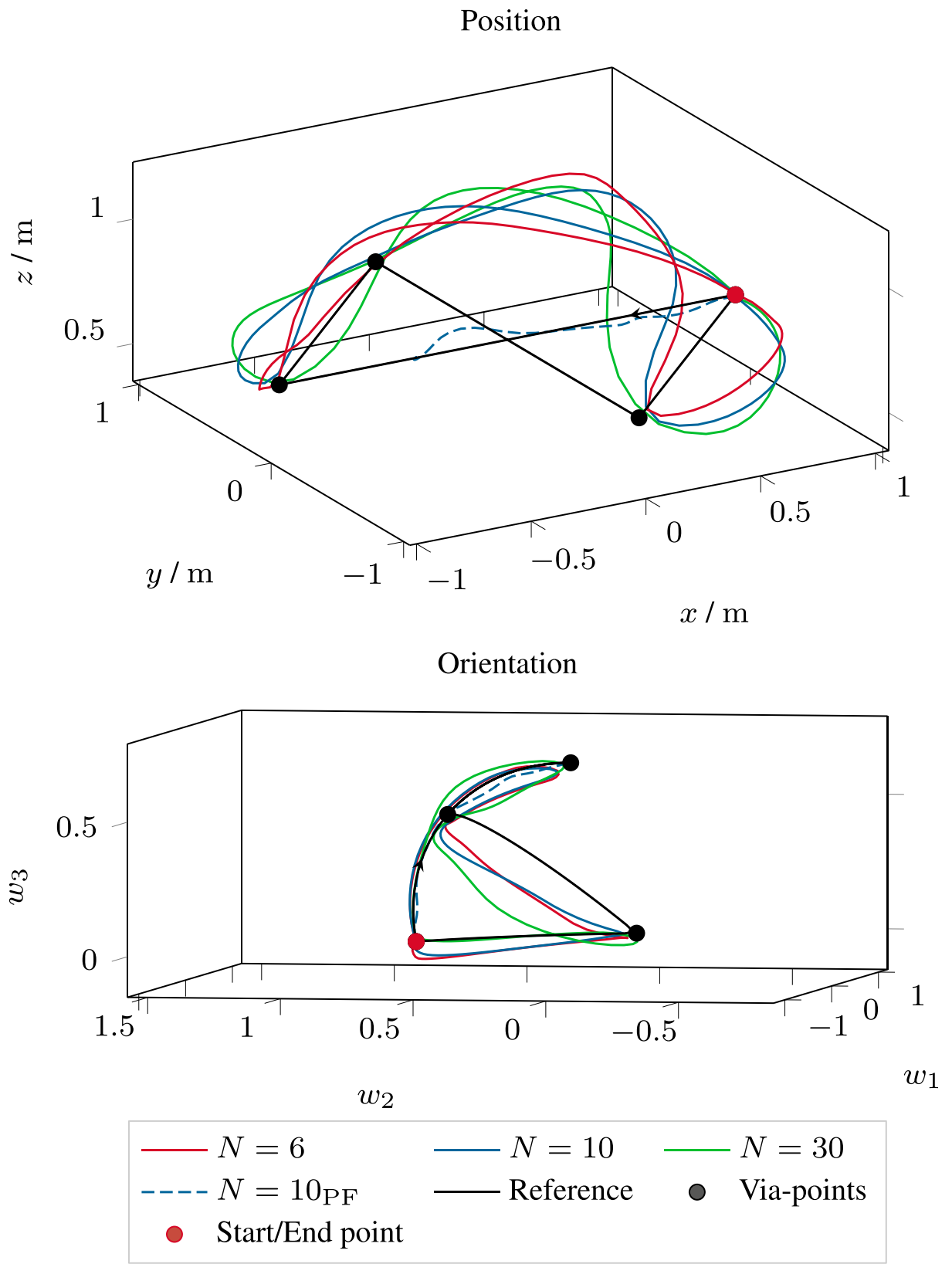

**Figure 9.** Comparison of the position and orientation trajectories of the MPC solutions for different horizon lengths $N$. The rotation trajectory is shown as the vector part of the quaternion representation of orientations $\mathbf{r}_q^T = [w_1, w_2, w_3, w]$.

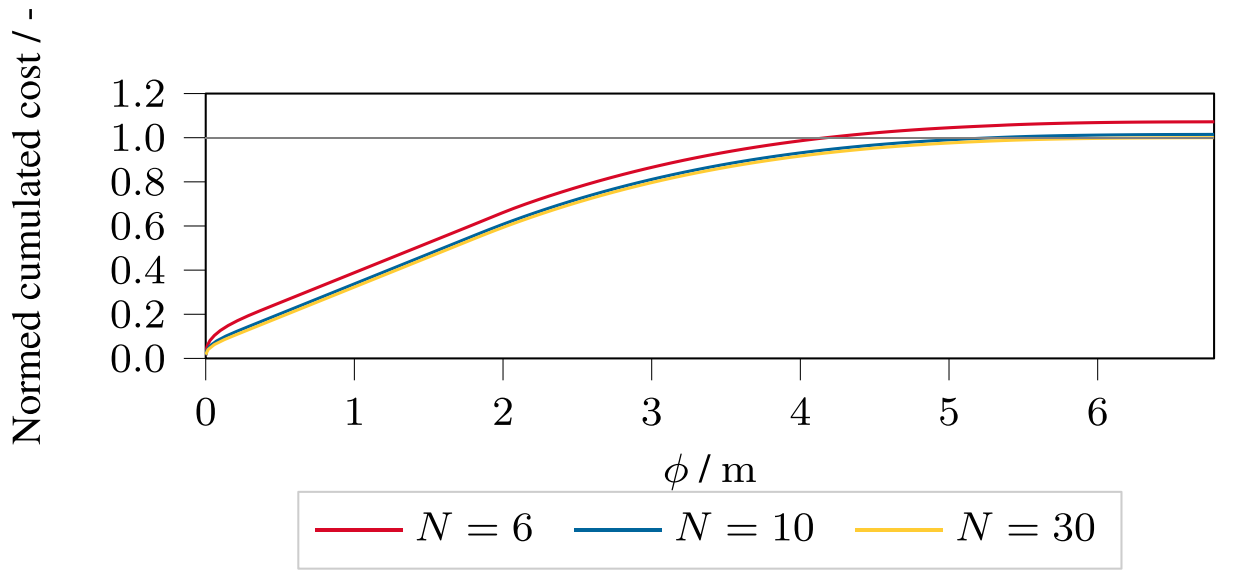

**Figure 10.** Cumulated costs of the trajectories for different horizon lengths $N$. The curves are normalized to the cumulated cost value of the trajectory for $N = 30$ at $\phi = \phi_f$. Each curve is the cumulated sum of the objective (45) along the resulting trajectories.

In order to study the influence of the parameters on the performance, a simple example with a reference path containing four linear segments is chosen. The path starts and ends at the same pose. The list of poses at the via-points is given in Table 3, and the position reference path is visualized with the robot in Figure 8. The trajectory covers a large portion of the robot's workspace.

A comparison of the position and orientation trajectories for different horizon lengths $N$ is shown in Figure 9. The orientation is plotted as the vector part of the unit quaternion representing the orientation. All solutions pass the via-points and converge to the final point without violating the constraints. The normalized cumulated costs of the trajectories are depicted in Figure 10. Longer horizon lengths lead to lower cumulated costs. Figure 11 shows the norm of the tangential error $\mathbf{e}_p^{\|}$ and the orthogonal error $\mathbf{e}_p^{\perp}$ for different horizon lengths $N$. The tangential errors are smaller than the orthogonal errors due to the large weights $w_{\|,p}$ and $w_{\|,o}$ in (45). It can be seen that the tangential errors increase close to the via-points, which can be traced back to the sudden change of the direction in the piecewise linear reference path. The orthogonal errors at the via-points (gray vertical lines in Figure 11) do not reach zero exactly due to the discretization of the control problem. Between the via-points, the MPC exploits the orthogonal deviation from the reference path within the given bounds to further minimize the objective function (45). For shorter horizons, planning is more difficult since the next via-point is not always within the planning horizon, which explains the higher cumulated costs in Figure 10. A longer horizon thus allows the robot to find better solutions, which is visible in the first and third segment of $\|\mathbf{e}_p^{\perp}\|$ in Figure 11, where a long horizon of $N = 30$ can adapt the position errors more appropriately for the upcoming via-point. This is also visible in the position trajectory in Figure 9.

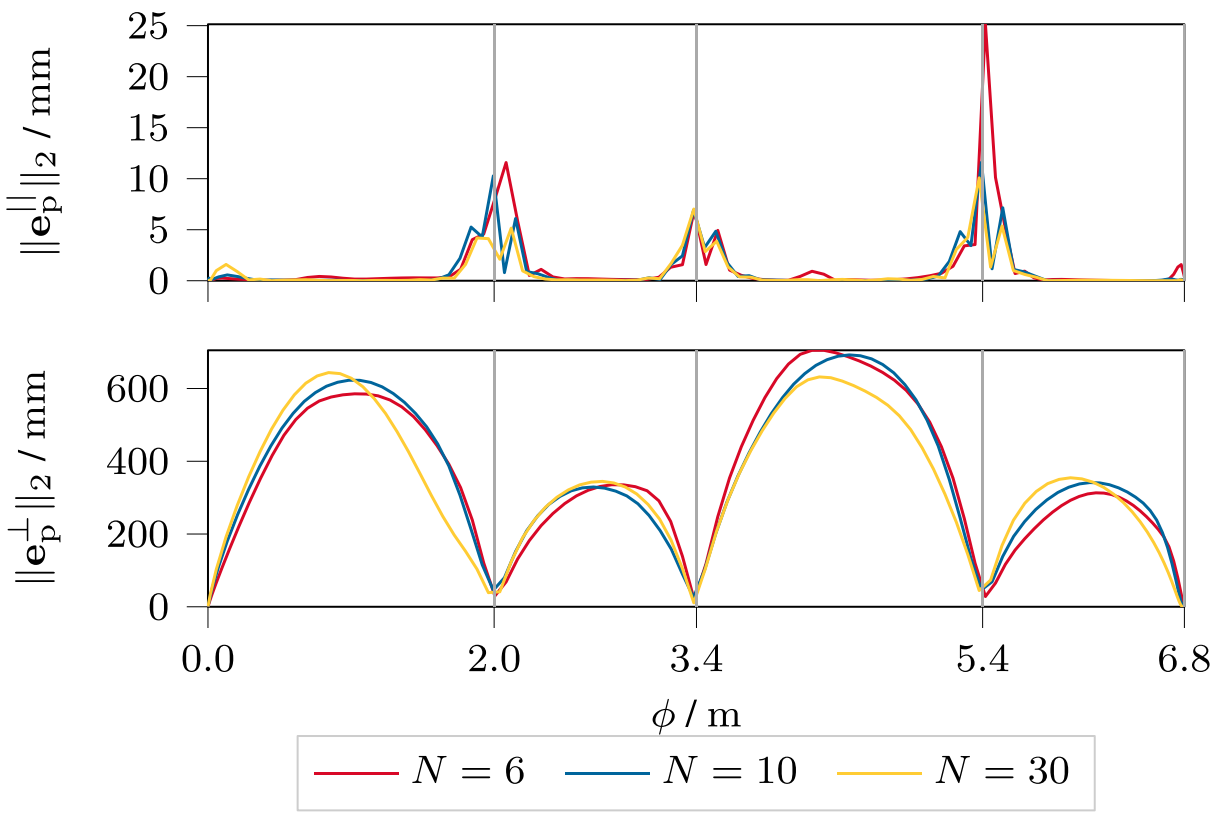

**Figure 11.** Norm of the tangential and orthogonal position error of the MPC solutions for different horizon lengths $N$.

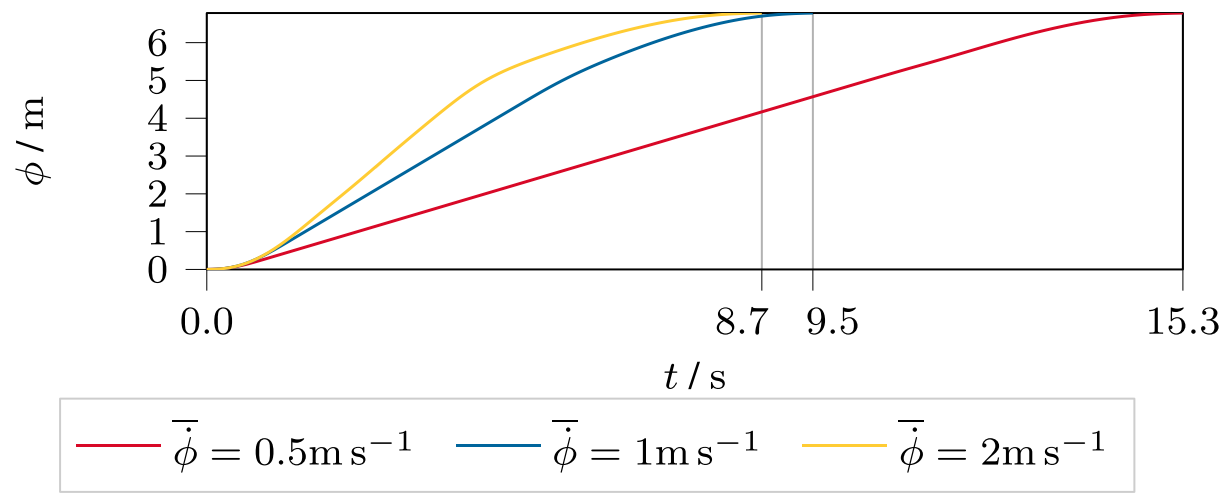

**Figure 12.** Influence of the path velocity limit $\bar{\dot{\phi}}$ on the path progress $\phi(t)$.

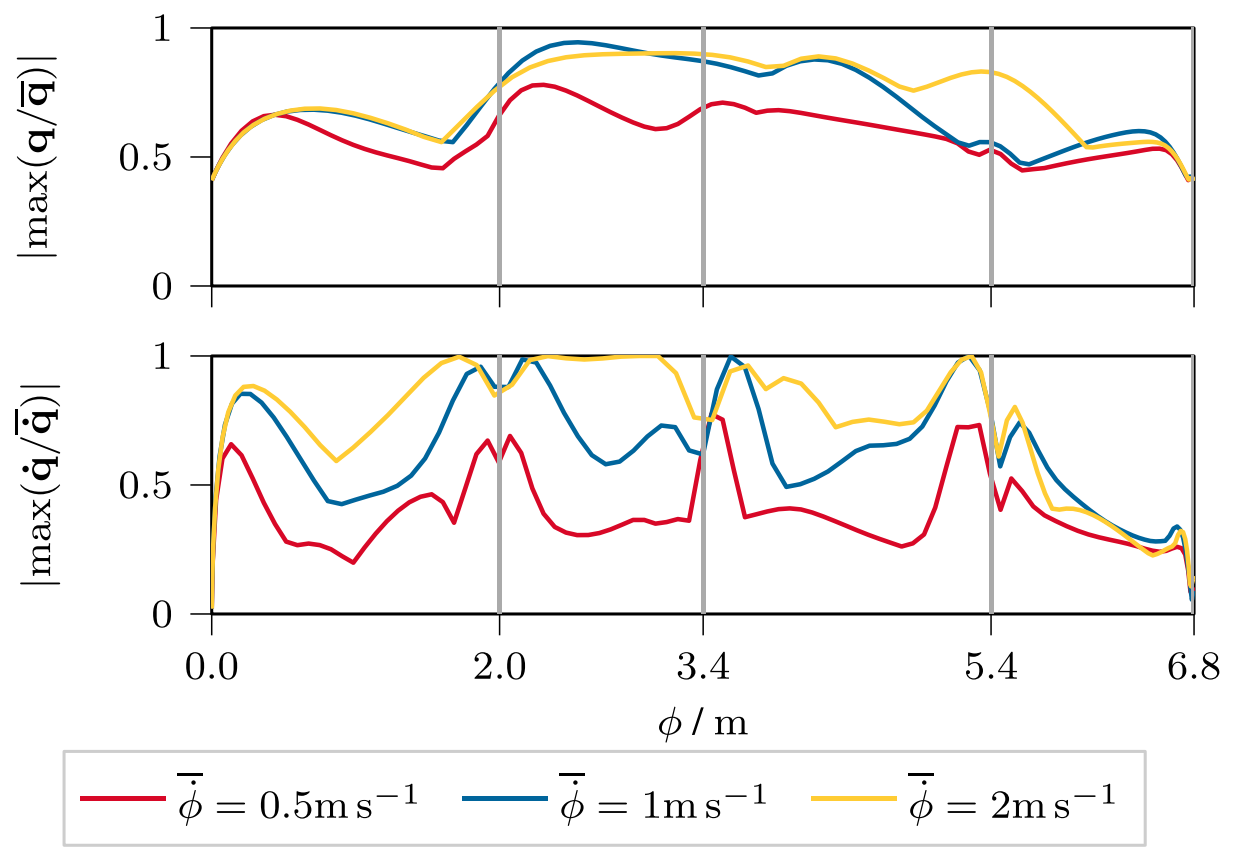

**Figure 13.** Exploitation of the joint position and velocity limits for different values of the path velocity limit $\bar{\dot{\phi}}$. The vertical axis shows the maximum absolute normalized joint position and velocity with respect to the joint and velocity upper bounds $\bar{\mathbf{q}}$ and $\bar{\dot{\mathbf{q}}}$.

**Table 4.** Trajectory duration $T$ in s for different parameter sets. The parameter values corresponding to the different columns are given in Table 1.

| Parameter | Value 1 | Value 2 | Value 3 |
|---|---|---|---|
| $N$ | 16.2 | 15.3 | 15.7 |
| $\bar{\dot{\phi}}$ | 8.7 | 9.5 | 15.3 |
| $s_0 = s_f$ | 15.3 | 15.3 | - |

**Table 5.** Statistics of computation time $T_{\mathrm{comp}}$ in ms for different horizon lengths $N$.

| $N$ | 6 | 10 | 30 |
|---|---|---|---|
| $\min(T_{\mathrm{comp}})$ | 8 | 17 | 139 |
| $\max(T_{\mathrm{comp}})$ | 38 | 64 | 848 |
| $\mathrm{mean}(T_{\mathrm{comp}})$ | 16 | 39 | 235 |

Table 4 shows the trajectory durations $T$ for the parameter sets introduced in Table 1. It is confirmed that the horizon length has a negligible impact on the trajectory duration $T$ due to the upper bound $\bar{\dot{\phi}}$ on the path velocity $\dot{\phi}$. The decreased cumulated costs still favor larger horizons. However, large horizon lengths lead to longer computation times and might undermine the real-time capability of the proposed concepts; see the minimum, maximum, and mean computation times for different horizon lengths $N$ in Table 5. A horizon length of $N = 10$ was chosen as a compromise for the experiments in Section 10. The optimization problem can thus be reliably solved within one sampling time of $T_s = 0.1$ s. While large horizons are more likely to find the globally optimal solution, they are not real-time capable. Thus, the choice of the horizon length is a trade-off between optimality and computational feasibility.

The influence of the path velocity bound $\bar{\dot{\phi}}$ on the path progress is depicted in Figure 12. This figure shows that a larger bound reduces the trajectory duration (see also Table 4), since the path velocity $\dot{\phi}$ allows faster progress along the path. The linear progression of the path parameter $\phi$ indicates that the bound $\bar{\dot{\phi}}$ is reached. For $\bar{\dot{\phi}} = 1\ \mathrm{m\,s^{-1}}$ and $\bar{\dot{\phi}} = 2\ \mathrm{m\,s^{-1}}$, the robot cannot reach the maximum velocity $\bar{\dot{\phi}}$ until the end of the trajectory, which causes the slope in Figure 12 to decrease. Further, the maximum absolute normalized joint position and velocity values are depicted in Figure 13. A value of 1 indicates that at least one axis is in a position or velocity limit. The MPC manages to exploit the limits without exceeding them. Larger values of the path velocity bound $\bar{\dot{\phi}}$ utilize the joint velocity range to a larger extent. Thus, the proposed approach uses the whole workspace of the robot with maximum velocities and without violating any limits.

The parameters $s_0$ and $s_f$ are used in the error bounds in (48) to specify the slopes of the bounding functions at the via-points; see the comparison of the orthogonal errors in Figure 14. The dashed lines indicate the error bounds, which differ only by their slope parameters at the via-points. These parameters influence the error bounds' shape but do not change the maximum value $\Upsilon_{\max}$. A low slope intuitively means that the error bounds around the via-points change slowly; hence, the via-points become more like a via-corridor. A high slope means that this via-corridor becomes very narrow. This is important when passing a via-point (see Figure 14). A longer via-corridor forces the orthogonal path errors to be small for a longer time, and thus, the error cannot be exploited well between the via-points. The effect of the slope $s_0 = s_f$ on the trajectory duration $T$, as presented in Table 4, can be neglected.

**Remark.** A repeated tracking of Cartesian paths, and specifically the paths described in Table 3, is possible due to the stabilization of the joint nullspace by the term $\|\mathbf{x}_i\|^2_{\mathbf{W}_x}$ in (45).

## 10. Experiments

The proposed trajectory planning framework *BoundMPC* features several important properties, such as position and orientation path error bounding, adhering to via-points, and online replanning. Six scenarios with different experiments are considered in the following to demonstrate the capabilities of the proposed framework. We grouped the following experiments into *tutorial* scenarion, which demonstrate the features and performance of BoundMPC in more detail, and *application* scenarios, which showcase everyday tasks. The first four scenarios are performed in a static environment and the last two scenarios change the task description online, demonstrating the online replanning capabilities. The scenarios are as follows:

S1: *(Tutorial Static)* Moving a large object between obstacles: A large object is moved from a start to a goal pose with obstacles. These obstacles constraint the position and orientation. The position and orientation paths are synchronized using via-points. This scenario is an example of understanding the components of the BoundMPC planning framework.
- *Experiment 1:* Moving the object in a highly constraint space.

S2: *(Application Static)* Opening a drawer: The robot's end-effector hooks into a cabinet drawer and opens it in a straight motion. A linear reference path segment with tight error bounds encodes the task-specific constraint of a straight motion. This is an application scenario that fulfills an everyday task.
- *Experiment 2:* Opening the drawer with different desired trajectory durations.

S3: *(Application Static)* Table wiping: Wiping a table with straight sponge motions across the surface. The task-specific straight motions require tight error bounds during the wiping motion but allow larger bounds in the transition segments. This is an application scenario that fulfills an everyday task.
- *Experiment 3:* Wiping the table with the sponge.

S4: *(Tutorial Static)* Collision avoidance for full robot: An object is transferred from one shelf in front of the robot to a shelf behind the robot. The environment contains six additional obstacles close to the robot base. Collision avoidance terms are added to the objective function (45) for collision avoidance of the entire robot's kinematic chain.
- *Experiment 4:* Transferring the object to the shelf behind the robot.

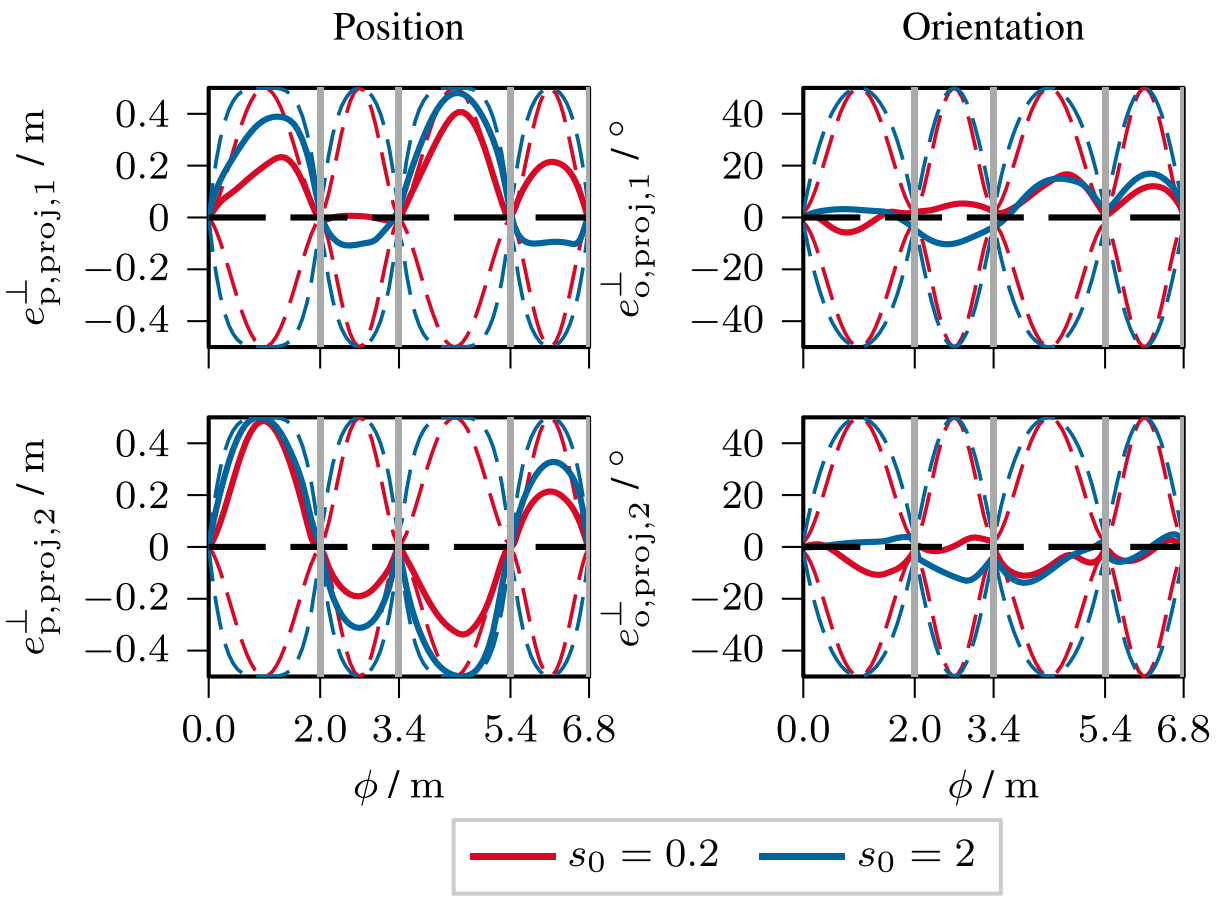

**Figure 14.** Comparison of the orthogonal error trajectories of the position and orientation for different slopes $s_0 = s_f$. The bounds are shown as dashed lines of the corresponding color. Via-points are indicated by vertical gray lines.

- *Experiment 5:* Evaluation of the task success for shifted obstacles along the $x$-axis.

S5: *(Tutorial Replanning)* Object grasp from a table: An object is grasped using one via-point at the pre-grasp pose. The path must be replanned during the robot's motion in real-time to adapt to the sudden change of the object pose. Collisions of the end-effector are avoided by using asymmetric error bounds. This scenario is used to showcase BoundMPC's replanning capabilities.
- *Experiment 6:* Grasping the object with three replanning steps showcasing the replanning capabilities.
- *Experiment 7:* Randomly updating the object's pose until replanning fails.
- *Experiment 8:* Planning 50 random poses with a comparison to state-of-the-art motion planners; see below.

S6: *(Application Replanning)* Transfer of a filled open container: Transferring a filled open container from an initial position onto a table while keeping it upright to avoid liquid spillage. A path segment with tight orientation error bounds encodes the task-specific upright constraint. This is an application scenario that fulfills an everyday task.
- *Experiment 9:* Transferring the open container.
- *Experiment 10:* Randomized put-down positions with replanning for the container.

The 7-DoF robot KUKA LBR iiwa 14 R820 is used for the experiments. The BoundMPC implementation is the same across all experiments, and the weights for the cost function (45) used in all experiments are listed in Table 2. Choosing the planning frequency $f_c$ = 10 Hz and the number of planning grid points to be $N = 10$ leads to a time horizon of $T_c = N/f_c = 1$ s. The controller considers the current and upcoming $n = 4$ segments of the reference path. Additionally, BoundMPC was applied to human–robot handovers in Oelerich et al. (2024).

The motion planning pipeline consists of two components:

1. The higher-level planner provides the linear reference path segments $\boldsymbol{\pi}_{\mathrm{p},l}$ and $\boldsymbol{\pi}_{\mathrm{o},l}$, $l = 1, \ldots, L - 1$, together with the basis vectors $\mathbf{b}_{j,m,l}$, and the orthogonal error bound parameters $\Upsilon_{j,m,l}$, $\mathbf{e}_{j,\mathrm{u},l}$, and $\mathbf{e}_{j,\mathrm{l},l}$, $j \in \{\mathrm{p,o}\}$, $m = 1, 2$, to enable collision-free movement within the bounds. In this work, these reference path segments and bound parameters are manually chosen for the experiments. Future work will deal with a more systematic way of finding the reference path and its bounds based on measured data from the environment. Since linear reference paths are easy to specify, this planner can replan the reference path during the robot's motion.
2. BoundMPC is used to compute a trajectory to follow the reference paths optimally while exploiting the bounds.

The path bounds are assumed to be obstacle-free, which is ensured by the user input or the higher-level planner, as discussed in Section 6. In this work, collision freedom is only considered for the end-effector, and future work will extend this principle to the whole kinematic chain of the robot. Since the bounding functions are specified in Cartesian space, task-specific constraints are easily encoded in this framework.

In order to show the advantages of using the BoundMPC framework, it is compared to the following state-of-the-art motion planners:

- *Joint trajectory tracking MPC (JT-MPC)*: This MPC framework uses a joint-space reference trajectory. Trajectory tracking MPCs are popular real-time capable planning methods (see, e.g., Arbo et al., 2017; Dai et al., 2021; Schoels et al., 2020a; Toussaint et al., 2022; Wang et al., 2023). The reference trajectories for the tracking MPC are computed by optimizing a spline trajectory through the inverse kinematic solutions of the given via-points, (see Toussaint et al., 2022). The reference trajectory planning is fast but does not consider collisions. The collision avoidance is performed through a collision term in the MPC objective function based on potential functions (Vu et al., 2020). In order to enable a fair comparison, collision avoidance is only performed for the robot's end-effector except for scenario S6 where the entire kinematic chain is checked for collisions. A time horizon of 2 s is used.
- *Cartesian trajectory tracking MPC (CT-MPC)*: The CT-MPC uses the same formulation as JT-MPC but with Cartesian reference trajectories for the positions and orientation instead of joint-space reference trajectories. A time horizon of 1 s is used.
- *Path tracking BoundMPC (BoundMPC-PT)*: This variant of BoundMPC minimizes $\mathbf{e}_{\mathrm{p}}$ and $\mathbf{e}_{\mathrm{o}}$ instead of $\mathbf{e}_{\mathrm{p}}^{\|}$ and $\mathbf{e}_{\mathrm{o}}^{\|}$ in the objective function (45). This leads to a formulation similar to Arbo et al. (2017), Astudillo et al. (2022b), and Mehrez et al. (2017) but uses a suboptimal reference path.

The output of all planners is a joint-space trajectory, which is executed by the robot using the computed-torque controller (2). Thus, the planner outputs are comparable.

**Remark.** Importantly, BoundMPC requires a pre-planned collision-free reference path. While this aspect is not the focus of this work preliminary experiments with planning paths based on collision-free convex sets show promising results. The method proposed in Deits and Tedrake (2015) can efficiently compute these sets. Thus, the comparisons lack the computational overhead of computing the reference paths.

The planners are numerically compared for the following metrics:

- The planning time $t_{\mathrm{plan}}$ is the time the trajectory planner takes each step to generate the joint-space trajectory for the next time step.
- The geometric path length $l_{\mathrm{path}}$ is the length of the end-effector's position path of the resulting trajectory for each planner.
- The time it takes for the robot to traverse the trajectory, indicated by $t_{\mathrm{traj}}$.
- The success rate $r_{\mathrm{s}}$ is the percentage of successful planning attempts relative to the failed ones.

## 10.1. Moving a large object between obstacles

*10.1.1. Goal.* This task aims to move a large object from a start to a goal pose through a confined space defined by obstacles.

*10.1.2. Setup.* The object is positioned on a table, from which it is grasped and then transferred to the goal pose. The obstacles create a narrow passage in the middle of the path and at the goal pose. Since the object is too large to pass the obstacles in the initial orientation, the robot must turn it by 90° in the first segment, then pass the obstacles, and finally turn the object to the goal orientation to put it down safely. BoundMPC allows setting via-points to traverse the bottlenecks without collision where no stopping at the via-points is required. The path parameter $\phi(t)$ synchronizes the end-effector's position and orientation at the via-points. Thus, the position and orientation via-points are reached at the same time, making it possible to traverse the path safely within the given bounds. A video of the experiment can be found at https://www.acin.tuwien.ac.at/42d0/.

*10.1.3. Results.* Figure 15 illustrates the robot configuration at four time instances of the trajectory to visualize the robot's motion in the scenario. For further understanding of the basis vectors, Figure 16 displays the position basis vectors $\mathbf{b}_{\mathrm{p},1}$ and $\mathbf{b}_{\mathrm{p},2}$ for each linear path segment. The planned trajectory is shown in Figure 17 with the projected error bounds on the Cartesian planes. Five via-points and four line segments are used to construct the reference path. In the second segment, the robot moves with the large object between two obstacles. At the end of the trajectory, the robot with the large object in its gripper puts the object down between two more obstacles at the desired goal position. Only four via-points are visible in the orientation plot in Figure 17 because the last two via-points coincide. All via-points are passed for the position and orientation.

Further details are given in the error trajectories in Figure 18. The low bound in the direction of the basis vector $\mathbf{b}^{\mathrm{T}}_{\mathrm{p},1,2} = [0, 0, 1]$ for $e^{\perp}_{\mathrm{p,proj},1}$ (also depicted in Figure 16 by the red arrow in the second segment) ensures low deviations in $z$-direction. This bounding does not affect the other error direction $\mathbf{b}^{\mathrm{T}}_{\mathrm{p},2,2} = [-0.83, 0.55, 0]$ for $e^{\perp}_{\mathrm{p,proj},2}$ in this segment, as seen in Figure 18. Furthermore, the last segment of the position path is bounded to have low errors for both directions $\mathbf{b}_{\mathrm{p},1,4}$ and $\mathbf{b}_{\mathrm{p},2,4}$ with the orthogonal errors $e^{\perp}_{\mathrm{p,proj},1}$ and $e^{\perp}_{\mathrm{p,proj},2}$, to avoid the obstacles around the goal pose. The orientation errors $e^{\perp}_{\mathrm{o,proj},2}$ and $e^{\perp}_{\mathrm{o,proj},2}$ in the second segment are constrained to be small since only the rotation around the normed reference path angular velocity $\boldsymbol{\omega}^{\mathrm{T}}_{\mathrm{r},2}/\|\boldsymbol{\omega}_{\mathrm{r},2}\|_2 = [0, 0, 1]$ is allowed to ensure a collision-free trajectory. Similar to the position, the orthogonal orientation errors $e^{\perp}_{\mathrm{o,proj},1}$ and $e^{\perp}_{\mathrm{o,proj},2}$ are bound to be low to avoid collisions with the obstacles around the goal pose. The robot passes the obstacles in the confined space and exploits the orthogonal error bounds between the obstacles, finding the optimal trajectory to solve the task. Figure 18 additionally shows the trajectory when using large error bounds with a maximum position error of 100 mm and a maximum orientation error of 20° for each segment. This trajectory moves outside the free space in the second and last segments and results in collisions between the robot and the obstacles. This shows the necessity of specifying appropriate bounds.

Furthermore, Figure 18 shows the bounded tangential orientation path error $e^{\parallel}_{\mathrm{o,proj}}$ in the last plot. The bounds force $e^{\parallel}_{\mathrm{o,proj}}$ to be low at the via-points but allow deviations between the via-points, as discussed in Section 6.3. This is, for example, utilized in the third segment. A large orientation change is required in this segment, while the last segment has no orientation change. Hence, the MPC successfully leverages the freedom in $e^{\parallel}_{\mathrm{o,proj}}$ to optimize the motion along the orientation path.

Figure 19 depicts the approximation error $\alpha_{\mathrm{approx}}$ of $\mathbf{e}^{\perp,1}_{\mathrm{o}}$ between the actual tangential and orthogonal orientation path errors obtained by (32) and their approximations (39). The approximation error is low at the start of the horizon and grows larger toward the end due to the integral approximation in (58). Furthermore, the approximation error is larger during the first and third segments, which contain significant changes in the orientation reference. A large change requires large angular accelerations, which causes the first-order approximation (29) to become less accurate. The approximation errors for $\mathbf{e}^{\parallel}_{\mathrm{o}}$ and $\mathbf{e}^{\perp,2}_{\mathrm{o}}$ are in a similar range.

*10.1.4. Comparison.* The proposed BoundMPC planner is compared to two different motion planners for this task, that is, JT-MPC and CT-MPC. The comparison metrics are the planning time $t_{\mathrm{plan}}$, the geometric path length $l_{\mathrm{path}}$, the trajectory duration $t_{\mathrm{traj}}$, the success rate $r_{\mathrm{s}}$, and formulation-specific collision avoidance aspects.

The results of the comparison are listed in Table 6. JT-MPC and CT-MPC solve the given task with computational effort comparable to BoundMPC. However, CT-MPC struggles to follow the given trajectory well. This is depicted in Figure 20. The Cartesian velocity $\mathbf{v}_{\mathrm{c}}$ is larger for CT-MPC in the second half of the trajectory. The norm of the tangential path error $\mathbf{e}^{\parallel}_{\mathrm{p}}$ remains low with the

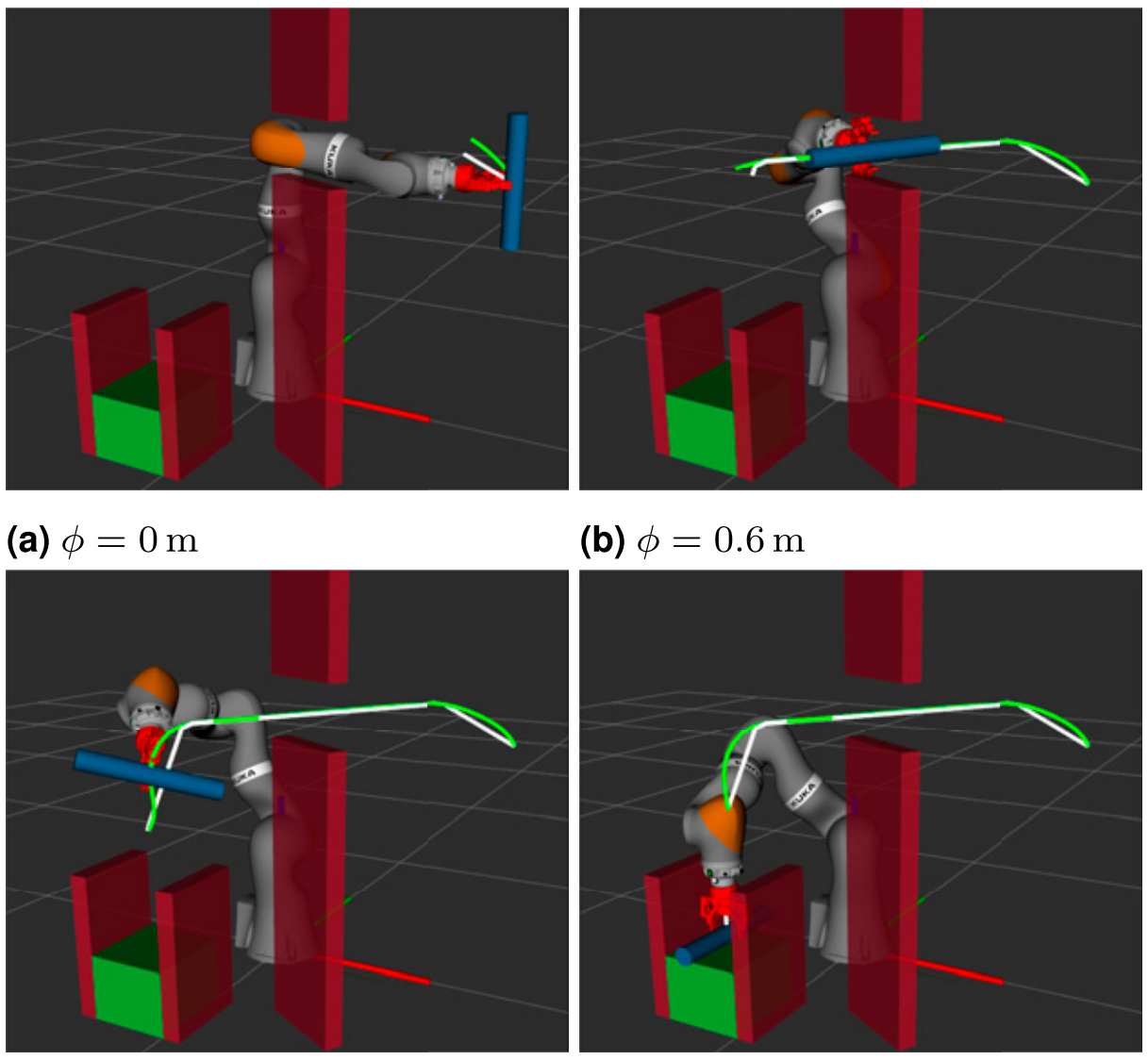

**(a)** $\phi = 0\,\mathrm{m}$ **(b)** $\phi = 0.6\,\mathrm{m}$ **(c)** $\phi = 1.1\,\mathrm{m}$ **(d)** $\phi = \phi_{\mathrm{f}} = 1.62\,\mathrm{m}$

**Figure 15.** Visualization of the robot's configurations in different time instances along the path: The end-effector in red is a gripper attached to the robot flange. The blue cylinder represents the object to be transferred to the green table. The red walls are obstacles to be avoided. The green and white lines represent the planned trajectory and the reference path of the end-effector, respectively. The world coordinate system is given by the RGB (xyz) axes at the robot's base. (a) $\phi$ = 0 m, (b) $\phi$ = 0.6 m, (c) $\phi$ = 1.1 m, and (d) $\phi = \phi_{\mathrm{f}}$ = 1.62 m.

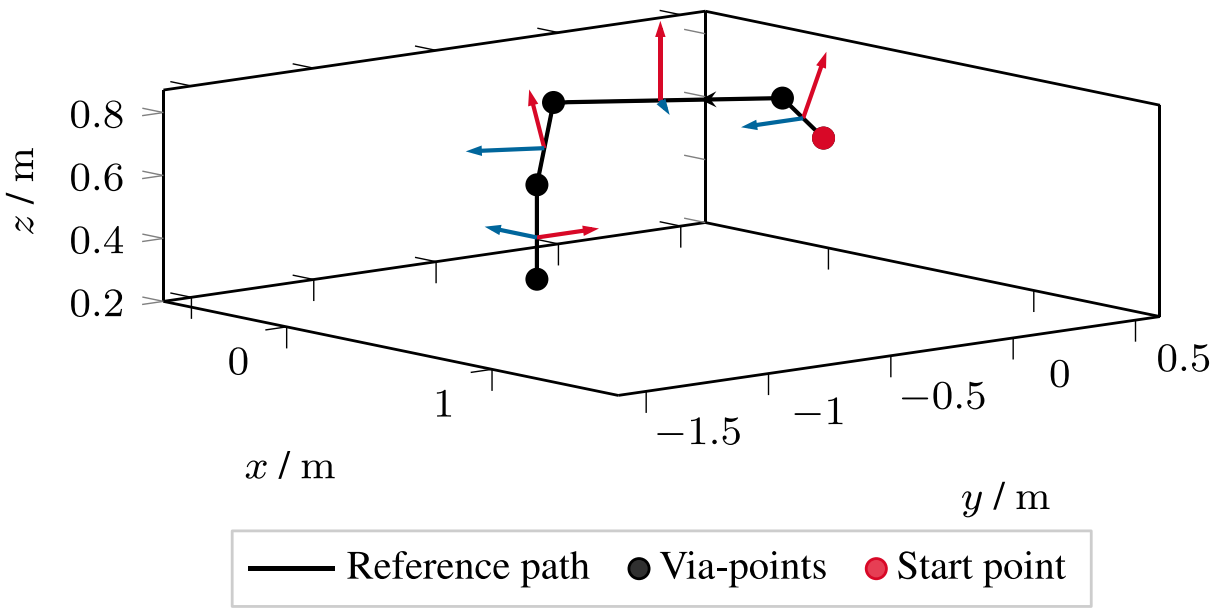

**Figure 16.** Position basis vectors for each linear reference path segment for the object transfer scenario. The red and blue arrows indicate the basis vectors $\mathbf{b}_{\mathrm{p},1}$ and $\mathbf{b}_{\mathrm{p},2}$, respectively, which span the orthogonal position error plane.

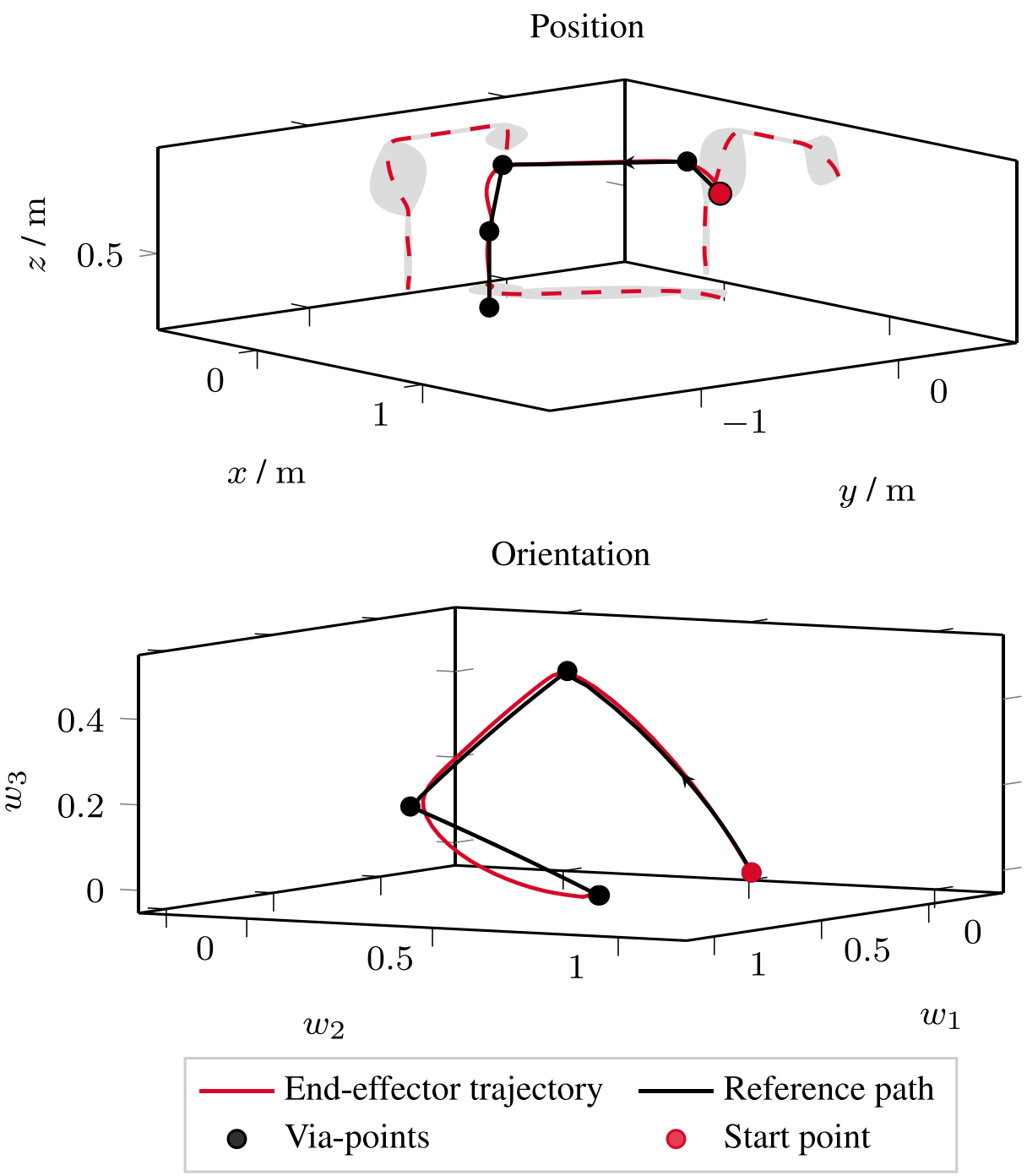

**Figure 17.** End-effector position and orientation trajectories for the object transfer scenario. The projected bounds (light gray areas) for the orthogonal position path error are shown on the Cartesian planes.

BoundMPC formulation throughout the entire trajectory. For CT-MPC, it shows large values for $t \geq 4$ s and large tracking errors at $t \geq 5$ s after the slow down at 4.5 s $\leq t \leq 5$ s. The behavior of CT-MPC is due to the trajectory following formulation and the collision avoidance, as discussed in Section 10.7.

## 10.2. Opening a drawer

*10.2.1. Goal.* This experiment aims to open a cabinet drawer using the gripper at the end-effector. The robot has to thread the gripper finger into the handle and then pull toward the opening direction.

*10.2.2. Setup.* The pose of the cabinet is assumed to be known. The robot has to avoid collisions with the cabinet but not with the handle since contact is necessary to exert a pulling force on the handle. Furthermore, moving the gripper finger into the handle from the top is essential, requires precise positioning, and is crucial for success. The developed framework achieves this by putting a via-point above the handle and descending the reference path into the handle with low orthogonal error bounds to ensure the gripper finger hooks into the handle. Since the reference path is independent of the time parametrization during the motion, the task can be performed successfully at different speeds. The physics simulation was performed in MuJoCo (Todorov et al., 2012).

*10.2.3. Results.* Three robot configurations during the opening of the drawer are shown in Figure 21. The reference paths $\boldsymbol{\pi}_{\mathrm{p}}$ and $\boldsymbol{\pi}_{\mathrm{o}}$ contain three linear segments of which the first positions the end-effector above the handle, the second hooks into the handle and the third segment opens the drawer. In order to avoid collisions and ensure a successful alignment with the handle, the orthogonal path errors $e^{\perp}_{\mathrm{p,proj},i}$ and $e^{\perp}_{\mathrm{o,proj},i}$ ($i = 1, 2$) are bound to be low in the second segment. The same holds for the third segment, which ensures a successful opening. This task is performed with the four different maximum speeds $\overline{\dot{\phi}} \in \{0.25\mathrm{ms}^{-1}, 0.3\mathrm{ms}^{-1}, 0.5\mathrm{ms}^{-1}, 1.0\mathrm{ms}^{-1}\}$ corresponding to the trajectory durations $t_{\mathrm{f}} \in \{3.7\mathrm{s}, 4.5\mathrm{s}, 6\mathrm{s}, 7\mathrm{s}\}$. The trajectories are compared to the Cartesian tracking CT-MPC formulation for which the reference trajectory was chosen to achieve approximately the same trajectory durations $t_{\mathrm{f}}$. A comparison of the end-effector trajectories in the $y$-$z$-plane for different motion planners is shown in Figure 22.

## 10.3. Table wiping

*10.3.1. Goal.* In this experiment, the robot wipes a table with a sponge. The sponge must stays in contact with the table during the wiping.

*10.3.2. Setup.* Each wiping motion is a straight line across the top of the table where the sponge attached to the robot's end-effector is in contact with the table. At the end of each wiping motion, the end-effector lifts off the table and transitions to the following wiping line, starting from the opposite side. Following the straight wiping paths to cover the tabletop area is essential. In order to achieve this, the wiping paths have low error bounds in all position and orientation directions. The first position basis vector $\mathbf{b}_{\mathrm{p},1}$ is the $z$-direction during the wiping motions, and the lower bound is set to zero so that the end-effector will not press too hard onto the table. We compare BoundMPC for this task with the joint-space trajectory tracking JT-MPC formulation to show the advantage of Cartesian paths.

*10.3.3. Results.* The robot is visualized during the table wiping task in Figure 23. A comparison of the end-effector position trajectories in the $x$-$y$-plane is shown in Figure 24 for BoundMPC and JT-MPC. BoundMPC can perform the task successfully while staying on the tabletop. Small deviations from the path are allowed and do not affect the task success. On the contrary, JT-MPC deviates largely from the desired wiping path.

## 10.4. Collision avoidance for entire kinematic chain

*10.4.1. Goal.* This task aims to move an object from a start to a goal pose. The previous scenarios use the bounded reference path to ensure collision-freedom for the end-effector only. However, it is straightforward to also account for collisions with the remaining links of the robot. In this scenario, the entire kinematic chain of the robot is checked for collisions. This is a very demanding task as the space where the robot can turn around is highly constrained due to the obstacles.

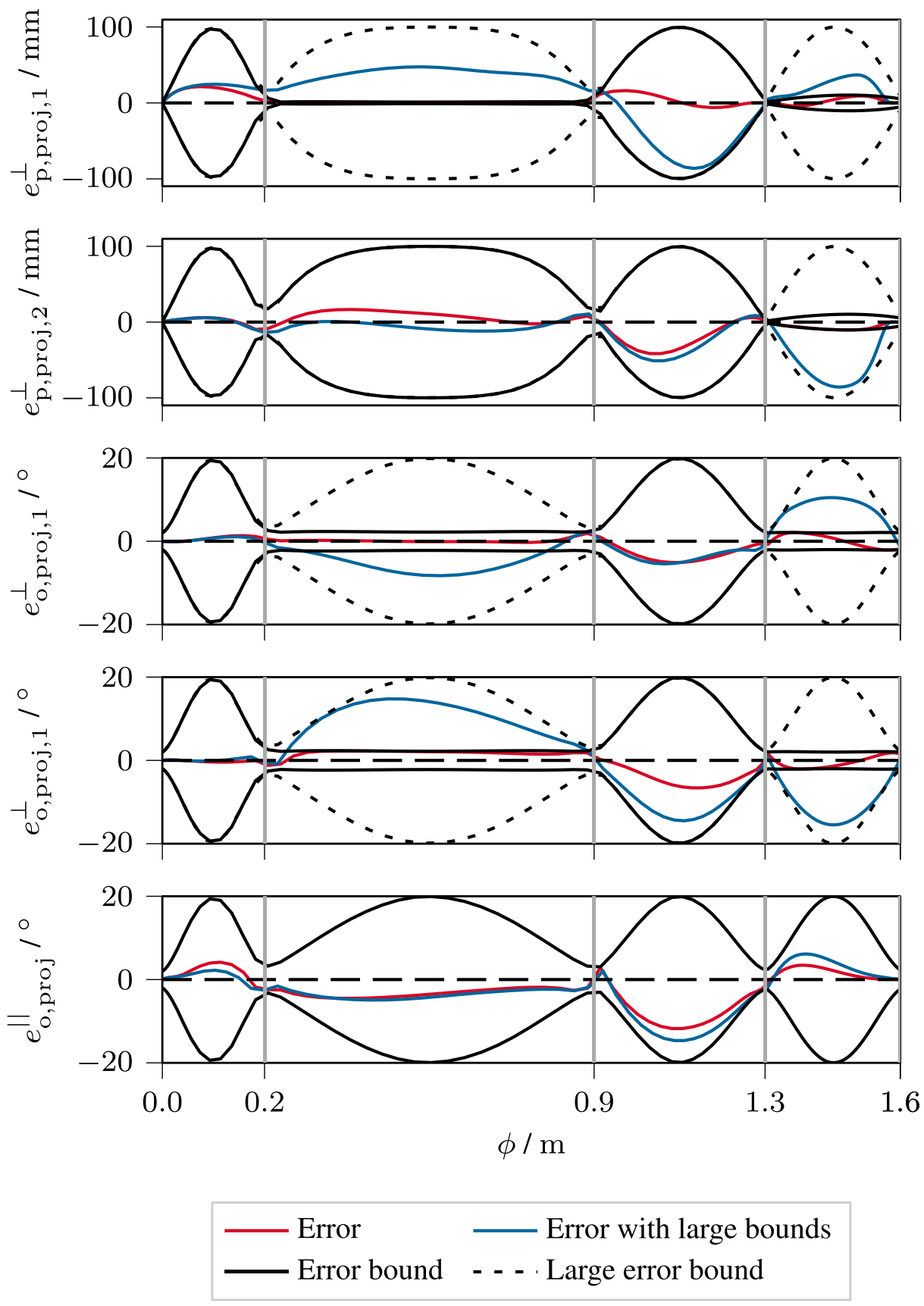

**Figure 18.** Orthogonal path error trajectories in the position and orientation for the object transfer. The upper two plots show the decomposed orthogonal position path errors and the lower three plots the respective orthogonal orientation path errors and the tangential orientation path error. The blue curves show the trajectory when using large error bounds in each segment. The via-points are indicated by vertical gray lines.

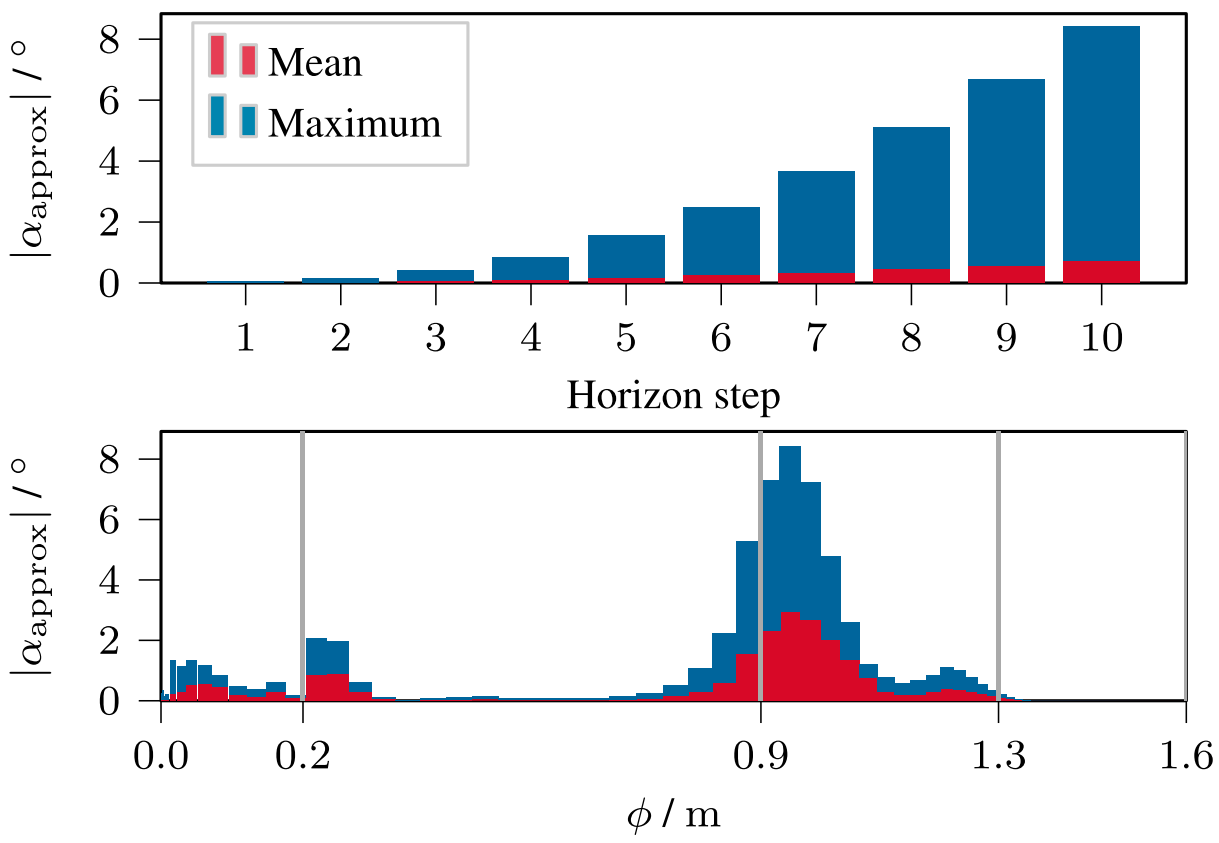

**Figure 19.** Mean and maximum values of the absolute orientation approximation error $|\alpha_{\text{approx}}| = |e^{\perp}_{\text{o, proj}, 1} - \alpha_{\text{true}}|$, where $\alpha_{\text{true}}$ is obtained from (32). The upper bar plot shows the mean and maximum values over the planning horizon of the MPC. The lower plot illustrates these values over the path parameter $\phi$, where each bar's width represents the horizon's geometric length. Vertical gray lines indicate the via-points.

*10.4.2. Setup.* The environment contains the shelves, one in front and one behind the robot, and six additional objects close to the robot's base. The collision avoidance of the end-effector is guaranteed through the collision-free reference path with the collision-free bounding functions. The collision with the remaining links of the manipulator is checked by adding collision avoidance terms similar to Vu et al. (2020), which are also used in the CT-MPC and JT-MPC formulations.

*10.4.3. Results.* Figure 25 visualizes the initial and final robot configurations during the object transfer. In Figure 26, the resulting end-effector trajectories of BoundMPC, CT-MPC, and JT-MPC are shown in the $x$-$y$-plane. BoundMPC is the only framework that successfully performs the task. The CT-MPC gets stuck at about $x = 0$ while turning around due to the obstacles. Furthermore, using JT-MPC, a collision of a robot link with one of the collision spheres is detected, as shown in Figure 27.

In order to further evaluate the performance of BoundMPC in this task, the obstacles were shifted along their $x$-coordinate in the range $-0.4$ m and 0.4 m. The evaluation of the resulting nine experiments is depicted in Table 7. JT-MPC always collides with the environment due to the collisions in the reference trajectory. The CT-MPC formulation gets stuck at obstacles. BoundMPC is the most successful framework, finishing the task successfully each time without collisions.

## *10.5. Object grasp from a table*

*10.5.1. Goal.* This experiment aims at grasping an object from a table. The object pose is changed before the robot can grasp it, causing a replanning toward the new object pose. This replanning shows the strength of the proposed BoundMPC framework.

*10.5.2. Setup.* The grasping pose is assumed to be known. The robot's end-effector must not collide with the table or the obstacle on the table during grasping. These requirements can be easily incorporated into the path planning by appropriately setting the upper and lower bounds for the orthogonal position path error. Note that collision avoidance is only considered for the end-effector. The replanning of the reference path allows a sudden change in the object's pose or a decision to grasp a different object during motion. The newly planned trajectory adapts the position and orientation of the end-effector to ensure a successful grasp. This replanning is performed every time the object changes its position or orientation. To this end, a simple task-specific high-level path planner is implemented. This planner sets the via-points such that the linear paths themselves are collision-free. Then, the bounds are planned based on the distances of the linear segments to the obstacles. First, an example of three object pose changes is constructed, which results in three re-planning instances. Second, a simulation of random object pose changes at random times is performed and the robustness of the approach is evaluated.

**Table 6.** Comparison of BoundMPC with different motion planners for the object transfer scenario, in terms of planning time $t_{\text{plan}}$, geometric path length $l_{\text{path}}$, trajectory duration $t_{\text{traj}}$, and success rate $r_s$. Each planner was run 10 times and the rows report the minimum, mean and maximum values for each planner.

| Planner | $t_{\text{plan}}$ / s | $l_{\text{path}}$ / m | $t_{\text{traj}}$ / s | $r_s$ |
|---|---|---|---|---|
| JT-MPC | 0.01 | 1.75 | 5.0 | 100 % |
| | 0.018 | | | |
| | 0.045 | | | |
| CT-MPC | 0.006 | 2.16 | 8.2 | 100 % |
| | 0.012 | | | |
| | 0.043 | | | |
| BoundMPC | 0.015 | 1.59 | 5.2 | 100 % |
| | 0.025 | | | |
| | 0.038 | | | |

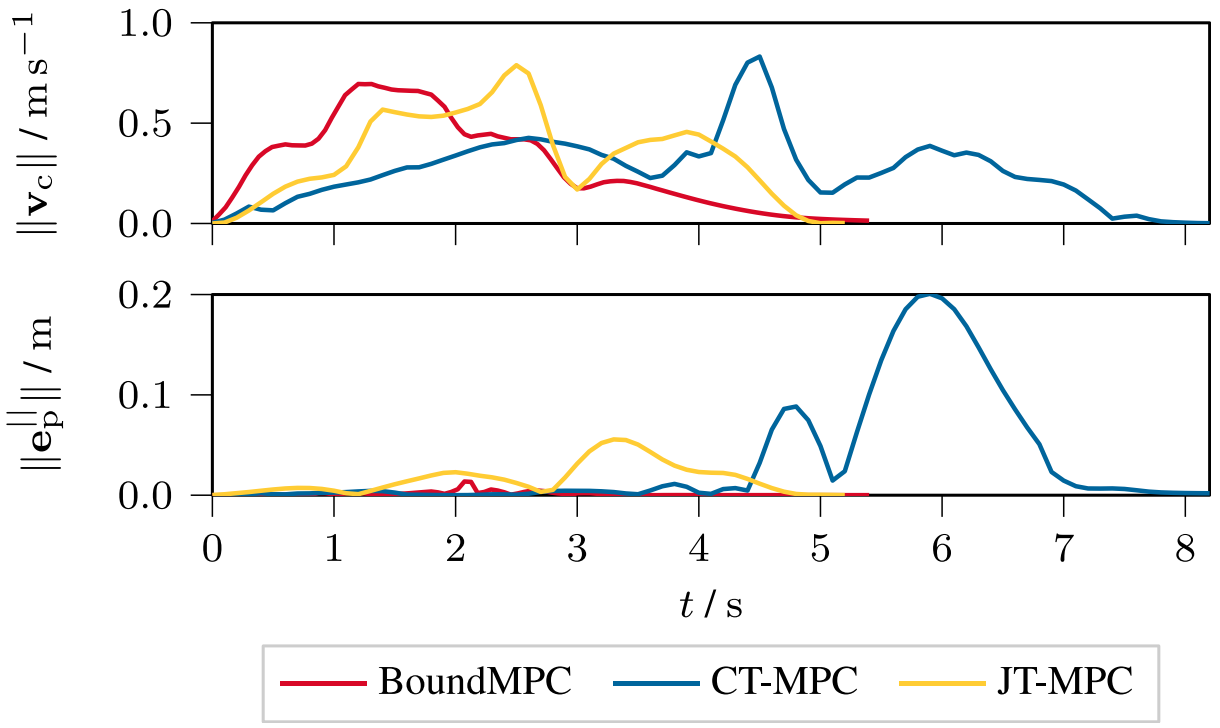

**Figure 20.** Comparison of the normed Cartesian velocity $\|\mathbf{v}_c\|$ (top) and the tangential path error norm $\|\mathbf{e}_p^{\|}\|$ (bottom) for BoundMPC, JT-MPC, and CT-MPC. The reference trajectories for JT-MPC and CT-MPC are chosen to perform the task as quickly as possible. The tangential path error $\mathbf{e}_p^{\|}$ was computed for JT-MPC and CT-MPC by projecting onto the Cartesian normalized reference trajectory velocity.

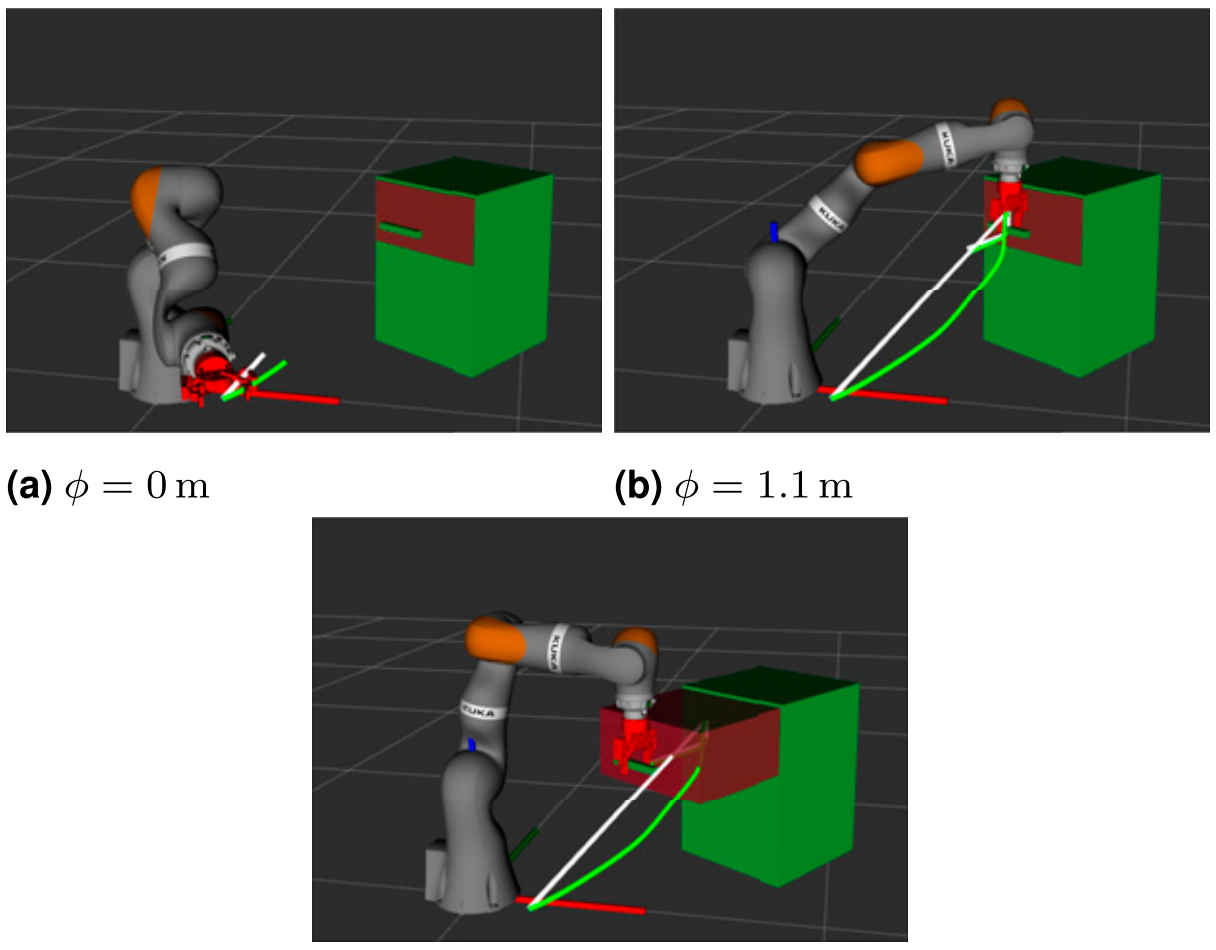

(a) $\phi = 0$ m (b) $\phi = 1.1$ m (c) $\phi = 1.44$ m

**Figure 21.** Visualization of three robot configurations at different time instances along the path. The green box represents a cabinet. The red drawer can be opened by pulling the green handle. The world coordinate system is given by the RGB (xyz) axes at the robot's base. (a) $\phi = 0$ m, (b) $\phi = 1.1$ m, and (c) $\phi = 1.44$ m.

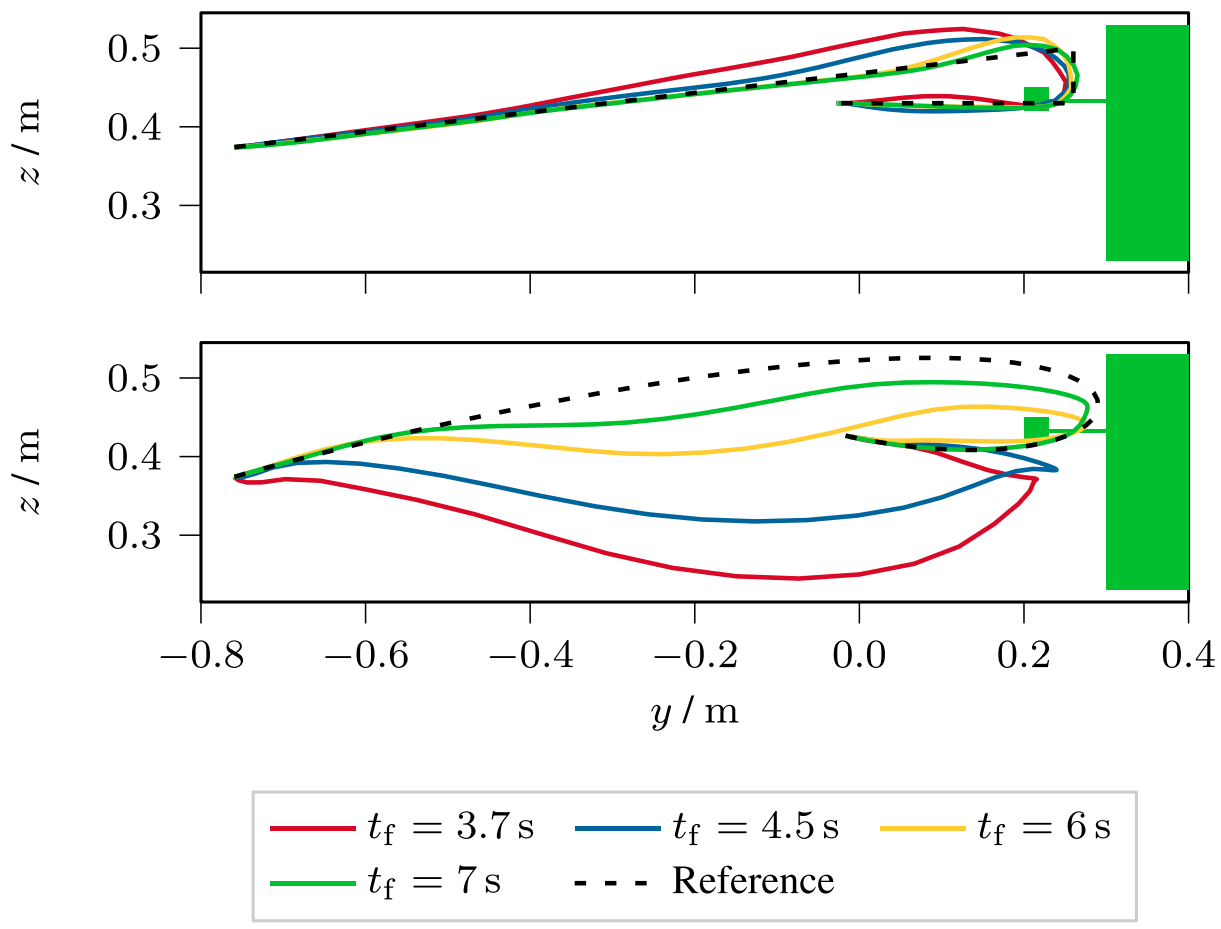

**Figure 22.** End-effector trajectories in the $y$-$z$-plane of BoundMPC (upper) and CT-MPC (lower) for the drawer opening scenario. The cabinet with the attached handle is depicted in green.

*10.5.3. Results.* In order to visualize the robot's motion in the scenario for three replanning instances, Figure 28 illustrates the robot configuration in four time instances of the trajectory. The robot's end-effector follows the reference path and deviates from it according to the bounds between the via-points. The replannings occur at $t_r \in [2$ s, $4.5$ s, $8.5$ s$]$. At each replanning time instance $t_r$, the optimization has to adapt the previous solution significantly due to the sudden change of the reference path.

The error bounds $e_{\text{p,u},m}$, $e_{\text{p,l},m}$, and $e_{\text{o,u},m}$, $e_{\text{o,l},m}$, $m = 1, 2$, are chosen such that collisions with the obstacle on the table, the table itself, and the object to be grasped are avoided. This is visualized in Figure 29 for the trajectory between the second and the third replanning ($4.5 \text{ s} \le t \le 8.4$ s). At this time, the robot's end-effector must pass over the obstacle on the table, approach the object without collision, and then perform the grasp. The error bounds at $t = 4.5$ s allow large deviations to simplify the replanning and ensure feasibility.

During $4.5 \text{ s} \le t \le 8.4$ s, the projected position path error $e^{\perp}_{\text{p, proj}, 1}$ in Figure 29 is bounded to avoid collisions. In the first segment, $\mathbf{b}^{\text{T}}_{\text{p}, 1, 1}$ is chosen to point toward the closest obstacle, either the table or the obstacle on the table.

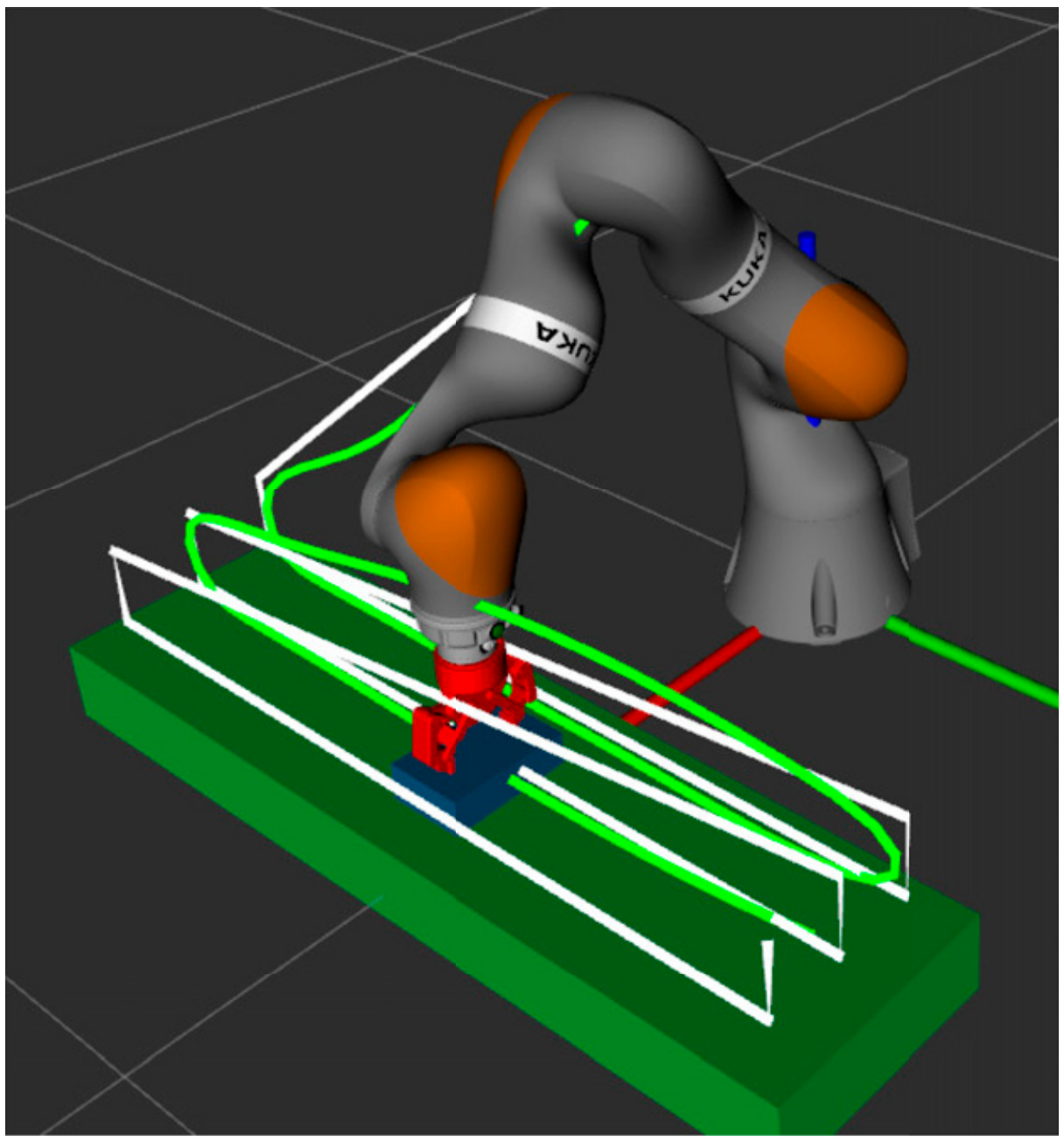

**Figure 23.** Visualization of the robot during the table wiping task: The blue box in the robot's gripper represents the sponge used to wipe the green table along the white reference path. The world coordinate system is given by the RGB (xyz) axes at the robot's base.

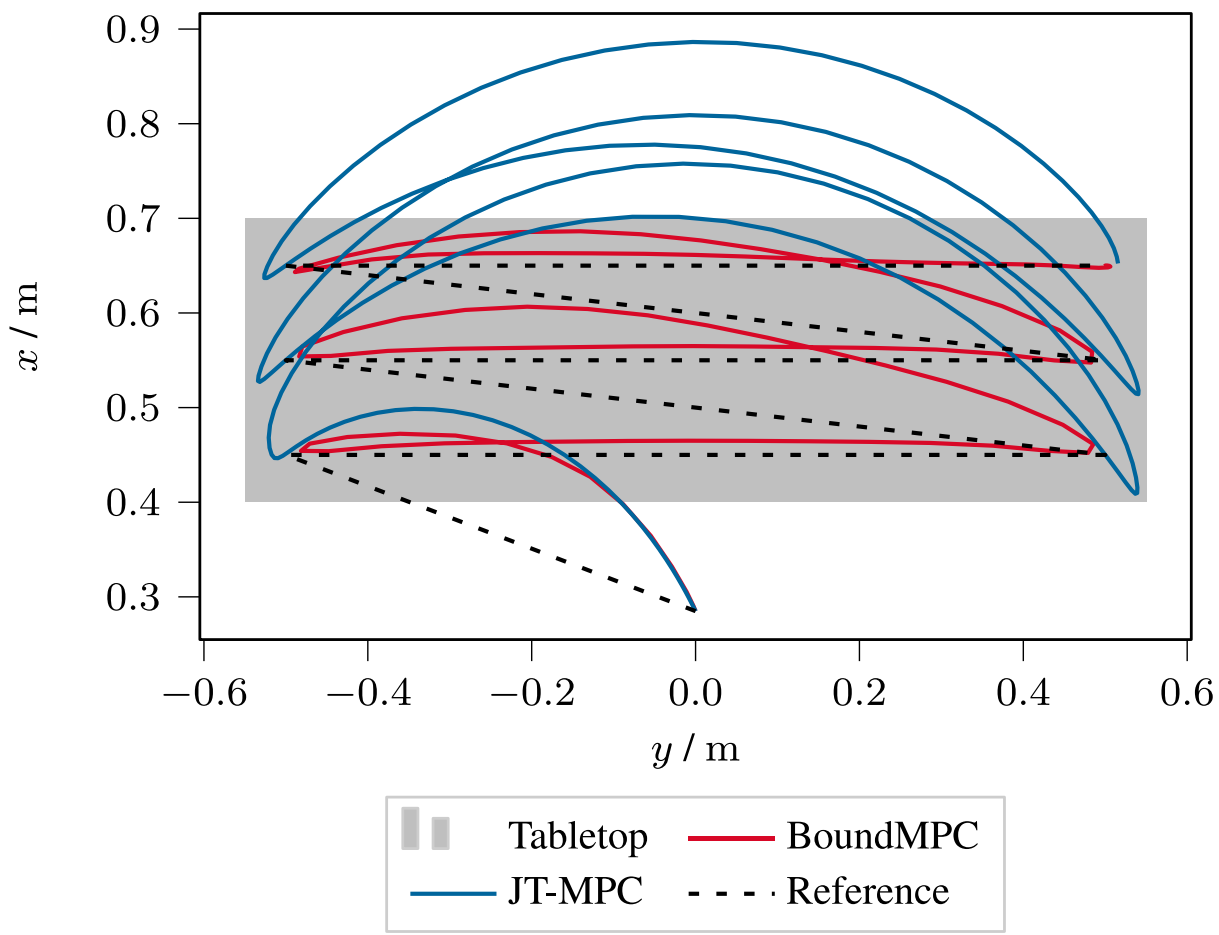

**Figure 24.** End-effector positions trajectories in the $x$-$y$-plane during the table wiping task for the BoundMPC formulation and JT-MPC: The Cartesian reference of the BoundMPC formulation is also shown. The wiping direction is from left to right and the diagonal lines of the reference path are the transitions between the wiping movements.

However, as these are further away from the reference path than 150 mm, the error $e^{\perp}_{\mathrm{p,proj},1}$ is bounded by ± 150 mm. In the second segment, the error $e^{\perp}_{\mathrm{p,proj},1}$ is bounded from below along the basis directions $\mathbf{b}^{\mathrm{T}}_{\mathrm{p},1,2} = [0, 0, 1]$ to avoid collisions with the obstacle. In the third segment, the basis vector $\mathbf{b}_{\mathrm{p},1,3}$ is chosen to point from the path toward the object. The fourth segment ($t \geq 8.1$) is the approach segment, in which the object is approached correctly to ensure a successful grasp. In this segment, the position path error in the basis direction $\mathbf{b}_{\mathrm{p},1,4}$ is bound to be low. However, the path is replanned at 8.5 s before this grasp happens. The orientation path errors are chosen to be symmetrically bounded by 20° to avoid collisions.

Furthermore, Figure 30 shows the path parameter $\phi$ and the speed along the path $\dot{\phi}$ during the trajectory. At the replanning points, the path state jumps according to (61), and the velocity $\dot{\phi}$ becomes negative, for example, at the third replanning at $t = 8.5$ s. A video of the experiment can be found at https://www.acin.tuwien.ac.at/42d0/.

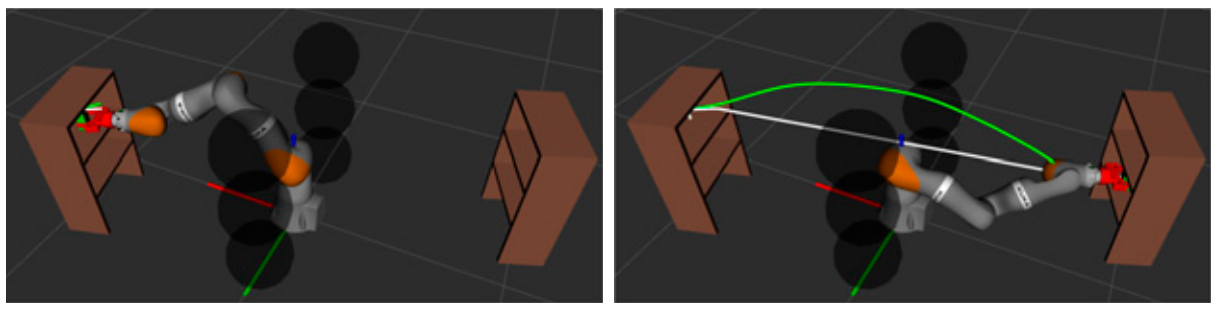

**(a)** $\phi = 0\,\mathrm{m}$ **(b)** $\phi = 2\,\mathrm{m}$

**Figure 25.** Visualization of two robot configurations at different times along the path: The green box is the object to be transferred between the two shelves. Six obstacles are shown as black spheres close to the robot's base. The RGB (xyz) axes give the world coordinate system at the robot's base. (a) $\phi = 0$ m and (b) $\phi = 2$ m.

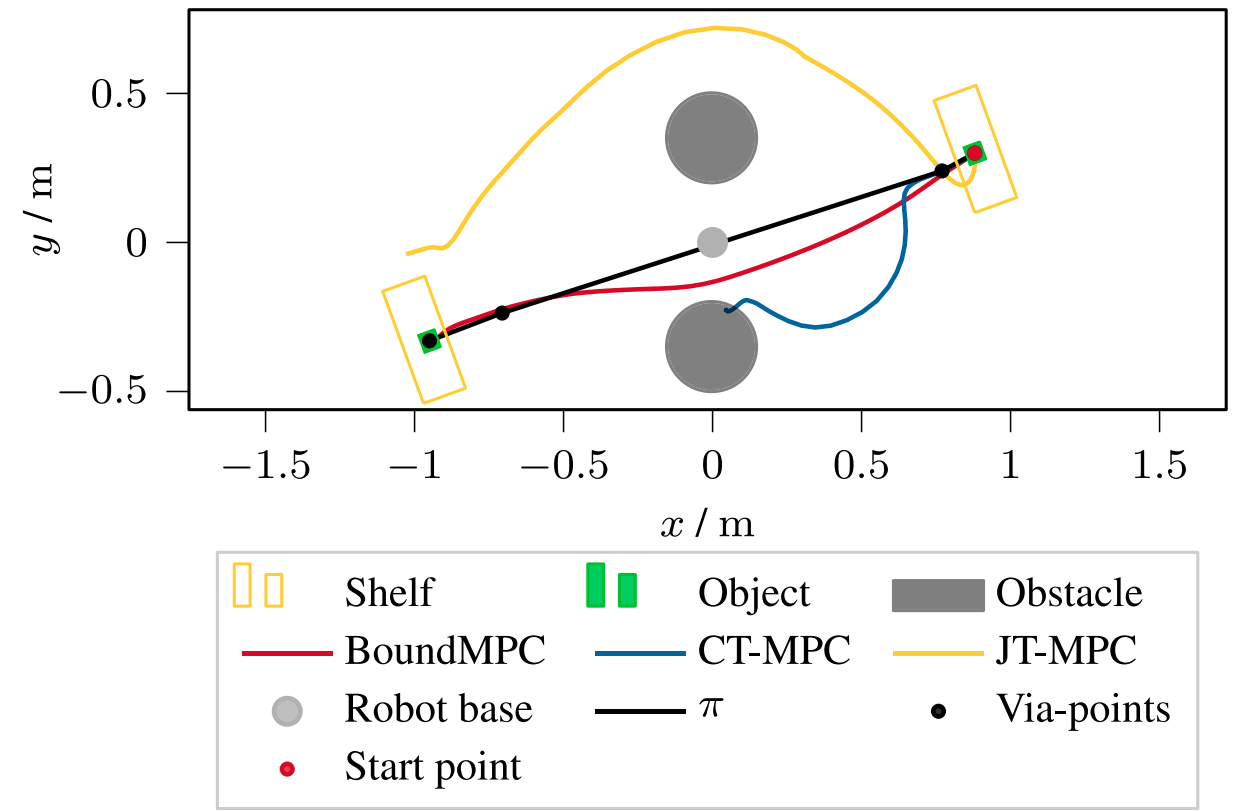

**Figure 26.** End-effector trajectories in a top-down perspective ($x$-$y$-plane) for the object transfer between two shelves: The trajectories of BoundMPC, CT-MPC, and JT-MPC are shown.

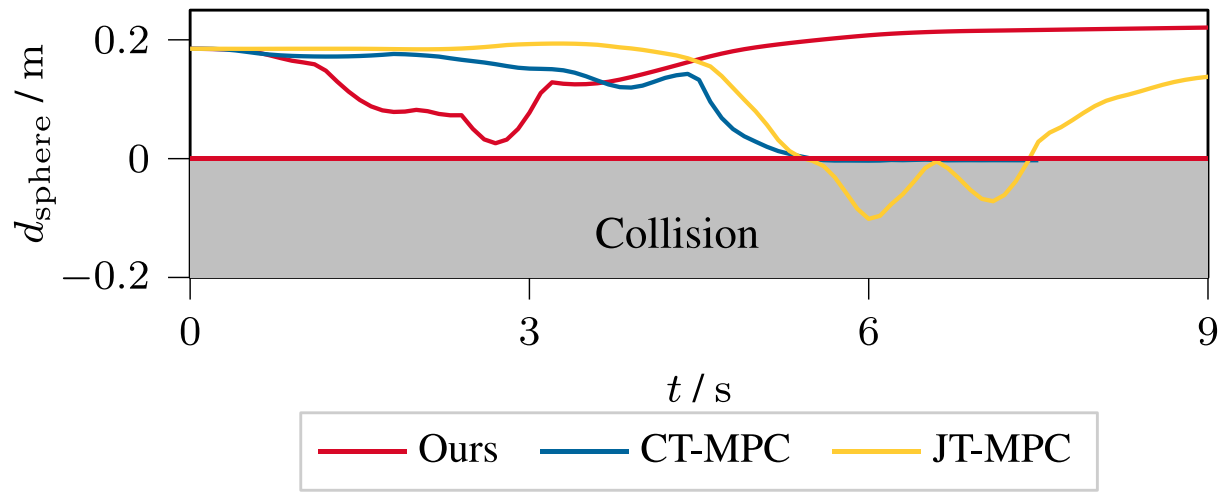

**Figure 27.** The minimum distance $d_{\mathrm{sphere}}$ of any robot link to a collision sphere: Each trajectory has to fulfill $d_{\mathrm{sphere}} > 0$ m to ensure collision freedom.

**Table 7.** Comparison of BoundMPC with offline and online planners for the shelf transfer task with collision avoidance of the robot's full kinematic chain: Nine experiments are compared, in which the six obstacles in Figure 25 are shifted along the $x$-axis between $-0.4$ m and 0.4 m with a step length of 0.1 m.

| | JT-MPC | CT-MPC | BoundMPC |
|---|---|---|---|
| Successes | 0/9 | 2/9 | 9/9 |
| Collisions | 9/9 | 1/9 | 0/9 |
| Stuck at joint limits | 0/9 | 1/9 | 0/9 |
| Stuck at obstacles | 0/9 | 7/9 | 0/9 |

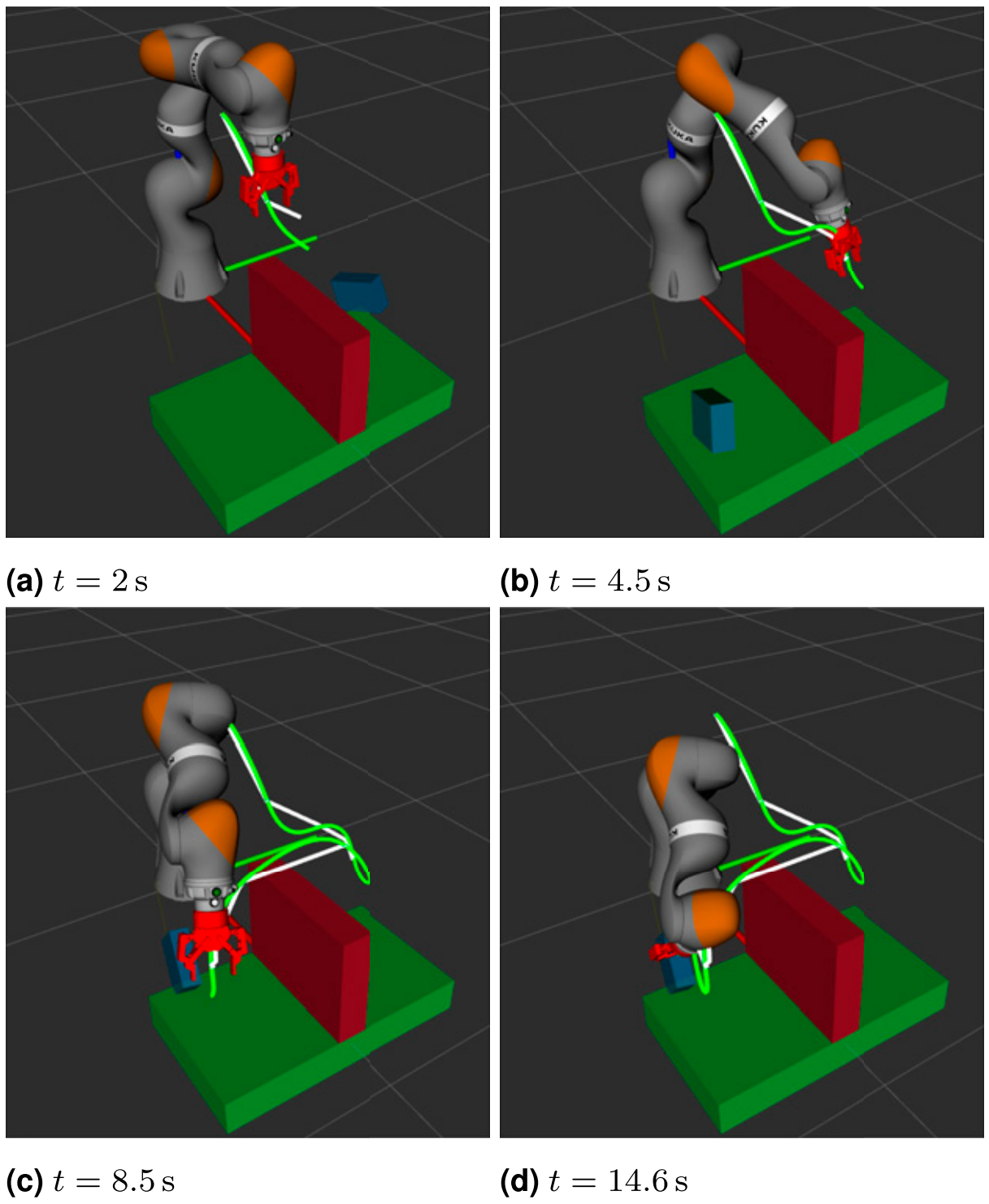

**Figure 28.** Visualization of the robot's configuration at different time instances along the path. The end-effector in red is a gripper attached to the last robot link. The blue box represents the object to grasp, which is above a green table. The red wall is an additional obstacle to be avoided. The first three time instances represent the time of a replanning, where the object's pose changes. The green and white lines show the end-effector's planned trajectory and the reference path. The world coordinate system is given by the RGB (xyz) axes at the robot's base. (a) $t = 2$ s, (b) $t = 4.5$ s, (c) $t = 8.5$ s, and (d) $t = 14.6$ s.

The framework was further tested in the scenario of Figure 28 with randomly sampled object positions

$$\mathbf{p}_{\text{object, random}} = \begin{bmatrix} 0.6 \pm 0.15\text{m} \\ 0.0 \pm 0.35\text{m} \\ 0.1 \pm 0.1\text{m} \end{bmatrix} \tag{65}$$

and object orientations, given by Euler angles in convention ZYZ in the range

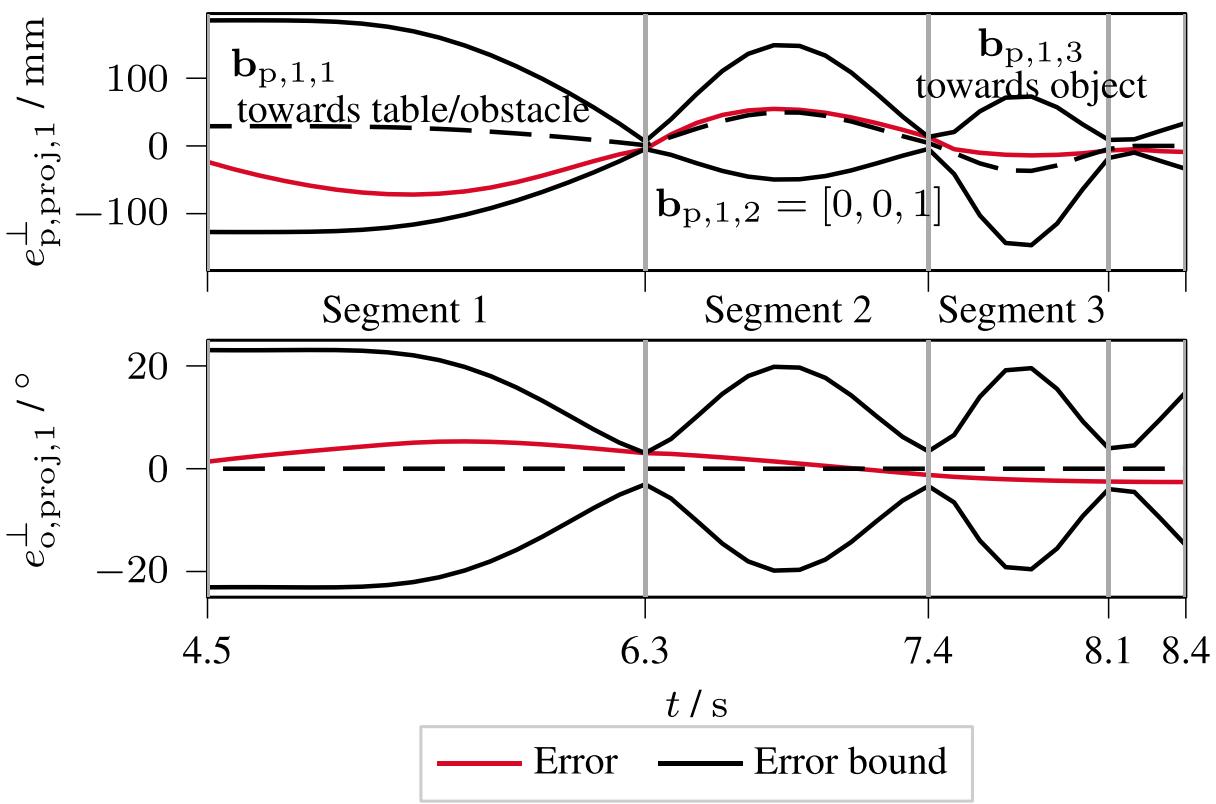

**Figure 29.** Path error trajectories of the orthogonal position path error $e^{\perp}_{\mathrm{p,proj,1}}$ and the orthogonal orientation path error $e^{\perp}_{\mathrm{o,proj,1}}$ for the object grasp between the second and third replanning at 4.5 s $\leq t \leq$ 8.4 s. The via-points are indicated by vertical gray lines.

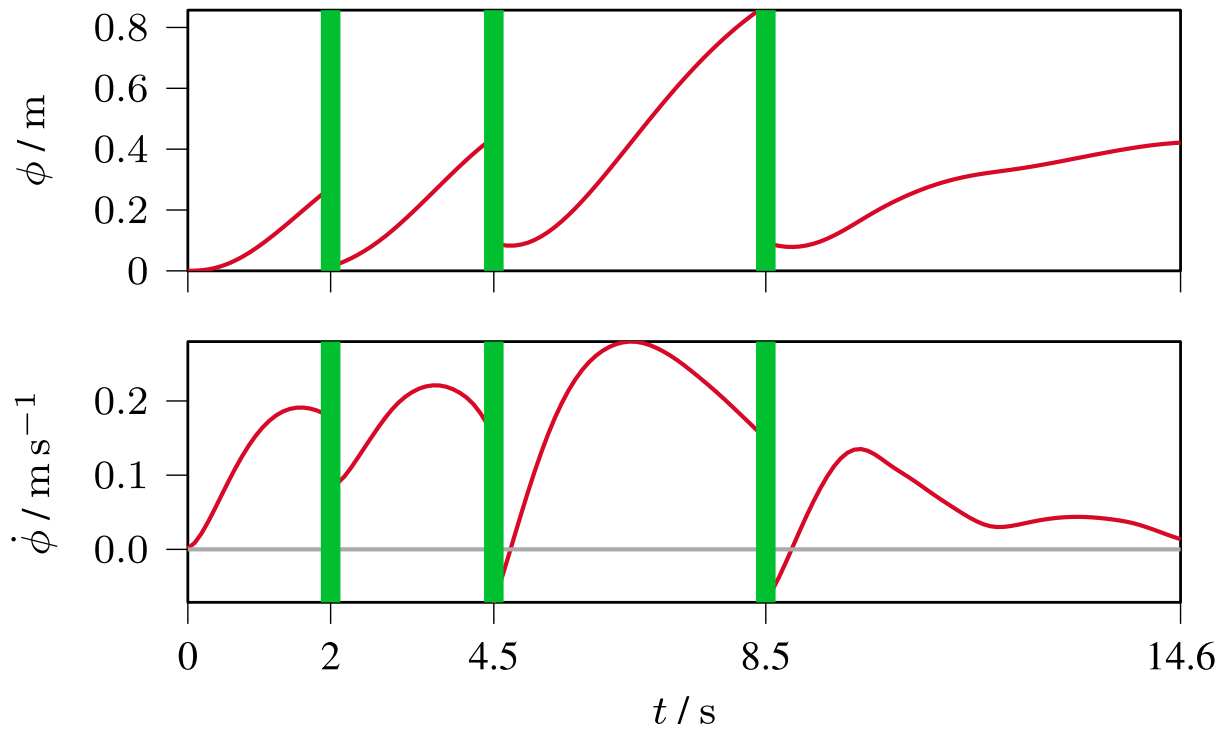

**Figure 30.** Trajectories of the path parameter $\phi$ and its time derivative $\dot{\phi}$ for the object grasping experiment with replanning. The time instances of path replanning are indicated by vertical green lines.

$$\boldsymbol{\theta}_{\text{euler, random}} = \begin{bmatrix} \pm 45° \\ \pm 45° \\ \pm 45° \end{bmatrix}. \tag{66}$$

One experiment was to perform permanent replanning every 0.5 s to 1.5 s until the MPC did not find a solution. This experiment was conducted 10 times. On average, the MPC performed 95 replanning steps per experiment, corresponding to a failure rate of about 1.1 % of all replanning steps. The proposed framework stayed within the path error bounds to ensure collision-freedom by the design of the reference paths. Hence, no collisions between the robot and the table, the obstacle on the table, or the object were observed.

In order to compare this performance to the state-of-the-art, 50 random poses were sampled according to the procedure mentioned above. JT-MPC, CT-MPC, and BoundMPC were used to plan trajectories to each of the sampled poses. Once the pose was reached, the motion to the next sampled pose was planned. The comparison metrics and results are shown in Table 8.

**Table 8.** Comparison of BoundMPC with different motion planners for the object grasp scenario. In total, 50 random object poses were evaluated, and it was ensured that all poses were reachable without collisions. The minimum, mean and maximum values are reported for each planner.

| Planner | $t_{\text{plan}}$/s | $l_{\text{path}}$/m | $t_{\text{traj}}$/s | $r_{\text{s}}$ |
|---|---|---|---|---|
| JT-MPC | 0.01 | 0.16 | 1.8 | 100 % |
| | 0.02 | 0.71 | 4.1 | |
| | 0.05 | 1.57 | 6.0 | |
| CT-MPC | 0.002 | 0.13 | 2.3 | 100 % |
| | 0.02 | 0.71 | 5.1 | |
| | 0.09 | 1.52 | 8.2 | |
| BoundMPC | 0.008 | 0.15 | 2.6 | 100 % |
| | 0.03 | 0.74 | 5.1 | |
| | 0.09 | 1.60 | 6.6 | |

## 10.6. *Transfer of a filled open container*

*10.6.1. Goal.* An open container filled with a liquid is grabbed using a two-finger gripper and is to be placed onto a table on the opposite side of the robot's workspace. In order to avoid spillage, the container needs to remain upright. A via-point above the table and an approach via-point are used to perform the motion.

*10.6.2. Setup.* The requirement for the container to stay upright translates easily to the orientation path error bounds of the end-effector. For BoundMPC, a maximum of ±2.5° orientation path error is allowed for the rotations around the $x$ and $y$ axis, that is, limiting $e^{\perp}_{\text{o, proj}, i}$ ($i$ = 1, 2), which correspond to the RPY angles $\alpha$ and $\gamma$ defined in (41). The $z$-axis represents the upright direction, thus, a rotation around this axis, represented by $\beta$, is not essential for spilling. The position path errors $e^{\perp}_{\text{p, proj}, i}$ have large bounds to allow a successful motion. The CT-MPC formulation is also applied to this problem, focusing on position tracking (p) or orientation tracking (o) by changing the weights in the cost function. Furthermore, BoundMPC is compared to its path-tracking variant BoundMPC-PT.

**Remark.** Note that, the liquid inside the container also moves due to the forces exerted by robot's motion. Hence, the given requirement is only an approximation.

*10.6.3. Results.* The start and end configurations of the robot are depicted in Figure 31. Figure 32 shows the trajectories of the RPY angles $\alpha$, $\beta$, and $\gamma$ for BoundMPC and the two CT-MPC formulations. The BoundMPC framework keeps the RPY angles $\alpha$ and $\gamma$ within the specified bounds shown as gray areas and successfully reaches the goal state ($d_{\text{goal}} \rightarrow 0$). The CT-MPC formulations reach the goal point but struggle to find good solutions. When the cost function is parameterized for position tracking (p), the upright requirement is not fulfilled. However, the requirement is fulfilled for orientation tracking (o), but the trajectory's smoothness, indicated by the norm of the Cartesian acceleration $\mathbf{a}_{\text{c}}$, is very poor. The path-tracking formulation BoundMPC-PT cannot fulfill the task as it gets stuck (see Figure 32), where $d_{\text{goal}}$ does not reach zero.

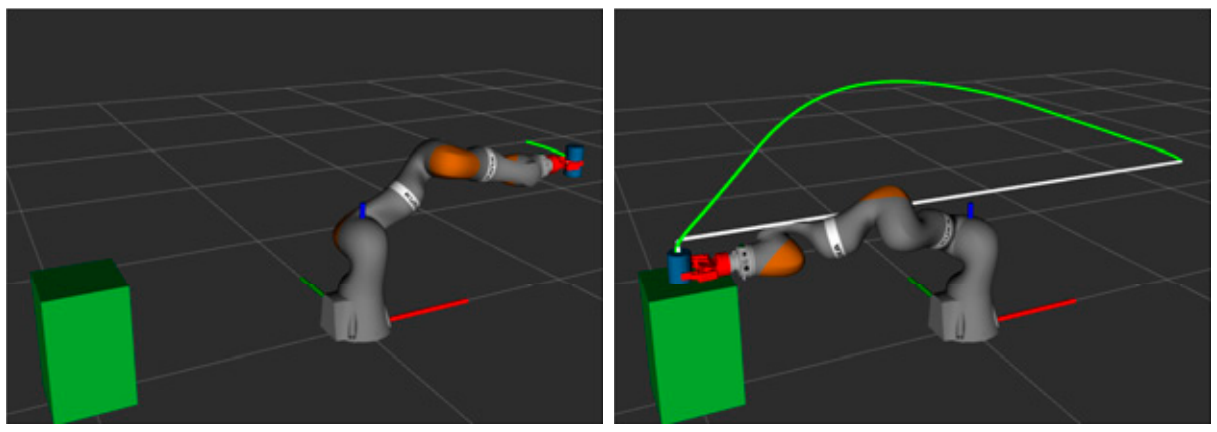

**(a)** $\phi = 0\,\text{m}$ **(b)** $\phi = 2.1\,\text{m}$

**Figure 31.** Visualization of two robot configurations in different time instances along the path. The blue cylinder in the robot's gripper represents the filled open container to be placed onto the green table. The world coordinate system is given by the RGB (xyz) axes at the robot's base. (a) $\phi$ = 0 m and (b) $\phi$ = 2.1 m.

Randomization is applied to this problem where the container placement pose is randomized around a ±90° arc around the position straight behind the robot. The new goal pose is provided to the planner randomly during the approach of the initial pose to investigate online replanning. The results of 20 experiment executions are depicted for BoundMPC and the CT-MPC framework parametrized to minimize the orientation errors in Table 9.

## 10.7. *Discussion*

The evaluated experiments show the versatility and the strengths of the proposed BoundMPC formulation which are discussed in detail in this section.

*10.7.1. General.* BoundMPC can perform the tasks in all six scenarios well and can compute the joint-space trajectories online as shown, for example, in Sections 10.1 and 10.5. The orientation approximation Figure 19 used by BoundMPC does not restrict the performance throughout the experiments. Specifically, the approximation error $\alpha_{\text{approx}}$ in Figure 19 is low for the first horizon step, and the optimization is performed at each time step in a receding horizon manner. Thus, the applied linear approximation (39) is valid for finding good orientation trajectories.

*10.7.2. Collision avoidance.* BoundMPC uses bounded path errors for the collision avoidance of the end-effector, whereas CT-MPC and JT-MPC use cost terms in the MPC objective function. The advantage of the proposed formulation becomes clear in Section 10.1. The formulation with collision-free bounds is independent of the obstacles's amount, shape, and size, making even highly constrained scenarios easily accessible. This way, the obstacle-free space is parameterized instead of the obstacle-occupied space as for CT-MPC and JT-MPC. This leads to small bounds at the end of the trajectory to place the object between the walls (see Figure 18), but does not impede performance. Although the reference trajectory guides the CT-MPC formulation, it struggles to find a collision-free trajectory for the end-effector, resulting in a longer path length $l_{\text{path}}$. Furthermore, for the transfer of an object between two shelves in Section 10.4 the collision avoidance of the entire robot's kinematic chain is demonstrated. In this experiment the collision-free reference path guides the robot with the bounding functions toward the goal. CT-MPC and JT-MPC struggle in this scenario because they lack the guiding of the bounding functions. For JT-MPC, a trade-off between minimizing the trajectory error and collision avoidance leads to an unsafe trajectory of the robot, as seen in Figure 27, because the joint-space reference trajectory guides the end-effector around the outside of the spheres, as seen in Figure 26, which is impossible without collisions.

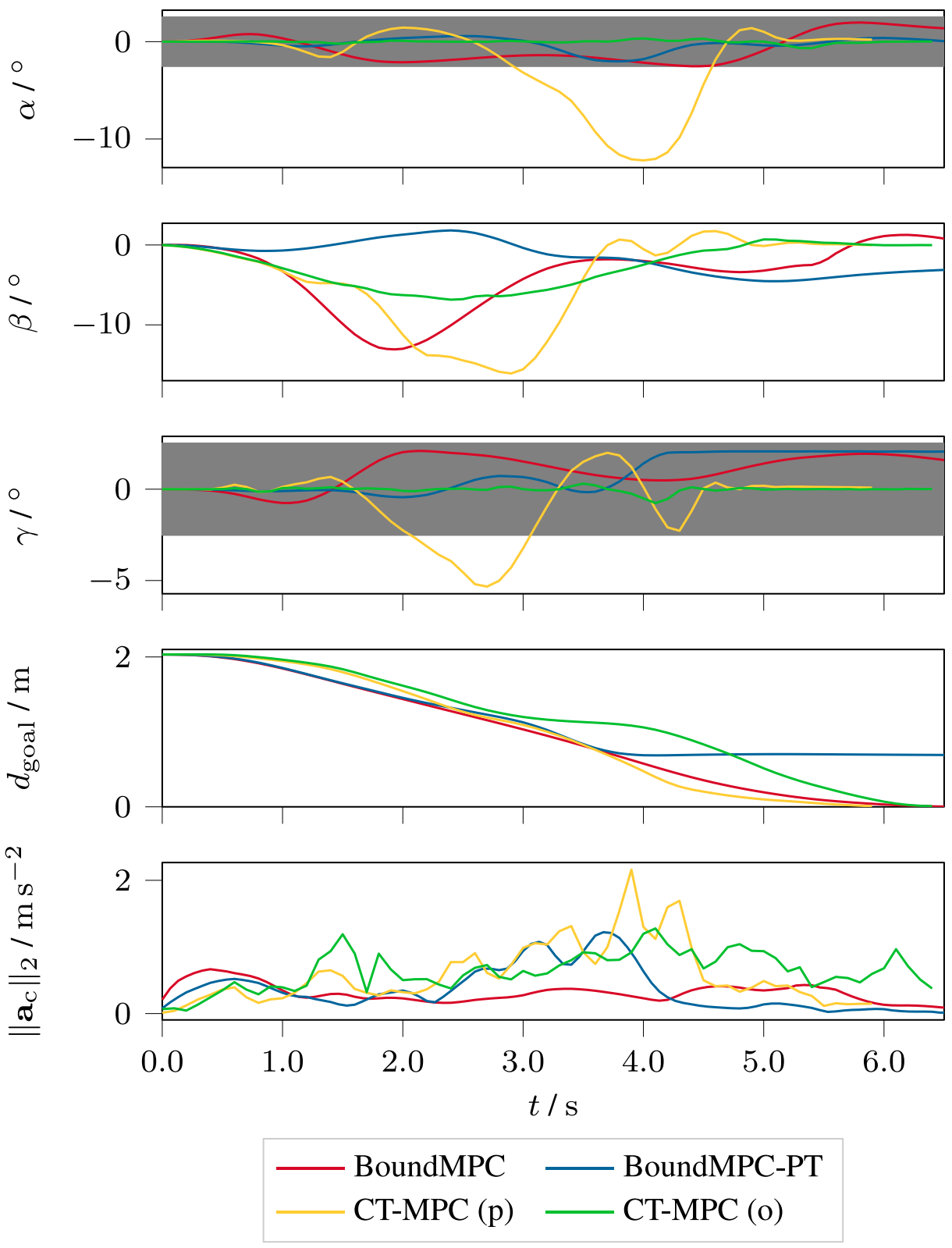

**Figure 32.** End-effector orientation trajectories given as RPY error angles defined in (31a) for the filled container transfer scenario. The angles $\alpha$ and $\gamma$ must stay within the gray area to fulfill the upright requirement. The trajectories of BoundMPC are compared to CT-MPC parametrized for either position (p) or orientation (o) tracking. The fourth plot shows the distance $d_{\text{goal}}$ to the goal point and the last plot depicts the norm of the Cartesian acceleration $\mathbf{a}_{\text{c}}$ as a measure of trajectory smoothness.

**Table 9.** Comparison of the BoundMPC and CT-MPC framework parametrized for orientation following. The metrics are the maximum RPY angle deviations of the upright $\alpha$ and $\gamma$ and the mean norm of the Cartesian acceleration $\|\mathbf{a}_{\text{c}}\|$, where the maximum and mean are taken over the trajectory. The minimum, mean and maximum values over 20 tries are provided for each metric.

| Formulation | max $\alpha$ | max $\gamma$ | mean$\|\mathbf{a}_{\text{c}}\|$ |
|---|---|---|---|
| BoundMPC | 1.35 | 1.85 | 0.30 |
| | 1.98 | 2.02 | 0.33 |
| | 2.22 | 2.43 | 0.38 |
| CT-MPC (o) | 0.70 | 1.35 | 0.69 |
| | 3.15 | 4.47 | 0.81 |
| | 7.62 | 9.47 | 1.08 |

*10.7.3. Path following.* The path-following formulation used by BoundMPC exhibits advantages over the trajectory tracking used by JT-MPC and CT-MPC. It makes it optimizing the speed along the path possible such that following a complicated reference path causes the end-effector to slow down for the necessary joint movements. A Cartesian reference path or trajectory does not guarantee that the robot's end-effector can follow it perfectly. Especially the coupling of position and orientation makes it complicated to plan suitably for the end-effector. When planning a trajectory, the time parametrization causes additional problems since the nonlinear robot kinematics must be considered. In the object transfer scenario in Section 10.1, BoundMPC slows down at $t \geq 3.0$ s to approach the goal smoothly and slowly, as depicted in Figure 20. In contrast, CT-MPC cannot follow the reference trajectory due to the issues mentioned above, forcing the robot to slow down at $4.5 \text{ s} \leq t \leq 5$ s. However, since the reference trajectory progresses, the tracking error in the cost function increases, leading to large Cartesian velocities to decrease the tracking error at $t \geq 5$ s. This issue is less pronounced for the joint-space planner JT-MPC because of its large planning horizon. However, making JT-MPC faster causes the end-effector to collide with obstacles. These collisions are caused by a trade-off between collision avoidance and minimizing the tracking error, which the path-following formulation proposed in this paper avoids. This effect is also observed in Section 10.4.

*10.7.4. Bounded path errors.* As discussed above, the bounded path errors are used for collision avoidance but are also important for task success. Some scenarios require certain path error bounds to finish the task successfully. In Section 10.2, BoundMPC finds suitable trajectories for all trajectory durations $t_{\text{f}}$ and performs the task successfully (see Figure 22). For long trajectory durations ($t_{\text{f}} \geq 6$ s), CT-MPC also manages to solve the task. However, if the durations become small, for example, $t_{\text{f}} \leq 4.5$ s, CT-MPC struggles to follow the reference and fails at opening the drawer. CT-MPC reduces tracking errors by shortcutting, which leads to task failure. The BoundMPC formulation stays within the path error bounds and adapts the path velocity $\dot{\phi}$ accordingly to solve the task. This allows BoundMPC to perform this task 2.3 s faster and more robustly than CT-MPC. These bounded path errors are further used in Section 10.3 to follow a straight line on the table but deviate from the path during the transition segments. The upright constraint in Section 10.6 further demonstrates how orientation constraints can be encoded. In the container-transfer experiment, BoundMPC guarantees low-angle errors for the upright direction, but this is not true for CT-MPC. The latter often violates the upright requirement of $|\alpha|, |\beta| < 2.5°$. In Section 10.4, the bounded path errors keep the end-effector in the space between the collision spheres, which guides it successfully toward the goal without collisions. JT-MPC and CT-MPC lack this guidance and struggle with the task.

*10.7.5. Purposeful path deviations.* The path error in BoundMPC is not minimized but bounded so that the robot can deviate from the path when necessary. This is advantageous in avoiding local minima due to the robot's kinematic. As seen in Section 10.6 and Figure 32, BoundMPC-PT gets stuck during the container transfer because the reference trajectory is suboptimal and hard to follow precisely. As the tracking errors $\mathbf{e}_p$ and $\mathbf{e}_o$ are minimized, BoundMPC-PT refrains from leaving the path, which would be necessary in this context. This highlights that the error bounds around the reference path enhance performance and lead to successful task execution. This shows why Astudillo et al. (2022a) require pre-optimized reference paths for their path-tracking formulation. In the same scenario, the deviations do not create additional costs for BoundMPC such that smoother trajectories are obtained than those of CT-MPC.

*10.7.6. Cartesian reference paths.* The Cartesian reference paths in BoundMPC show advantages over planners working in the joint space. In the table-wiping task in Section 10.3, the robot has to follow a reference path on the table. It is not possible to finish this task successfully with the JT-MPC formulation as the joint-space reference trajectory does not correspond to a wiping motion, emphasizing the need for Cartesian reference paths. In the transfer of a filled open container in Section 10.6, JT-MPC cannot perform this task without additional cost terms or constraints, as the joint-space reference trajectory has no notion of orientation. However, BoundMPC can perform this task successfully due to the Cartesian reference paths.

*10.7.7. Path error decomposition.* The BoundMPC formulation decomposes the position and orientation path error into tangential and orthogonal directions. This is helpful in Section 10.6 since it becomes possible to bound only the relevant error directions for the upright constraint but leave the remaining directions with large allowed path errors. Table 9 and Figure 32 show that smoother and more successful trajectories are obtained due to this advantage over CT-MPC. A major issue for CT-MPC is that the end-effector's reference trajectory is not perfectly followable. Thus, CT-MPC has to deviate in either position or orientation, while lacking information on which direction deviations are allowed. If the position tracking error is weighted stronger than the orientation tracking error, the MPC violates the upright requirement. If the orientation is weighted stronger, large non-smooth position deviations occur since the position error grows large, and eventually, the MPC tries to decrease it back to zero. Parametrizing CT-MPC for orientations minimizes all orientation errors. Hence, the error in $\beta$ is also minimized, which is not critical for spilling. The decomposition is also helpful in Section 10.4 as the position path error is constrained stronger in the direction that points toward the spheres, thus favoring a deviation in the other basis directions. Due to this, BoundMPC successfully solves the task.

*10.7.8. Replanning.* BoundMPC can perform robust online replanning to react to environmental changes and avoid collisions with the obstacles. Global trajectory planners, for example, Jankowski et al. (2023) and Marcucci et al. (2023), that plan the entire trajectory from the start to the goal pose are generally not real-time capable and therefore not applicable to this scenario. The replanning capabilities are demonstrated in Section 10.5. A simple task-specific higher-level path planner was implemented for this scenario, but it did not consider the robot's kinematics or dynamics. For complex scenarios a more general and sophisticated planner is required. The design of such a planner is part of our future work. In the comparison in Table 8, CT-MPC performs as well as BoundMPC in this environment, and JT-MPC slightly outperforms it. This is expected as the Cartesian and joint-space reference trajectories are collision-free in most cases However, the maximum planning times $t_{\text{plan}}$ are the same for BoundMPC and CT-MPC, so both concepts are equivalent in this context. Notably, some poses are hard to reach for the manipulator such that the trajectory tracking issues discussed above also apply here, that is, CT-MPC offers no considerable advantage in this application. JT-MPC is fine with hard-to-reach poses as it plans in the joint space. The replanning experiments for the scenario with the filled container in Section 10.6 show that BoundMPC can robustly replan even with task-specific constraints which is not possible for CT-MPC, as seen in Table 9.

## 11. Conclusion

This work proposes a novel model-predictive trajectory planning for robots in Cartesian space, called *BoundMPC*, which systematically considers via-points, tangential and orthogonal error decomposition, and asymmetric error bounds. The joint jerk input is optimized to follow a reference path in Cartesian space. The motion of the robot's end-effector is decomposed into a tangential motion, the path progress, and an orthogonal motion, which is further decomposed geometrically into two basic directions. A novel way of projecting orientations allows a meaningful decomposition of the orientation error using the Lie theory for rotations. Additionally, the particular case of piecewise linear reference paths in orientation and position is derived, simplifying the optimization problem's online calculations. Online replanning is crucial to adapt the considered path to dynamic environmental changes and new goals during the robot's motion.

Simulations and experiments were carried out on a 7-DoF KUKA LBR iiwa 14 R820 manipulator, where the flexibility of the new framework is demonstrated in six distinct scenarios.

First, the transfer of a large object through a slit, where lifting of the object is required, is solved by utilizing via-points and specific bounds on the orthogonal error in the position and orientation reference paths, which are synchronized by the path parameter. Second, an object has to be grasped from a table, the commanded grasping point changes during the robot's motion, and collision with other objects must be avoided. This case shows the advantages of systematically considering asymmetric Cartesian error bounds and the dynamic online replanning. Further scenarios show the drawer opening, the transfer of an open container, and the table wiping. The advantages of path-following concepts compared to trajectory-following and Cartesian reference paths over joint-space reference paths are elaborated. The last scenario highlights the possibility of collision avoidance for the entire robot's kinematic chain.

Due to the advantageous real-time replanning capabilities and the easy asymmetric Cartesian bounding, we plan to use *BoundMPC* in dynamically changing environments in combination with cognitive decision systems that adapt the pose reference paths online according to situational needs.

## Acknowledgments

The authors acknowledge TU Wien Bibliothek for financial support through its Open Access Funding Program.

## Declaration of conflicting interests

The author(s) declared no potential conflicts of interest with respect to the research, authorship, and/or publication of this article.

## Funding

The author(s) received no financial support for the research, authorship, and/or publication of this article.

## ORCID iDs

Thies Oelerich https://orcid.org/0009-0009-1004-0173
Florian Beck https://orcid.org/0000-0003-3407-4588
Christian Hartl-Nesic https://orcid.org/0000-0003-3054-9435

## Supplemental Material

Supplemental material for this article is available online.